\documentclass[lettersize,journal]{IEEEtran}
\usepackage{amsmath,amsfonts}
\usepackage{algorithmic}
\usepackage{algorithm}
\usepackage{array}
\usepackage[caption=false,font=footnotesize]{subfig}
\usepackage{textcomp}
\usepackage{stfloats}
\usepackage{url}
\usepackage{verbatim}
\usepackage{graphicx}
\usepackage{cite}
\usepackage{amsthm,amssymb}

\newtheorem{theorem}{Theorem}

\usepackage{xcolor}
\usepackage{multirow, booktabs, colortbl, threeparttable, makecell, rotating}  
\usepackage[]{hyperref}

\def\ie{\emph{i.e.}}
\def\eg{\emph{e.g.}}
\def\cf{\emph{cf.}}

\def\urone{\uppercase\expandafter{\romannumeral1}}
\def\urtwo{\uppercase\expandafter{\romannumeral2}}

\begin{document}

\title{Enhanced Ideal Objective Vector Estimation for Evolutionary Multi-Objective Optimization}

\author{Ruihao Zheng, Zhenkun Wang,~\IEEEmembership{Senior Member,~IEEE}, Yin Wu, and Maoguo Gong~\IEEEmembership{Fellow,~IEEE}
\thanks{R. Zheng is with the School of Automation and Intelligent Manufacturing, Southern University of Science and Technology, Shenzhen, 518055, P.R. China. (e-mails: 12132686@mail.sustech.edu.cn, wangzhenkun90@gmail.com)}
\thanks{Z. Wang is with the School of Automation and Intelligent Manufacturing and also with the Department of Computer Science and Engineering, Southern University of Science and Technology, Shenzhen 518055, P.R. China. (e-mail: wangzhenkun90@gmail.com)}
\thanks{Y. Wu is with the Thrust of Artificial Intelligence, Hong Kong University of Science and Technology (Guangzhou), Guangzhou 511400, P.R. China. (e-mail: ywu450@connect.hkust-gz.edu.cn)}
\thanks{M. Gong is with the Academy of Artificial Intelligence, College of Mathematics Science, Inner Mongolia Normal University, Hohhot 010022, P.R. China, and also with the Key Laboratory of Collaborative Intelligence Systems, Ministry of Education, School of Electronic Engineering, Xidian University, Xi'an, 710071, P.R. China. (e-mail: gong@ieee.org)}
\thanks{Corresponding author: Zhenkun Wang.}

}




\maketitle

\begin{abstract}
The ideal objective vector, which comprises the optimal values of the $m$ objective functions in an $m$-objective optimization problem, is an important concept in evolutionary multi-objective optimization. Accurate estimation of this vector has consistently been a crucial task, as it is frequently used to guide the search process and normalize the objective space. Prevailing estimation methods all involve utilizing the best value concerning each objective function achieved by the individuals in the current or accumulated population.
However, this paper reveals that the population-based estimation method can only work on simple problems but falls short on problems with substantial bias. The biases in multi-objective optimization problems can be divided into three categories, and an analysis is performed to illustrate how each category hinders the estimation of the ideal objective vector. Subsequently, a set of test instances is proposed to quantitatively evaluate the impact of various biases on the ideal objective vector estimation method.
Beyond that, a plug-and-play component called enhanced ideal objective vector estimation (EIE) is introduced for multi-objective evolutionary algorithms (MOEAs). EIE features adaptive and fine-grained searches over $m$ subproblems defined by the extreme weighted sum method. EIE finally outputs $m$ solutions that can well approximate the ideal objective vector.
In the experiments, EIE is integrated into three representative MOEAs. To demonstrate the wide applicability of EIE, algorithms are tested not only on the newly proposed test instances but also on existing ones. The results consistently show that EIE improves the ideal objective vector estimation and enhances the MOEA's performance.
\end{abstract}


\section{Introduction}
The multi-objective optimization problem (MOP) can be written as
\begin{equation}
    \begin{array}{ll}
        \text{min.} & \mathbf{f}(\mathbf{x} )=\left(f_1(\mathbf{x}), \ldots, f_m(\mathbf{x})\right)^{\intercal},
        \\
        \text{s.t.} & \mathbf{x} \in \Omega,
    \end{array}
\end{equation}
where $\Omega\subset\mathbb{R}^n$ is the feasible region, and $\mathbf{x} = \left( x_1,\ldots,x_n\right)^\intercal$ is an $n$-dimensional decision vector (also called solution). $\mathbf{f}: \mathbb{R}^n \rightarrow \mathbb{R}^m$ consists of $m$ objective functions, and $\mathbf{f}(\mathbf{x})$ is the objective vector corresponding to $\mathbf{x}$. The feasible objective region can be expressed as $Z=\{\mathbf{f}(\mathbf{x}) \mid \mathbf{x} \in \Omega\}$.

Given two $m$-dimensional vectors $\mathbf{u}=\left(u_1, \ldots, u_m\right)^{\intercal}$ and $\mathbf{v}=\left(v_1,\ldots, v_m\right)^{\intercal}$, $\mathbf{u}$ is said to \textbf{\em dominate} $\mathbf{v}$ if and only if $u_i \leq v_i$ for all $i\in \{1,\ldots,m\}$ and $u_{j}<v_{j}$ for at least one $j\in \{1,\ldots,m\}$. A solution $\mathbf{x}^* \in \Omega$ is said to be \textbf{\em Pareto-optimal} if there is no solution $\mathbf{x} \in \Omega$ such that $\mathbf{f}(\mathbf{x})$ dominates $\mathbf{f}(\mathbf{x}^*)$. Correspondingly, $\mathbf{f}(\mathbf{x}^*)$ is referred to as a Pareto-optimal objective vector. The set of all Pareto-optimal solutions is called the \textbf{\em Pareto set} ($PS$), and the set $\{\mathbf{f}(\mathbf{x}) \mid \mathbf{x} \in PS\}$ is known as the \textbf{\em Pareto front} ($PF$). The \textbf{\em ideal objective vector} $\mathbf{z}^{ide}=\left(z_1^{ide}, \ldots, z_m^{ide}\right)^{\intercal}$ is defined as $z_i^{ide}=\inf_{\mathbf{x} \in \Omega} f_i(\mathbf{x})$ for $i=1,\ldots,m$, while the \textbf{\em nadir objective vector} $\mathbf{z}^{nad} = (z^{nad}_1,\ldots,z^{nad}_m)^\intercal$ is defined as $z^{nad}_i = \sup_{\mathbf{x}\in PS} f_{i}(\mathbf{x})$ for $i=1,\ldots,m$.

The multi-objective optimization evolutionary algorithm (MOEA) is widely used to solve MOPs by virtue of its population-based search for the entire $PF$~\cite{deb2002fast,zitzler2004indicator,zhang2007moea,li2015bi,zheng2023mathematical}. The ideal objective vector, as a fundamental concept in multi-objective optimization, also plays a key role in the MOEA:
\begin{itemize}
    \item Search Guidance. Many MOEAs employ the ideal objective vector to guide the population's evolution towards the $PF$. For example, decomposition-based MOEAs often adopt it as the reference point~\cite{zhang2007moea,deb2014evolutionary,asafuddoula2018enhanced}. Besides, some convergence indicators are defined based on the ideal objective vector~\cite{liu2017many}. Moreover, the Pareto-optimal solutions associated with the ideal objective vector can guide the MOEAs' populations and improve their performance~\cite{gong2023effects}. 
    \item Normalization. In practice, the $m$ objective functions of an MOP often have different value ranges, and normalization of the objective space is essential~\cite{he2021survey,he2023effects}. Each component of the ideal objective vector is often used as the lower bound for normalizing the corresponding objective function~\cite{li2015interrelationship,ishibuchi2017effect,su2023adapting}.
\end{itemize}


Despite the essential role of the ideal objective vector, it cannot be known before solving the MOP. Several estimation methods have been proposed. Typically, the estimated ideal objective vector (denoted as $\mathbf{z}^{e}$) is determined by the objective function's best values currently obtained by the MOEA~\cite{zhang2007moea,deb2014evolutionary}. In each iteration, $\mathbf{z}^{e}$ is updated by
\begin{equation}\label{eqn:pop_ide_iter}
    z_i^{e}=\min \left\{z_i^{e}, \min_{\mathbf{x} \in Q} f_i\left(\mathbf{x}\right)\right\} \text{ for } i=1,\ldots,m,
\end{equation}
where $Q$ is the new offspring set. Alternatively, $z_i^{e}$ can be updated by taking the minimum value from the combined solution set of $Q$ and the current population~\cite{cai2018constrained,tian2019strengthened}. Some studies consider a more optimistic estimation~\cite{qi2017utopian,wang2019generator}, \ie,
\begin{equation}
    z_i^{e} = z_i^\prime - \beta_i \text{ for } i=1,\ldots,m,
\end{equation}
where $z_i^\prime$ can be calculated by Eq.~\eqref{eqn:pop_ide_iter} and $\beta_i$ is a non-negative value. $\beta_i$ is also suggested to decrease as the number of generations increases~\cite{ishibuchi2016reference, wang2017effect}. Furthermore, the method in a recent study~\cite{wu2023decomposition} uses previously generated solutions to build a lightweight surrogate model for each objective function, and then predicts the ideal objective vector according to the built models. Overall, these methods all estimate the ideal objective vector indirectly, \ie, leverage only the information provided by current and previous populations of an MOEA.

\begin{figure*}[ht]
    \centering
    \subfloat[DTLZ2]{\includegraphics[width = 0.33\linewidth]{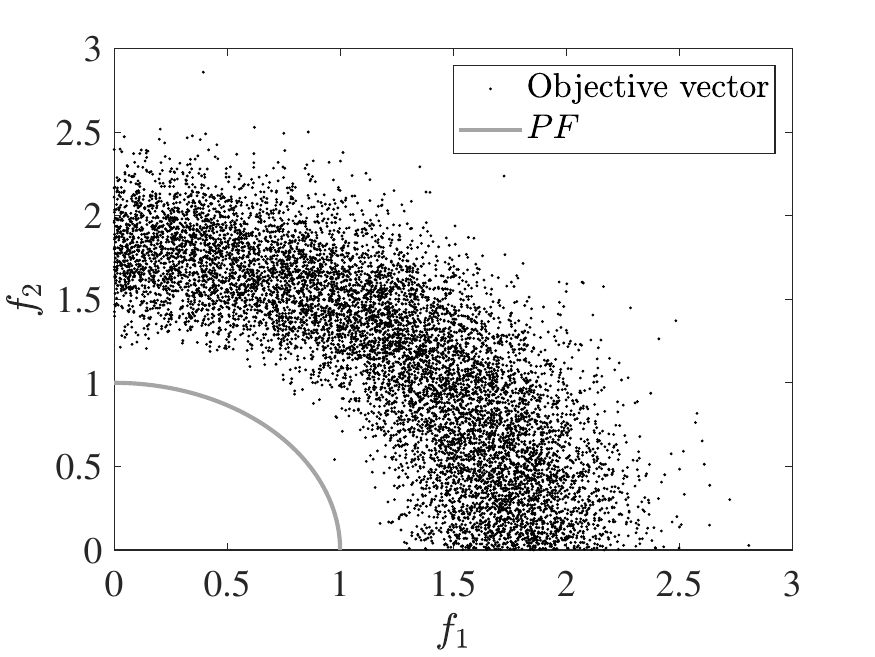}\label{fig:DTLZ2_sample}}
    \subfloat[Four bar truss design problem]{\includegraphics[width = 0.33\linewidth]{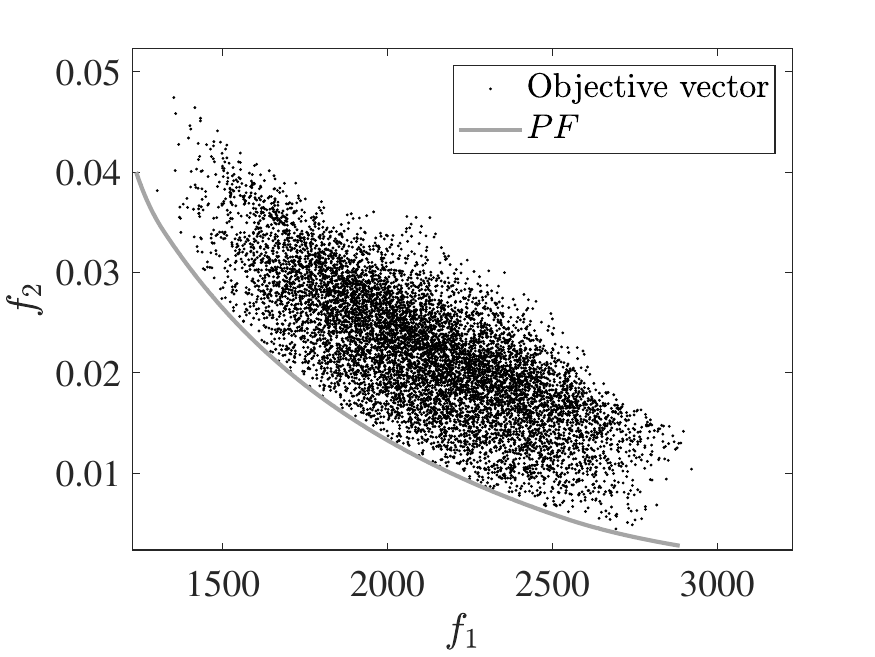}\label{fig:RE241_sample}}
    \subfloat[Multi-objective knapsack problem]{\includegraphics[width = 0.33\linewidth]{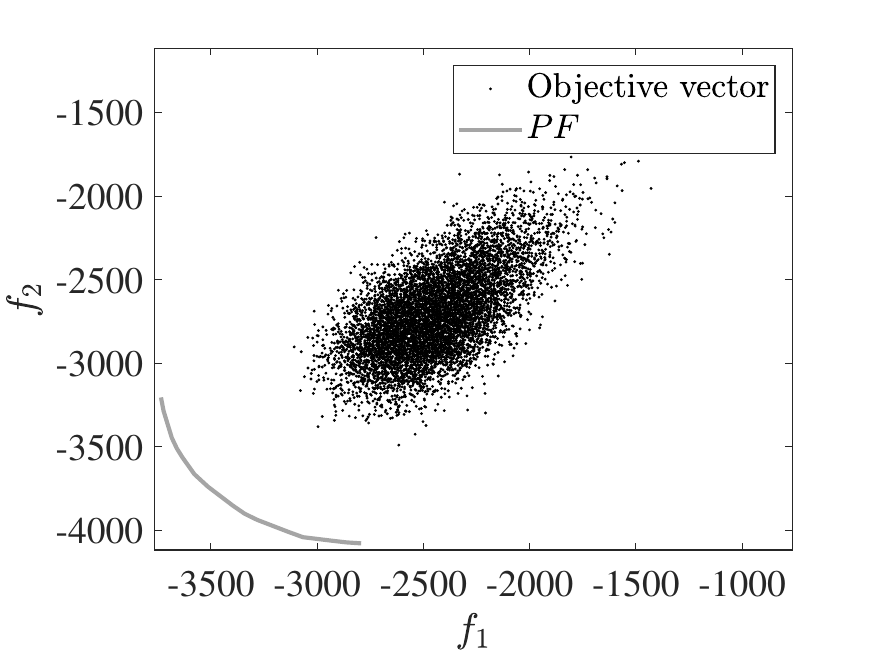}\label{fig:MOKP_sample}}
    \caption{Plots of 10000 random solutions. The $PF$ of (b) is taken from~\cite{tanabe2020easy}. The $PF$ of (c) is obtained by the exact solver.}
\end{figure*}

These population-based methods have demonstrated their effectiveness in some test problems such as DTLZ2~\cite{deb2005scalable}. As shown in \figurename~\ref{fig:DTLZ2_sample}, these test problems have a special feature: the randomly generated solutions have a very broad distribution in their objective space. In other words, the ideal objective vector can be well estimated even from the randomly generated solutions. However, real-world MOPs, such as the four bar truss design problem~\cite{cheng1999generalized} (\cf~\figurename~\ref{fig:RE241_sample}) and the multi-objective knapsack problem~\cite{zitzler1999multiobjective} (\cf~\figurename~\ref{fig:MOKP_sample}), do not have such a feature. \figurename~\ref{fig:RE241_sample} shows that the objective space of the four bar truss design problem is biased towards the center of the $PF$, and the objective vectors of random solutions fail to reach the two boundaries of the $PF$. As shown in \figurename~\ref{fig:MOKP_sample}, the objective vectors of random solutions to the multi-objective knapsack problem are far from the $PF$ and lack diversity in broadness. In these biased cases, the population-based method may face challenges in estimating the ideal objective vector.

In this work, the biases in multi-objective optimization are divided into three categories, \ie, distance-related bias, position-related bias, and mixed bias. We reveal that each category can bring significant challenges to population-based estimation methods. To experimentally assess the performance of MOEAs with different estimation methods, we generalize the works in~\cite{li2017biased,wang2019generator} and introduce a novel test problem generator. Test problems are often constructed by designing position functions and distance functions. To encompass a variety of biases, both position and distance functions in the generator contain bias-related components. As a result, the generator can produce test instances exhibiting biases of various categories and magnitudes. 16 biased test instances are finally determined to evaluate the algorithm's ability in estimating the ideal objective vector.

To address the limitations of population-based estimation methods, we propose a plug-and-play component called enhanced ideal objective vector estimation (EIE) for MOEAs. Unlike population-based methods, EIE directly drives the population of the MOEA to search for the ideal objective vector. Specifically, EIE employs the extreme weighted sum (EWS) method to convert the MOP into $m$ subproblems. By solving these $m$ EWS-based subproblems, $m$ Pareto-optimal solutions that determine the ideal objective vector can be obtained. In EIE, the optimizer for each subproblem can perform a fine-grained search to handle the biased scenarios. Furthermore, EIE executes concurrently with the MOEA. In each iteration, EIE autonomously determines the computational resources to be consumed based on the search status. Subsequently, the solutions generated by EIE are merged into the population of the MOEA, thereby guiding the population toward the ideal objective vector.
In the experimental studies, EIE is integrated into three representative algorithms, each corresponding to a distinct category of MOEAs: dominance-based, decomposition-based, and indicator-based. The test instances include both the 16 proposed ones and the 55 existing ones to validate the broad applicability of EIE. The experimental results indicate that MOEAs with EIE significantly outperform their original versions as well as the MOEAs with population-based estimation methods in terms of both the accuracy of ideal objective vector estimation and the approximation to the $PF$. In other words, although EIE occupies the computational resources that belong to the original MOEA, its advantage in estimating the ideal objective vector leads to a substantial improvement in the overall performance.

The remainder of this paper is organized as follows. Section~\ref{sec:motivation} introduces bias and discusses the biases causing difficulties in estimating the ideal objective vector. Section~\ref{sec:problem} proposed the 16 biased test instances. Subsequently, Section~\ref{sec:algorithm} presents EIE in detail. In Section~\ref{sec:experiments}, experiments are conducted to investigate the effectiveness of EIE. Finally, Section~\ref{sec:conclusion} concludes this paper.

\begin{figure*}[ht]
    \centering
    \subfloat[Distance-related bias]{\includegraphics[width = 0.33\linewidth]{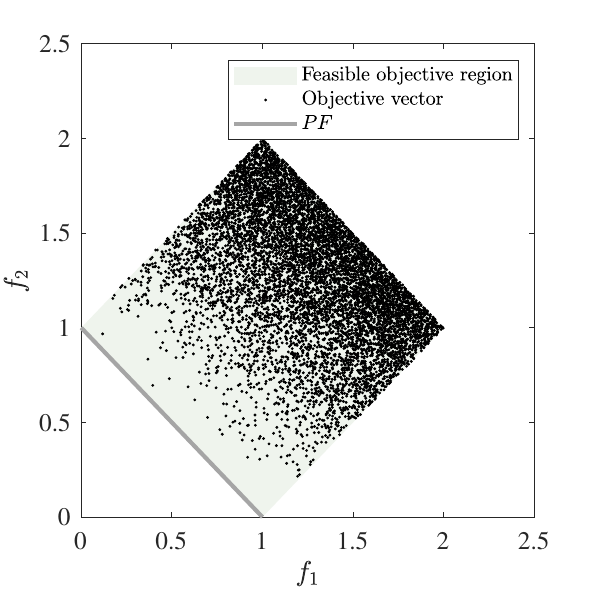}\label{fig:bias_dis_case}}
    \subfloat[Position-related bias]{\includegraphics[width = 0.33\linewidth]{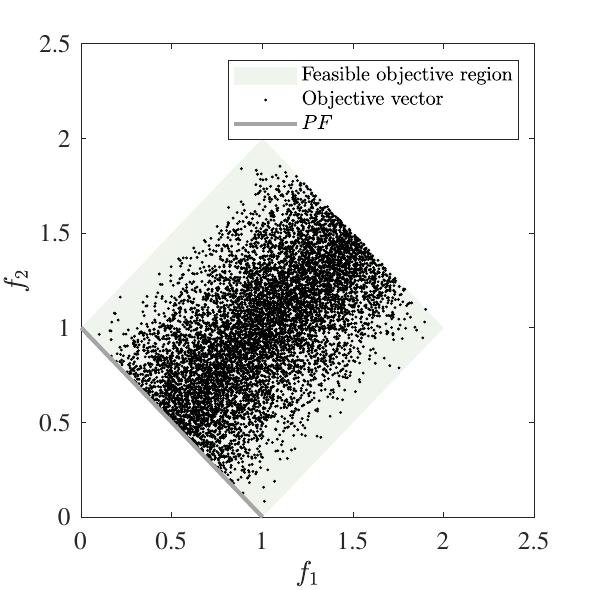}\label{fig:bias_pos_case}}
    \subfloat[Mixed bias]{\includegraphics[width = 0.33\linewidth]{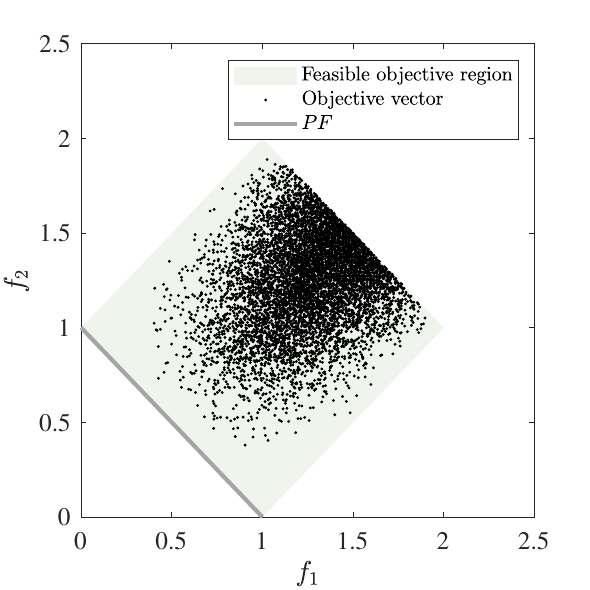}\label{fig:bias_mixed_case}}
    \caption{Plots of random solutions on a two-objective problem with different biases. Letting $x_1,x_2\in[0,1]$, we have: (a) $f_1(x_1,x_2)=x_1+x_2^{0.3}$, $f_2(x_1,x_2)=(1-x_1)+x_2^{0.3}$; (b) if $x_1<0.5$, $f_1(x_1,x_2)=\frac{(2x)^{0.3}}{2}+x_2$ and $f_2(x_1,x_2)=1-\frac{(2x)^{0.3}}{2}+x_2$, or else $f_1(x_1,x_2)=1-\frac{(2-2x)^{0.3}}{2}+x_2$ and $f_2(x_1,x_2)=\frac{(2-2x)^{0.3}}{2}+x_2$; (c) if $x_1<0.5$, $f_1(x_1,x_2)=\frac{(2x)^{0.3}}{2}+x_2^{0.3}$ and $f_2(x_1,x_2)=1-\frac{(2x)^{0.3}}{2}+x_2^{0.3}$, or else $f_1(x_1,x_2)=1-\frac{(2-2x)^{0.3}}{2}+x_2^{0.3}$ and $f_2(x_1,x_2)=\frac{(2-2x)^{0.3}}{2}+x_2^{0.3}$.}
    \label{fig:bias_case}
\end{figure*}

\section{Preliminaries}\label{sec:motivation}

\subsection{Bias}
\begin{figure}[ht]
    \centering
    \includegraphics[width = 0.7\linewidth]{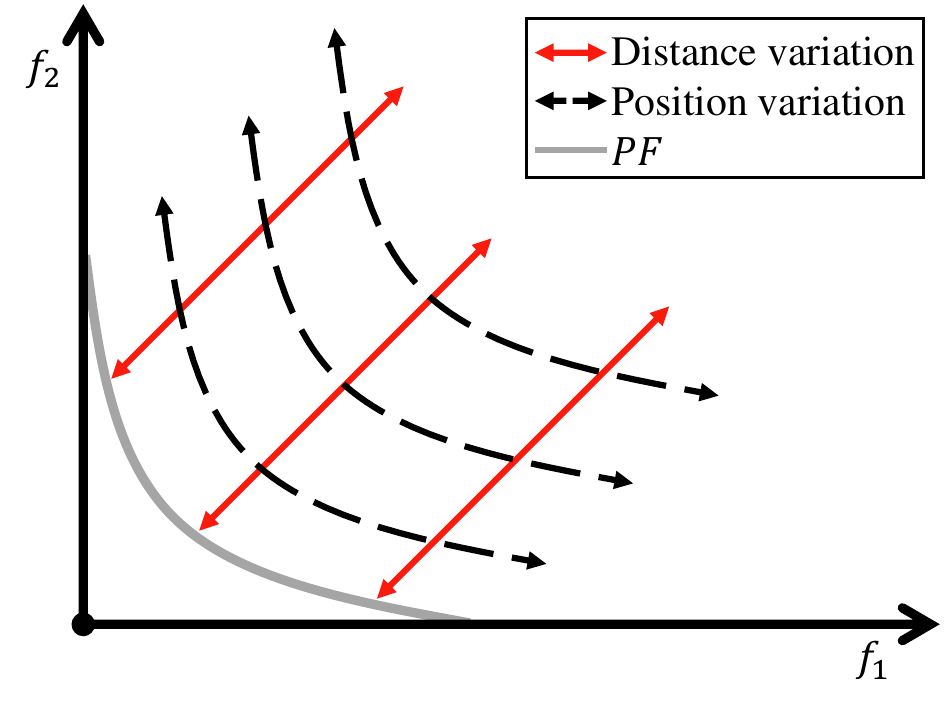}
    \caption{The comparison between the position-related bias and distance-related bias.}
    \label{fig:bias}
\end{figure}

The bias that causes uniformly distributed solutions in the decision space to be mapped into non-uniformly distributed objective vectors in the objective space is common in real-world scenarios and challenging for MOEAs~\cite{deb1999multi,huband2006review,wright2014multi,moshaiov2014mo}. As discussed in~\cite{huband2006review}, bias can be divided into three categories: distance-related bias, position-related bias, and mixed bias. The comparison between position-related bias and distance-related bias can be illustrated by \figurename~\ref{fig:bias}. Distance-related bias affects the distribution of objective vectors towards the $PF$ (\cf, the solid lines in \figurename~\ref{fig:bias}). Position-related bias means that the density of objective vectors is changed with the relative position to the $PF$ (\cf, the dash lines in \figurename~\ref{fig:bias}). Lastly, mixed bias holds both features.

We give some examples of MOPs with biases. To visualize the bias in an MOP, we can examine the density variation of objective vectors when a set of uniformly sampled solutions is considered.
\figurename~\ref{fig:bias_dis_case} shows the MOP with a distance-related bias. The density of objective vectors becomes increasingly low as the distance to the $PF$ decreases. The decreasing density hinders the population of the MOEA from effectively approximating the $PF$. 
\figurename~\ref{fig:bias_pos_case} shows the MOP with a position-related bias, where the density of objective vectors is low on the boundaries but high in the middle. In other words, the objective vectors become sparser as the relative distance to the center of the $PF$ (\ie, $(0.5,0.5)^\intercal$) increases, while the distance to the $PF$ does not affect the density. The MOEA tends to exploit the central part of the $PF$ in this case. Generally, the position-related bias makes the population approximate certain parts of the $PF$ easier than others.
Mixed bias that combines distance-related bias and position-related bias is more frequently encountered in real-world scenarios. In \figurename~\ref{fig:bias_mixed_case}, we can observe that most objective vectors of random solutions are far away from the $PF$ as well as two boundaries of the feasible objective region. Mixed bias can greatly increase the variety and complexity of MOPs. The MOEA has more difficulties in balancing between convergence and diversity on MOPs with mixed biases.

\subsection{Bias Challenges in Ideal Objective Vector Estimation}\label{sec:bias_ideal}
\begin{figure}[ht]
    \centering
    \subfloat[]{\includegraphics[width=0.45\linewidth]{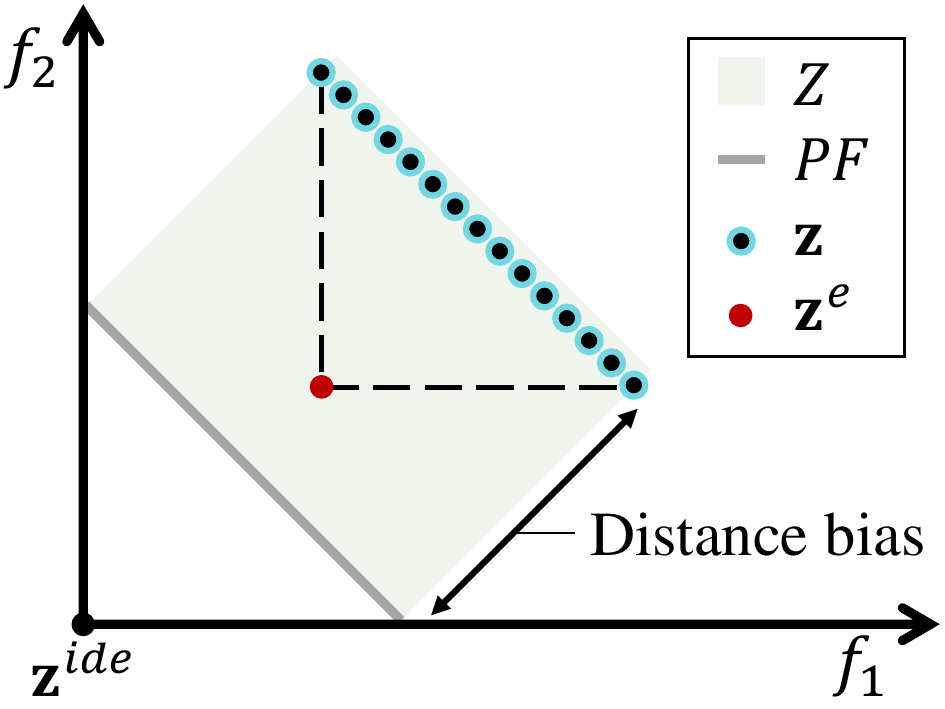}\label{fig:bias_dis_1}}
    \hfil
    \subfloat[]{\includegraphics[width=0.45\linewidth]{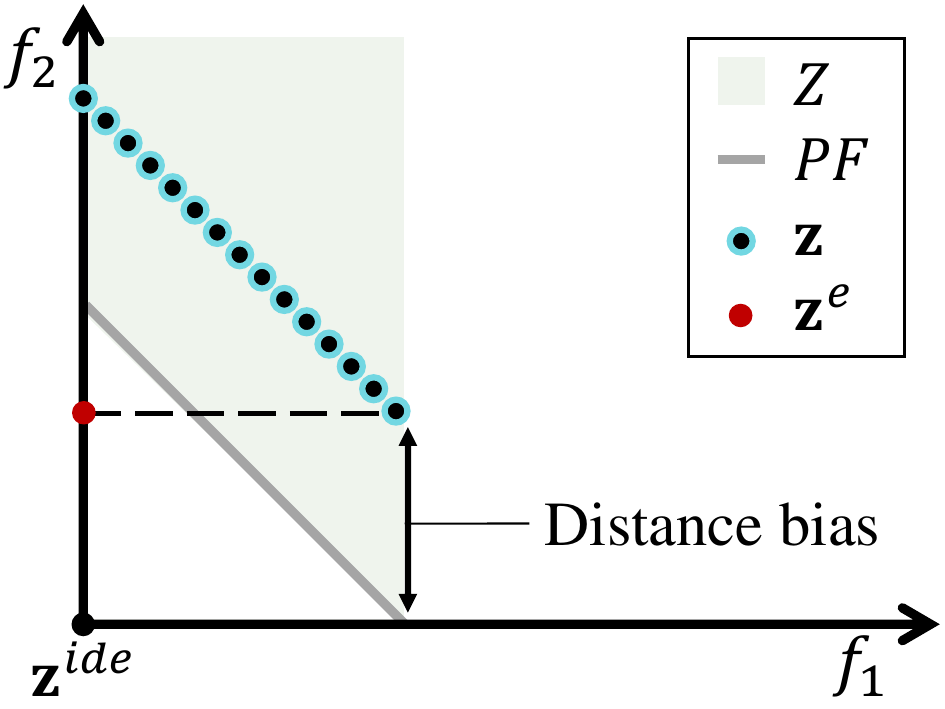}\label{fig:bias_dis_2}}
    \caption{Plots of the population-based estimated ideal objective vector on problems with distance-related bias.}
    \label{fig:bias_dis}
\end{figure}

\begin{figure}[ht]
    \centering
    \subfloat[]{\includegraphics[width = 0.45\linewidth]{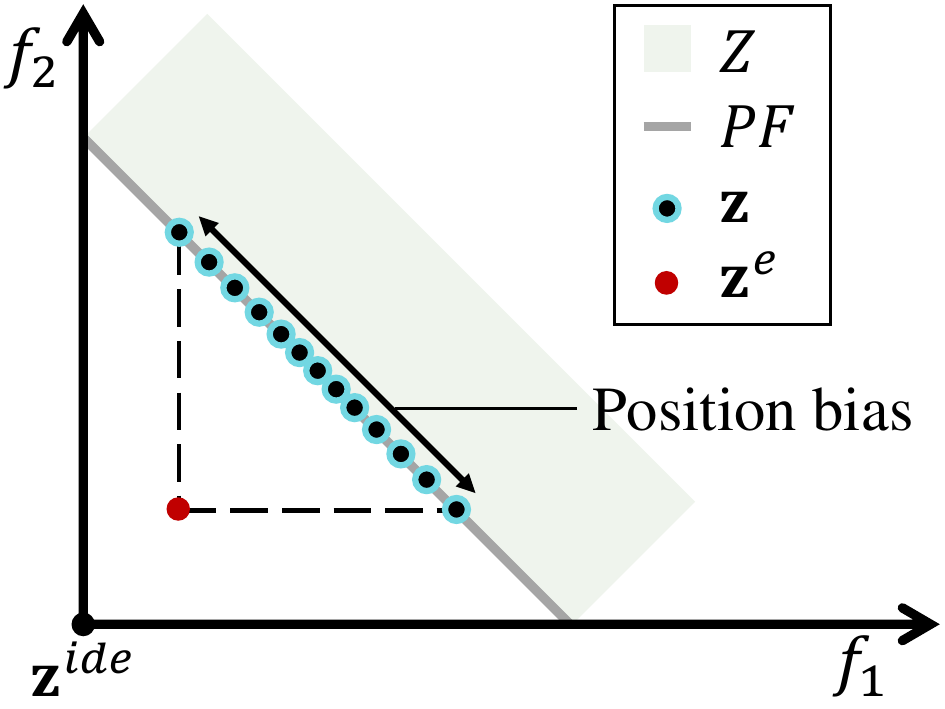}\label{fig:bias_pos_1}}
    \hfil
    \subfloat[]{\includegraphics[width = 0.45\linewidth]{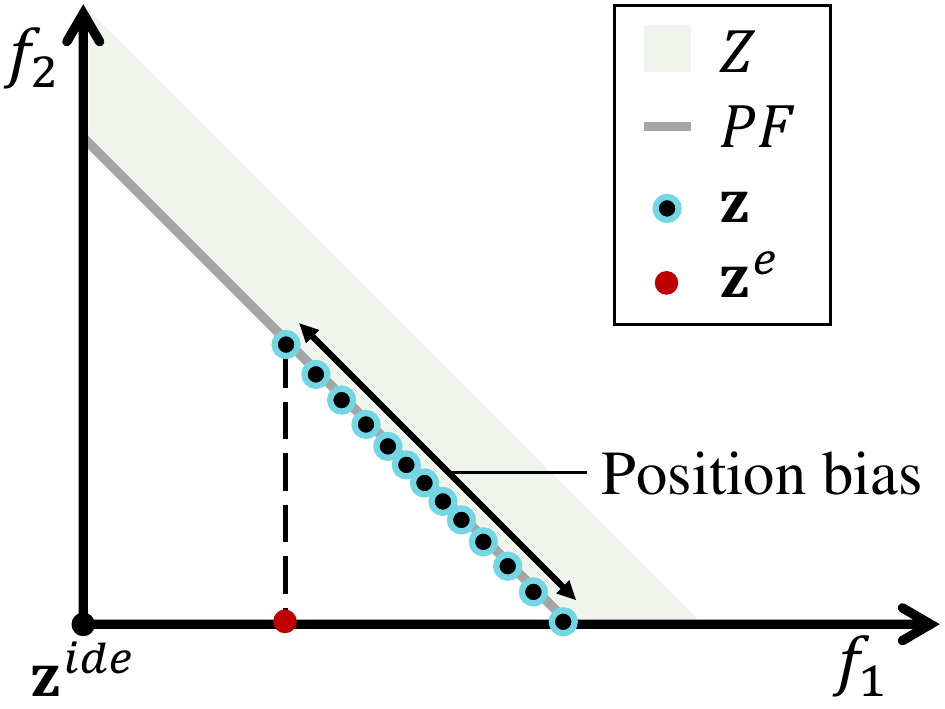}\label{fig:bias_pos_2}}
    \caption{Plots of the population-based estimated ideal objective vector on problems with position-related bias.}
    \label{fig:bias_pos}
\end{figure}

\begin{figure}[ht]
    \centering
    \subfloat[]{\includegraphics[width = 0.45\linewidth]{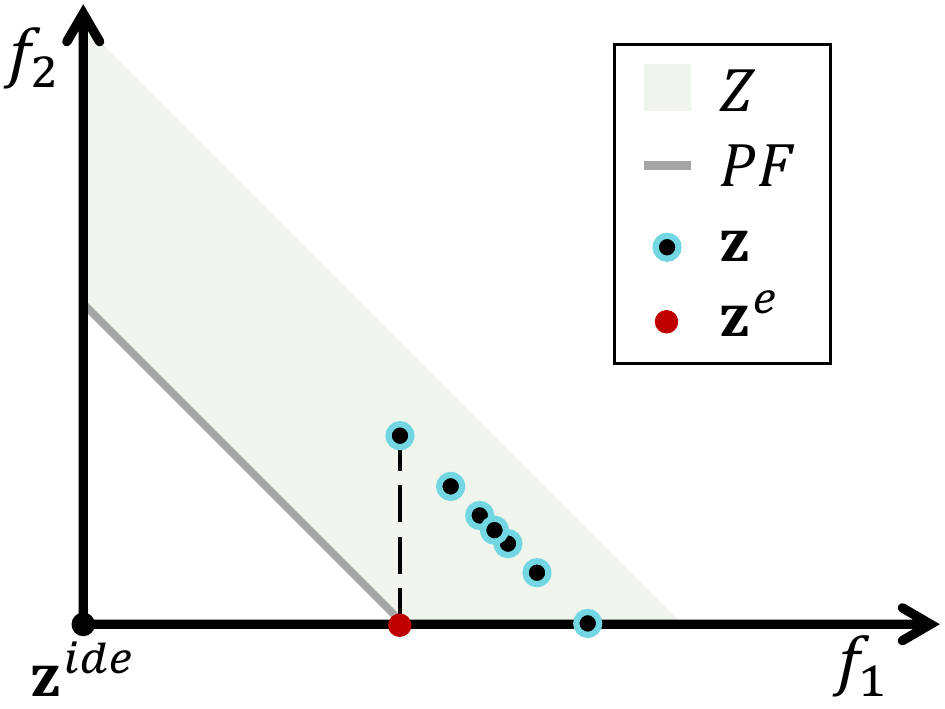}\label{fig:bias_mixed_1}}
    \hfil
    \subfloat[]{\includegraphics[width = 0.45\linewidth]{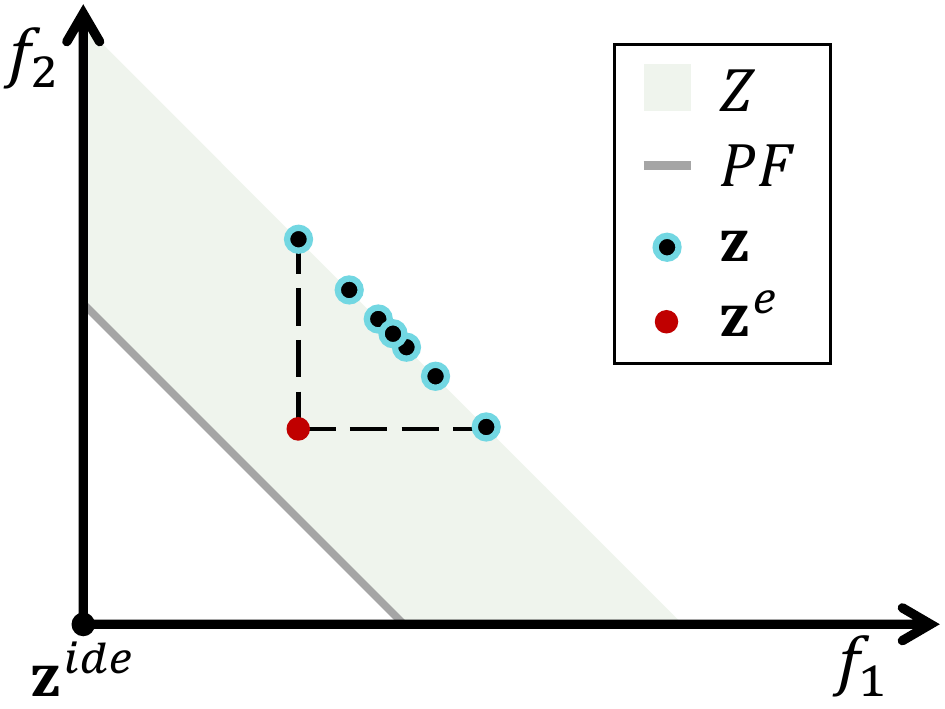}\label{fig:bias_mixed_2}}
    \caption{Plots of the population-based estimated ideal objective vector on problems with mixed bias.}
    \label{fig:bias_mixed}
\end{figure}

This paper focuses on the difficulties caused by biases for an MOEA in approximating the ideal objective vector. The biases affecting the ideal objective vector estimation can also be classified into three categories, \ie, distance-related bias, position-related bias, and mixed bias. In the following, we illustrate the shortcomings of population-based methods when encountering problems with these kinds of biases. In \figurename~\ref{fig:bias_dis}-\ref{fig:bias_mixed}, $\mathbf{z}$ represents an objective vector of the population, and $\mathbf{z}^e$ is the estimated ideal objective vector.

In \figurename~\ref{fig:bias_dis}, we use two examples to illustrate the impact on ideal objective vector estimation when having only distance-related biases but no position-related biases. As shown in \figurename~\ref{fig:bias_dis_1}, the distance-related bias results in a large distance between the estimated ideal objective vector and the true one. No component of the ideal objective vector can be accurately estimated. Another instance with a different feasible objective region, shown in \figurename~\ref{fig:bias_dis_2}, highlights a similar dilemma for the population-based method. In this case, the population-based method can only achieve $z_1^{ide}$ but not $z_2^{ide}$. If the feasible objective region of the MOP is changed to that of \figurename~\ref{fig:bias_pos_2} or \figurename~\ref{fig:bias_mixed} (like DTLZ2), having only distance-related biases no longer hinders the estimation of the ideal objective vector. The boundary objective vectors in the population already obtain the optimal value for each objective function (\cf~\figurename~\ref{fig:DTLZ2_sample}).

In \figurename~\ref{fig:bias_pos}, we use two examples to illustrate the impact on ideal objective vector estimation when having only position-related biases but no distance-related biases. \figurename~\ref{fig:bias_pos_1} shows that the population is biased towards the center of the $PF$, while \figurename~\ref{fig:bias_pos_2} displays that the population only explores the bottom region of the $PF$. In these two cases, the population-based method also fails to estimate the ideal objective vector on some or even all objectives.

As shown in \figurename~\ref{fig:bias_mixed}, the mixed bias further poses both dilemmas for the population. The population not only remains distant from the $PF$ but also lacks spread to some boundaries of the feasible objective region, occupying only a small portion. Consequently, the mixed bias brings greater error for the estimated ideal objective vector. Overall, biases lead to poor population convergence and diversity for an MOEA, causing population-based methods to fail to estimate the ideal objective vector accurately. It is important to develop more appropriate approaches for the ideal objective vector estimation.

\section{Biased Test Problem}\label{sec:problem}

\begin{figure}[t]
\centering
\includegraphics[width = 0.8\linewidth]{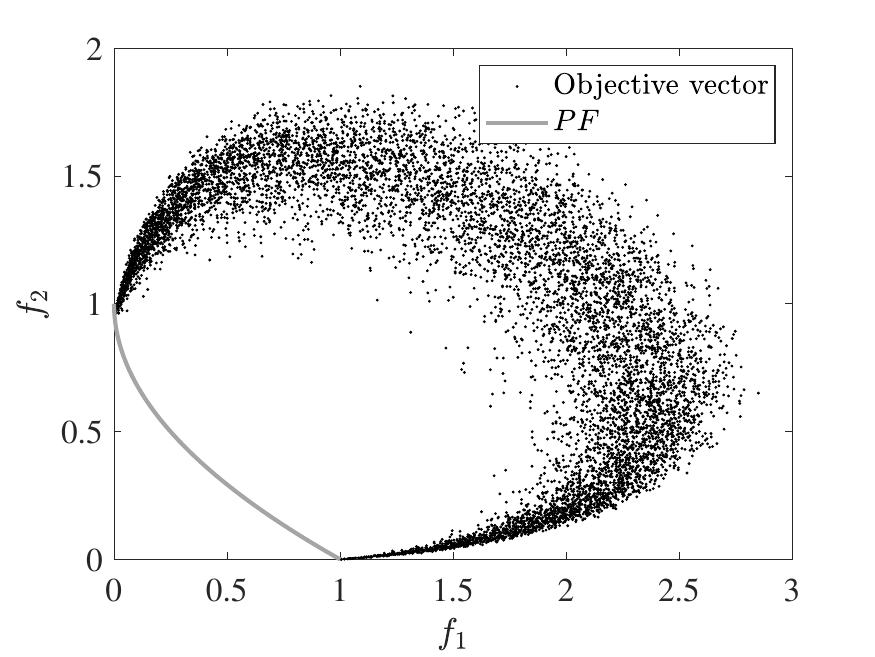}
\vspace{-0.5ex}
\caption{Plot of 10000 random solutions in the objective space of MOP1 in~\cite{liu2014decomposition}.}
\label{fig:MOEADM2M_F1_sample}
\end{figure}

\begin{figure}[t]
\centering
\includegraphics[width = 0.8\linewidth]{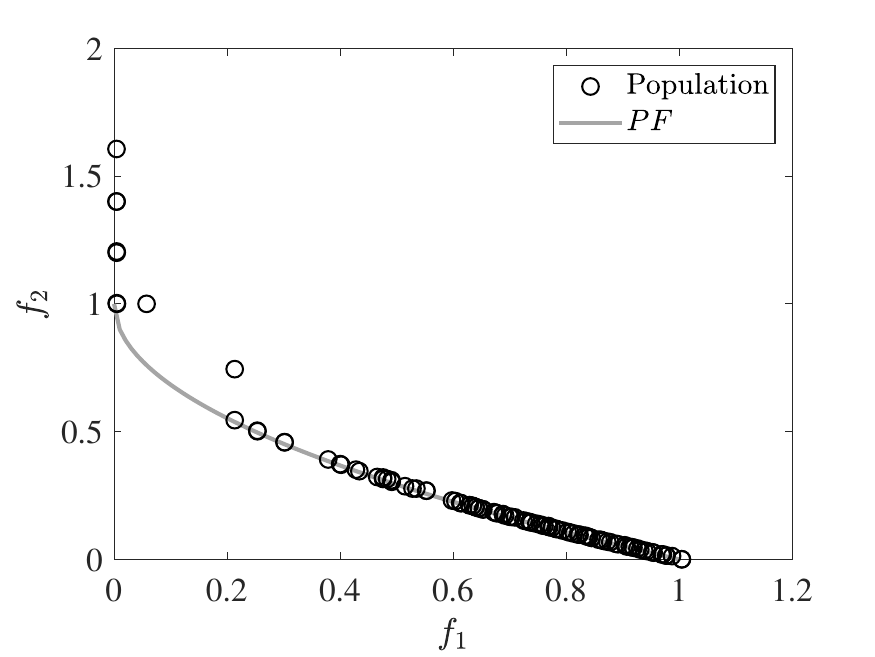}
\vspace{-0.5ex}
\caption{Plot of the final population of NSGA-{\urtwo}~\cite{deb2002fast} on BT3~\cite{li2017biased}. The maximum number of function evaluations is 50000.}
\label{fig:BT3_NSGAII_50000}
\end{figure}

\subsection{Existing Test Problems}\label{sec:existing_pro}
The objective function in defining the multi-objective test problem can be divided into distance function and position function. The position function defines the $PF$ while the distance function controls the distance from the objective vector of a solution to the $PF$. Distance-related bias can be constructed by introducing the power function for the distance function~\cite{deb2005scalable,zapotecas2023challenging}. Position-related bias can be achieved by modifying the position function~\cite{li2017biased,matsumoto2019multiobjective} or the distance function~\cite{liu2014decomposition,liu2017investigating,wang2019generator}.
However, none of these existing test problems are competent. Their drawbacks are listed as follows:
\begin{itemize}
    \item They lack flexibility and control over their biases, preventing full coverage of all biases discussed in Section~\ref{sec:bias_ideal}. The test problems in~\cite{wang2019generator} bias the population towards the central part of the $PF$, which only satisfies the case in \figurename~\ref{fig:bias_mixed_2}. Some test problems even have easy-to-approximate boundaries~\cite{liu2014decomposition,liu2017investigating,zhou2019set}. Their ideal objective vectors can be well estimated by randomly generated solutions. For example, in \figurename~\ref{fig:MOEADM2M_F1_sample}, the random solutions of MOP1 in~\cite{liu2014decomposition} can obtain the extreme points of the $PF$ where the ideal objective vector is extracted.
    \item Many test problems enable decision vectors with special values to achieve the optimal objective function value. When the distance function values are all zero, the lower and upper values of position-related variables map to the extreme points of their $PF$s. This is extremely rare in real-world scenarios, and one can easily obtain the optimal objective function value using the cutting-off operation~\cite{takahama2009solving}. For example, NSGA-{\urtwo}~\cite{deb2002fast} applies this operation and runs on BT3~\cite{li2017biased}. \figurename~\ref{fig:BT3_NSGAII_50000} depicts that NSGA-{\urtwo} can identify the optimum of $f_2$ without fully exploring the upper left portion of the $PF$.
\end{itemize}

We design a test problem generator to overcome the aforementioned issues. This generator features adjustable distance-related and position-related biases. The distance-related bias arises from the distance function, whereas the position-related bias can be controlled by both position and distance functions. Additionally, the generator allows for the customization of the feasible objective region. Any case stated in Section~\ref{sec:bias_ideal} can have corresponding instances obtained by the generator.

\subsection{Proposed Test Problem Generator}
Each objective function of the proposed MOP takes the following form:
\begin{equation}\label{eqn:obj_fun}
\begin{aligned}
    f_i(\mathbf{x}) = &~ w_i\left(h_i(\mathbf{x}_{\urone}) + g_i(\mathbf{x}_{\urone},\mathbf{x}_{\urtwo})\right) \\
    = &~ w_i\left(h_i(\mathbf{x}_{\urone}) + \sum_{j=1}^m\Theta_{ij}g_j^\prime(\mathbf{x}_{\urone},\mathbf{x}_{\urtwo})\right),
\end{aligned}
\end{equation}
where $\mathbf{x}\in\Omega = [0,1]^{s}\times [-1,1]^{n-s}\subset\mathbb{R}^n$, $\mathbf{x}_{\urone} =(x_1,\ldots,x_{s})^{\intercal}$, and $\mathbf{x}_{\urtwo}=(x_{s+1},\ldots,x_n)^{\intercal}$. $g_i(\mathbf{x}_{\urone},\mathbf{x}_{\urtwo})\geq 0$ is called the distance function and $h_i(\mathbf{x}_{\urone})$ is termed as the position function. If a solution $\mathbf{x}^*$ can make all distance function values equal to $0$, it is a Pareto-optimal solution. $\mathbf{w}$ is a scale vector, where $w_i>0$ is a scale factor for the $i$-th objective function. $\mathbf{w}$ can create different ranges for different objective functions. $\Theta$ is a weight matrix, where elements in the same row are weights of distance functions for each objective function. $\Theta$ can adjust the feasible objective region. When $\Theta=\left[\begin{smallmatrix} 0.5 & 0.5 \\ 0.5 & 0.5 \end{smallmatrix}\right]$, all objective functions share the same distance function and the feasible objective region is similar to those in \figurename~\ref{fig:bias_dis_1} and \figurename~\ref{fig:bias_pos_1}; when $\Theta=\left[\begin{smallmatrix} 0 & 0 \\ 0.5 & 0.5 \end{smallmatrix}\right]$, only the 2-nd objective function has a nonzero distance function and the feasible objective region is similar to that in \figurename~\ref{fig:bias_dis_2}; when $\Theta=\left[\begin{smallmatrix} 1 & 0 \\ 0 & 1 \end{smallmatrix}\right]$, different objective functions have different distance functions and the feasible objective region is similar to those in \figurename~\ref{fig:bias_pos_2} and \figurename~\ref{fig:bias_mixed}.

\subsubsection{Position Function}
First, we define $\mathbf{y}(\hat{\mathbf{x}}_{\urone}):[0,1]^{m-1}\rightarrow[0,1]^m$ as a mapping, which returns an $m$-dimensional vector $(y_1,\ldots,y_m)^\intercal$ as follows:
\begin{equation}\label{eqn:base}
    y_i(\hat{\mathbf{x}}_{\urone})=
    \begin{cases}
        1-\hat{x}_1,                             & i=1,            \\
        (1-\hat{x}_i)\prod_{j=1}^{i-1}\hat{x}_j, & i=2,\ldots,m-1, \\
        \prod_{j=1}^{m-1}\hat{x}_j,              & i = m.
    \end{cases}
\end{equation}
Any point generated by this mapping is guaranteed to lie within an ($m-1$)-dimensional unit simplex. We let
\begin{equation}\label{eqn:x_hat}
    \begin{aligned}
         & \hat{x}_i =
        \begin{cases}
            \left(\frac{2^\gamma}{\hat{c}_i^{\gamma-1}}\right)\left|\sigma_i-\frac{\hat{c}_i}{2}\right|^\gamma, & \sigma_i<\hat{c}_i,    \\
            -\frac{2^\gamma}{(1-\hat{c}_i)^{\gamma-1}}\left|\sigma_i-\frac{1+\hat{c}_i}{2}\right|^\gamma + 1,   & \sigma_i>\hat{c}_i, \\
            \sigma_i,   & \sigma_i=\hat{c}_i,
        \end{cases} \\
         & i=1,\ldots,m-1,
    \end{aligned}
\end{equation}
where
\begin{equation}\label{eqn:scalable_x}
\begin{aligned}
    & \sigma_i = \frac{\sum_{j\in J_i}x_j}{|J_i|}, \\
    & J_i=\left\{i,m-1+i,\ldots,\left\lfloor\frac{s-i}{m-1}\right\rfloor (m-1)+i\right\},
\end{aligned}
\end{equation}
and
\begin{equation}\label{eqn:fp_pbs}
    \hat{c}_i = \frac{1-\sum_{j=1}^i c_j^{pos}}{1-\sum_{j=1}^{i-1} c_j^{pos}}.
\end{equation}
The position function in Eq.~\ref{eqn:obj_fun} can be formulated as
\begin{equation}\label{eqn:pos_fun}
    h_i(\mathbf{x}_{\urone}|\mathbf{p},\mathbf{c}^{pos},\gamma) = y_i(\hat{\mathbf{x}}(\mathbf{x}_{\urone}|\mathbf{c}^{pos},\gamma))^{p_i},
\end{equation}
where $\mathbf{c}^{pos}$ and $\gamma$ are parameters related to the position-related bias. $\mathbf{c}^{pos}$ satisfies $\sum_{i=1}^m c_i^{pos}=1$ and $c_i^{pos}>0$ for each $i\in \{1,\ldots,m\}$. It determines the position of the easy-to-approximate part. $\gamma<1$ controls the degree of bias. A smaller $\gamma$ means a more difficult bias. $\mathbf{p}$ determines the $PF$ shape. If $p_i=1$ for $i=1,\ldots,m$, the $PF$ is a unit simplex. If $p_i>1$ for $i=1,\ldots,m$, the $PF$ is convex. If $p_i<1$ for $i=1,\ldots,m$, the $PF$ is concave. If $p_i>1$ for some $i$ and $p_j<1$ for some $j$, then the $PF$ contains both convex and concave segments (denoted as mixed $PF$).

\begin{figure}[ht]
    \centering
    \includegraphics[width = 0.8\linewidth]{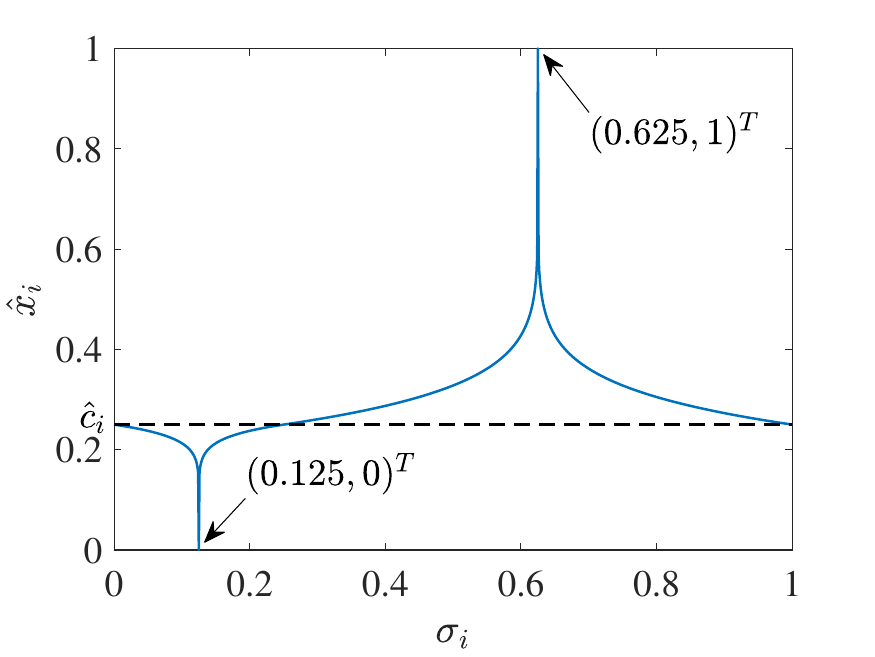}
    \caption{Plot of $\hat{x}_i$ with $\gamma=0.1$ and $\hat{c}_i=0.25$.}
    \label{fig:hat_x_2}
\end{figure}

The critical design of the position function is Eq.~\eqref{eqn:x_hat}.
To explain this, we begin with introducing Eq.~\eqref{eqn:fp_pbs}.
$\mathbf{y}(\hat{\mathbf{x}}_{\urone})$ is on the unit simplex and the $PF$ is constructed based on the unit simplex. Let $\mathbf{y}(\hat{\mathbf{x}}_{\urone})$ be the relative position of the objective vector to the $PF$. Eq.~\eqref{eqn:fp_pbs} is the solution to the following system of equations:
\begin{equation}
    \begin{cases}
        1-\hat{c}_1=c_i^{pos}, & i=1,            \\
        (1-\hat{c}_i)\prod_{j=1}^{i-1}\hat{c}_j=c_i^{pos}, & i=2,\ldots,m-1, \\
        \prod_{j=1}^{m-1}\hat{c}_j=c_i^{pos}, & i = m.
    \end{cases}
\end{equation}
It means that when $\hat{x}_i=\hat{c}_i$ for $i=1,\ldots,m$, the relative position of such a objective vector is exactly $\mathbf{c}^{pos}$. For example, when the distance function values are $0$ and $p_i=1$ for $i=1,\ldots,m$, the corresponding objective vector is $\mathbf{c}^{pos}$.
Then, Eq.~\eqref{eqn:scalable_x} enables scalability in the number of position-related variables through linear combination. The position-related bias is mainly caused by the mapping from $\sigma_i$ to $\hat{\mathbf{x}}_i$. The plot of $\hat{x}_i$ with respect to $\sigma_i$ is shown in \figurename~\ref{fig:hat_x_2}. The lower and upper values are acquired when $\sigma_i=0.125$ and $\sigma_i=0.625$ respectively. $\sigma_i$ has a more rapid change as $\hat{x}_i$ approximates 0 or 1. In contrast, $\sigma_i$ changes slowly when $\hat{x}_i$ approximates 0.25 (\ie, $\hat{c}_i$). Thereby, the relative position of a random objective vector is more likely to be trapped near $\mathbf{c}^{pos}$.

Note that the extreme objective vector (\eg, $(1,0,\ldots,0)^\intercal$) is obtained when $\hat{x}_i=0$ or $1$ for $i=1,\ldots,m$ and the corresponding distance function values are zero. $\sigma_i=0$ and $\sigma_i=1$ are both mapped to $\hat{x}_i=\hat{c}_i$, which are not the lower and upper values of $\hat{x}_i$. Therefore, the position function can avoid its minimal objective function value obtained by the cutting-off operation (see the second drawback in Section~\ref{sec:existing_pro}). 
Moreover, setting $\mathbf{c}^{pos}$ as any unit vector is not recommended. For example, we consider $\mathbf{c}^{pos}=(1,0,\ldots,0)^\intercal$. Then, $c_i$ is either $0$ or $1$. If $\sigma_i$ equals $0$ or $1$, the extreme value of $\hat{x}_i$ can be obtained. Additionally, when $\gamma=1$, the mapping from position-related variables to $\hat{x}_i$ can be regarded as linear. Consequently, the position-related bias in terms of the position function disappears.

In summary, we employ Eq.~\ref{eqn:base} as a basic configuration for the position function. In Eq.~\ref{eqn:x_hat}, we achieve adjustable difficulty and position of the position-related bias via $\gamma$ and $\mathbf{c}^{pos}$, respectively. Besides, the position function is scalable not only in terms of the number of objectives but also in terms of the number of variables as described in Eq.~\ref{eqn:scalable_x}. A parameter called $\mathbf{p}$ is finally introduced in Eq.~\ref{eqn:pos_fun} to enable adjustability of the $PF$ shape.

\subsubsection{Distance Function}
We generalize the distance function in~\cite{wang2019generator} to make it possess an easy-to-approximate part whose position is controllable. To be specific, the distance ratio in this distance function is modified. The new distance ratio (denoted as $\ell$) is calculated by
\begin{equation}
    \ell(\mathbf{x}_{\urone}|\mathbf{c}^{dis}) = \frac{\max(N\cdot(\mathbf{y}(\hat{\mathbf{x}}_{\urone})-\mathbf{c}^{dis}))}{\max\limits_{1\leq i\leq m}\Big(\max\big(N\cdot(\mathbf{e}_i-\mathbf{c}^{dis})\big)\Big)},
\end{equation}
where
\begin{equation}
    N_{ij}=
    \begin{cases}
        -\sqrt{\frac{m-1}{m}}, & i=j, \\
        \frac{1}{\sqrt{m(m-1)}}, & i\neq j,
    \end{cases}
    \quad\in\mathbb{R}^{m\times m},
\end{equation}
and $\mathbf{e}_i$ is a unit vector. $\mathbf{c}^{dis}$ also determines the position of the easy-to-approximate part, satisfying $\sum_{i=1}^m c_i^{dis}=1$ and $c_i^{dis}>0$ for $i=1,\ldots,m$. Each row of $N$ is a normal vector (with a length of 1) to the boundary of ($m-1$)-dimensional unit simplex. The normal vectors are all parallel to the ($m-1$)-dimensional unit simplex.
We let
\begin{equation}
    r=\max(N\cdot(\mathbf{y}(\hat{\mathbf{x}}_{\urone})-\mathbf{c}^{dis})).
\end{equation}
That is, $r$ is the perpendicular distance from $\mathbf{c}^{dis}$ to any boundary of a simplex. The simplex has $\mathbf{c}^{dis}$ as its center, with its boundary intersecting $\mathbf{y}(\hat{\mathbf{x}}_{\urone})$. Similarly, we formulate
\begin{equation}
    r_0=\max\limits_{1\leq i\leq m}(r_i^\prime)=\max\limits_{1\leq i\leq m}\Big(\max\big(N\cdot(\mathbf{e}_i-\mathbf{c}^{dis})\big)\Big).
\end{equation}
The computation of $r_i^\prime$ is similar to that of $r$, where $\mathbf{y}(\hat{\mathbf{x}}_{\urone})$ is replaced with the vertex of unit simplex (\ie, $\mathbf{e}_i$). And $r_0$ is the maximal value among $r_i^\prime,i=1,\ldots,m$ such that $0\leq\ell(\mathbf{x}_{\urone}|\mathbf{c}^{dis})\leq 1$.

\begin{figure}[ht]
    \centering
    \subfloat[]{\includegraphics[width = 0.45\linewidth]{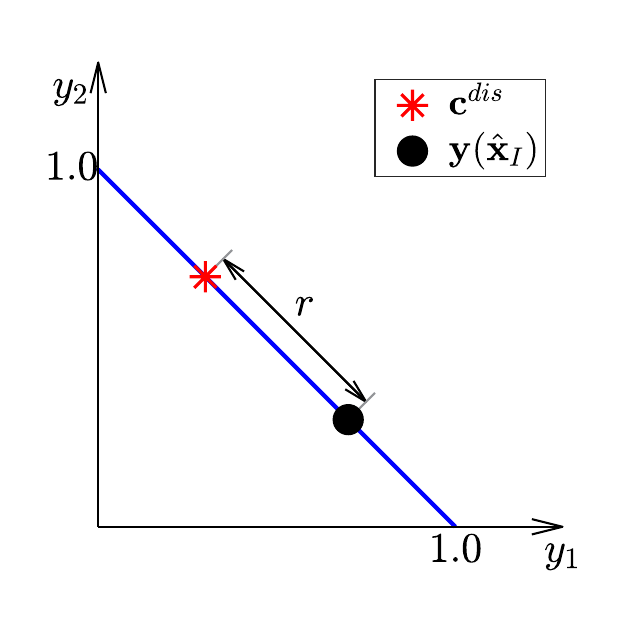}}
    \hfil
    \subfloat[]{\includegraphics[width = 0.45\linewidth]{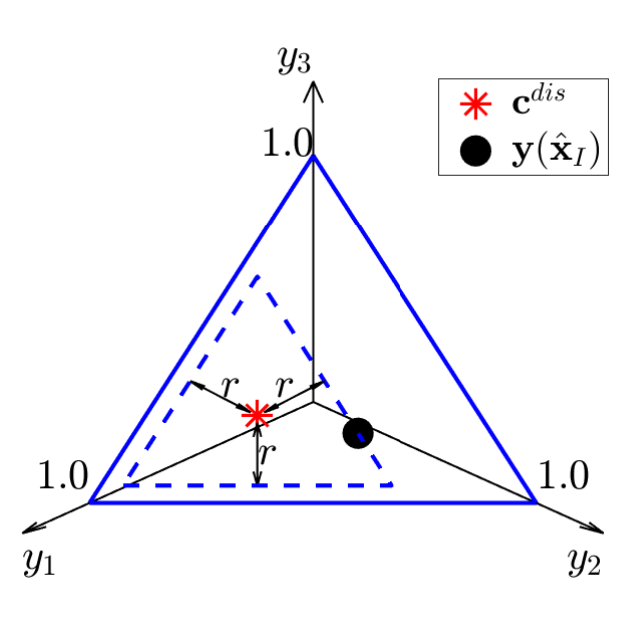}}
    \caption{Illustrations of $r$ in the 2-objective and 3-objective cases.}
    \label{fig:illustration_r}
\end{figure}

\figurename~\ref{fig:illustration_r} shows the computation of $r$. In the 2D case, the simplex is a line and its boundaries are the extreme points. In the 3D case, the simplex is a triangle and its boundaries are three lines. $\mathbf{c}^{dis}$ and $\mathbf{y}(\hat{\mathbf{x}}_{\urone})$ locate on the unit simplex. In the 2D case, $r$ can be simplified as the Euclidean distance from $\mathbf{c}^{dis}$ to $\mathbf{y}(\hat{\mathbf{x}}_{\urone})$. As for the 3D case, a simplex is introduced. The boundaries of this simplex are drawn by dash lines. The simplex satisfies: one of its boundaries passes through $\mathbf{y}(\hat{\mathbf{x}}_{\urone})$ and its central point is $\mathbf{c}^{dis}$. $r$ is the perpendicular distance from $\mathbf{c}^{dis}$ to any dash line. The dash lines can be viewed as a contour: all points on the dash lines have the same $r$ value. That is, $r$ is symmetrical in 2D and triangularly symmetrical in 3D with respect to $\mathbf{c}^{dis}$.

The rest of the distance function remains unchanged. We rewrite the distance function as follows.
The position-dependent scale function $b(\mathbf{x}_{\urone}|\beta,\mathbf{c}^{dis})$ is defined as
\begin{equation}\label{eqn:fun_b}
    b(\mathbf{x}_{\urone}|\beta,\mathbf{c}^{dis}) = \left(\sin\left(\frac{\pi}{2}\cdot\ell(\mathbf{x}_{\urone}|\mathbf{c}^{dis})^{m-1}\right)\right)^{\beta}.
\end{equation}
Then the distance function has the following form:
\begin{equation}\label{eqn:dis_fun}
\begin{aligned}
    g_i^\prime(\mathbf{x}_{\urone}, \mathbf{x}_{\urtwo} | &~ a_1,a_2,a_3,a_4,a_5,\mathbf{c}^{dis}) = \\
    &~ \frac{a_1b(\mathbf{x}_{\urone} | a_4,\mathbf{c}^{dis})+1}{|J_i|} \sum_{j\in J_i}|t_j|^{a_3},
\end{aligned}
\end{equation}
where
\begin{equation}
    J_i=\left\{s\!+\!i,m\!+\!s\!+\!i,\ldots,\left\lfloor\frac{n\!-\!s\!-\!i}{m}\right\rfloor m\!+\!s\!+\!i\right\},
\end{equation}
\begin{equation}\label{eqn:tj}
\begin{aligned}
    & t_j = x_j - \\
    & 0.9b(\mathbf{x}_{\urone}|a_2,\mathbf{c}^{dis}) \cos\left(a_5\pi\ell(\mathbf{x}_{\urone}|\mathbf{c}^{dis})+\frac{(n+2)j\pi}{2n}\right).
\end{aligned}
\end{equation}
Correspondingly, the $PS$ can be written as
\begin{equation}\label{eqn:ps}
    x_j \!=\! 0.9b(\mathbf{x}_{\urone}|a_2,\mathbf{c}^{dis})\cos\!\left(a_5\pi\ell(\mathbf{x}_{\urone}|\mathbf{c}^{dis}) \!+\! \frac{(n\!+\!2)\pi}{2n}\right)
\end{equation}
for $\mathbf{x}_{\urone}\in[0,1]^{s} \mbox{ and } s\geq j\geq n$.

In Eq.~\eqref{eqn:dis_fun}, $a_3$, $a_4$, and $\mathbf{c}^{dis}$ are three parameters related to bias. $a_3$ controls the distance-related bias. A smaller $a_3$ signifies that the objective vectors of random solutions are more sparse near the $PF$.
$a_4$ and $\mathbf{c}^{dis}$ control the position-related bias. Setting $a_4>0$ activates the position-related bias in terms of the distance function; and $\mathbf{c}^{dis}$ plays the same role with $\mathbf{c}^{pos}$. $\frac{a_1 b(\mathbf{x}_{\urone} | a_4,\mathbf{c}^{dis})+1}{|J_i|}$ is the position-related weight on the following summation, where $b(\mathbf{x}_{\urone} | a_4,\mathbf{c}^{dis})$ determines the weight. Note that the function value of $b(\mathbf{x}_{\urone}|\beta,\mathbf{c}^{dis})$ increases with a growing function value of $\ell(\mathbf{x}_{\urone}|\mathbf{c}^{dis})$. Then, the distance function value is large when $a_4>0$ and $r$ is large. In other words, the objective vectors with large relative distances to $\mathbf{c}^{dis}$ can be biased away from the $PF$. As a result, the objective vector density is getting lower as the relative distance to $\mathbf{c}^{dis}$ increases.

Besides, a larger $a_1$ can make a solution mapping into the objective space further away from the $PF$.
$a_2$ and $a_5$ adjust the $PS$ shape. $a_2>0$ scales the $PS$ according to the position of the objective vector. A larger $a_5$ means a more curved $PS$.
Last but not least, $a_1$ to $a_5$ serve the same purpose as $A$ to $E$ in~\cite{wang2019generator} respectively. For more information, please refer to~\cite{wang2019generator}.

\begin{table*}[t]
\footnotesize
  \centering
  \caption{Test instances. $w_i=10^{2(i-1)},i=1,\ldots,m$.}
    \scalebox{1}{
    \begin{threeparttable}
    \renewcommand{\arraystretch}{2}
	\setlength{\tabcolsep}{0.8mm}{
    \begin{tabular}{ccccccccccccccccccc}
    \toprule
    \multirow{2}[4]{*}{Instance} & \multirow{2}[4]{*}{$m$} & \multirow{2}[4]{*}{$n$} & \multicolumn{4}{c}{Position function} &       & \multicolumn{7}{c}{Distance function}                 &       & \multicolumn{3}{c}{Characteristics} \\
\cmidrule{4-7}\cmidrule{9-15}\cmidrule{17-19}          &       &       & $s$     & $\mathbf{p}$     & $\mathbf{c}^{pos}$  & $\gamma$ &       & $\Theta$ & $a_1$    & $a_2$    & $a_3$    & $a_4$    & $a_5$    & $\mathbf{c}^{dis}$  &       & Bias  & $PS$    & \multicolumn{1}{l}{$PF$} \\
    \midrule
    MOP1  & 2     & 7     & 5     & $(1,1)^\intercal$ & $(0.1,0.9)^\intercal$ & 0.1   &       & $\left[\begin{smallmatrix} 1 & 0 \\ 0 & 1 \end{smallmatrix}\right]$ & 1     & 0     & 1     & 0     & 0     & NA    &       & A$^\dagger$ & Linear & Linear \\
    MOP2  & 2     & 7     & 5     & $(0.5,0.5)^\intercal$ & $(0.5,0.5)^\intercal$ & 0.2  &       & $\left[\begin{smallmatrix} 1 & 0 \\ 0 & 1 \end{smallmatrix}\right]$ & 1     & 0     & 2     & 0     & 0     & NA    &       & A$^\dagger$ & Linear & Mixed \\
    MOP3  & 2     & 7     & 1     & $(1,1)^\intercal$ & $(0.3,0.7)^\intercal$    & 1     &       & $\left[\begin{smallmatrix} 0.5 & 0.5 \\ 0.5 & 0.5 \end{smallmatrix}\right]$ & 12     & 0     & 0.1   & 0     & 0   & NA    &       & B & Linear & Linear \\
    MOP4  & 2     & 7     & 1     & $(0.5,2)^\intercal$ & $(0.3,0.7)^\intercal$    & 1     &       & $\left[\begin{smallmatrix} 0 & 0 \\ 0.5 & 0.5 \end{smallmatrix}\right]$ & 6     & 0     & 0.1   & 0     & 0   & NA    &       & B & Linear & Mixed \\
    MOP5  & 2     & 7     & 1     & $(2,2)^\intercal$ & $(0.5,0.5)^\intercal$ & 0.1   &       & $\left[\begin{smallmatrix} 0.5 & 0.5 \\ 0.5 & 0.5 \end{smallmatrix}\right]$ & 6     & 0     & 0.25   & 0     & 0     & NA    &       & A$^\dagger$, B & Linear & Convex \\
    MOP6  & 2     & 7     & 1     & $(0.5,0.5)^\intercal$ & $(0.9,0.1)^\intercal$ & 0.2   &       & $\left[\begin{smallmatrix} 0.8 & 0.2 \\ 0.2 & 0.8 \end{smallmatrix}\right]$ & 3     & 0     & 0.5     & 0     & 0     & NA    &       & A$^\dagger$, B & Linear & Concave \\
    MOP7  & 2     & 7     & 1     & $(2,2)^\intercal$ & $(0,1)^\intercal$    & 1     &       & $\left[\begin{smallmatrix} 0.5 & 0.5 \\ 0.5 & 0.5 \end{smallmatrix}\right]$ & 6     & 4     & 2     & 4     & 3     & $(0.5,0.5)^\intercal$ &       & A$^\ddagger$ & Nonlinear & Convex \\
    MOP8  & 2     & 7     & 1     & $(0.5,2)^\intercal$ & $(0,1)^\intercal$    & 1     &       & $\left[\begin{smallmatrix} 0.8 & 0.2 \\ 0.2 & 0.8 \end{smallmatrix}\right]$ & 12    & 1     & 2   & 1     & 3   & $(0,1)^\intercal$ &       & A$^\ddagger$ & Nonlinear & Mixed \\
    MOP9  & 2     & 7     & 1     & $(2,2)^\intercal$ & $(0.5,0.5)^\intercal$ & 0.2   &       & $\left[\begin{smallmatrix} 0.8 & 0.2 \\ 0.8 & 0.2 \end{smallmatrix}\right]$ & 6     & 1     & 2     & 1     & 3     & $(0.5,0.5)^\intercal$ &       & A$^\S$ & Nonlinear & Convex \\
    MOP10 & 2     & 7     & 1     & $(0.5,2)^\intercal$ & $(0,1)^\intercal$ & 0.1   &       & $\left[\begin{smallmatrix} 1 & 0 \\ 0 & 1 \end{smallmatrix}\right]$ & 3     & 2     & 0.8   & 2     & 0   & $(0,1)^\intercal$ &       & A$^\S$, B & Nonlinear & Mixed \\
    MOP11 & 3     & 11    & 2     & $(2,2,0.5)^\intercal$ & $(0.2,0.2,0.6)^\intercal$    & 1     &       & $\left[\begin{smallmatrix} 0.33 & 0.33 & 0.33 \\ 0.33 & 0.33 & 0.33 \\ 0.33 & 0.33 & 0.33 \end{smallmatrix}\right]$ & 12     & 0     & 0.1   & 0     & 0   & NA    &       & B & Linear & Mixed \\
    MOP12 & 3     & 11    & 2     & $(0.5,0.5,0.5)^\intercal$ & $(0.33,0.33,0.33)^\intercal$ & 0.2   &       & $\left[\begin{smallmatrix} 0.6 & 0.2 & 0.2 \\ 0.2 & 0.6 & 0.2 \\ 0.2 & 0.2 & 0.6 \end{smallmatrix}\right]$ & 6     & 0     & 0.5     & 0     & 0     & NA    &       & A$^\dagger$, B & Linear & Concave \\
    MOP13 & 3     & 11    & 2     & $(2,2,2)^\intercal$ & $(0,0,1)^\intercal$    & 1     &       & $\left[\begin{smallmatrix} 0.33 & 0.33 & 0.33 \\ 0.33 & 0.33 & 0.33 \\ 0.33 & 0.33 & 0.33 \end{smallmatrix}\right]$ & 6     & 4     & 2     & 4     & 3     & $(0.33,0.33,0.33)^\intercal$ &       & A$^\ddagger$ & Nonlinear & Convex \\
    MOP14 & 3     & 11    & 2     & $(0.5,0.5,2)^\intercal$ & $(0,0,1)^\intercal$    & 1     &       & $\left[\begin{smallmatrix} 0.6 & 0.2 & 0.2 \\ 0.2 & 0.6 & 0.2 \\ 0.2 & 0.2 & 0.6 \end{smallmatrix}\right]$ & 12    & 1     & 2   & 1     & 3   & $(0.33,0.33,0.33)^\intercal$ &       & A$^\ddagger$ & Nonlinear & Mixed \\
    MOP15 & 3     & 11    & 2     & $(2,2,2)^\intercal$ & $(0.33,0.33,0.33)^\intercal$ & 0.2   &       & $\left[\begin{smallmatrix} 0.7 & 0.2 & 0.1 \\ 0.1 & 0.7 & 0.2 \\ 0.2 & 0.1 & 0.7 \end{smallmatrix}\right]$ & 6     & 1     & 2     & 1     & 3     & $(0.33,0.33,0.33)^\intercal$ &       & A$^\S$ & Nonlinear & Convex \\
    MOP16 & 3     & 11    & 2     & $(0.5,0.5,2)^\intercal$ & $(0,0,1)^\intercal$ & 0.1   &       & $\left[\begin{smallmatrix} 1 & 0 & 0 \\ 0 & 1 & 0 \\ 0 & 0 & 1 \end{smallmatrix}\right]$ & 3     & 2     & 0.8   & 2     & 0   & $(0,0,1)^\intercal$ &       & A$^\S$, B & Nonlinear & Mixed \\
    \bottomrule
    \end{tabular}%
    }
    \begin{tablenotes}
        \item[] NA: not applicable to the parameter.
        \item[] A: the $PF$ boundaries are difficult to approximate; B: the $PF$ is difficult to approximate.
        \item[] $\dagger$: the position-related bias caused by the position function; $\ddagger$: the position-related bias caused by the distance function; $\S$: the position-related bias caused by the position and distance functions.
    \end{tablenotes}
    \end{threeparttable}
    }
  \label{tab:instances}%
\end{table*}%

\subsection{Proposed Test Instances}

\begin{figure*}[ht]
\centering
\subfloat[MOP2]{
    \includegraphics[width = 0.24\linewidth]{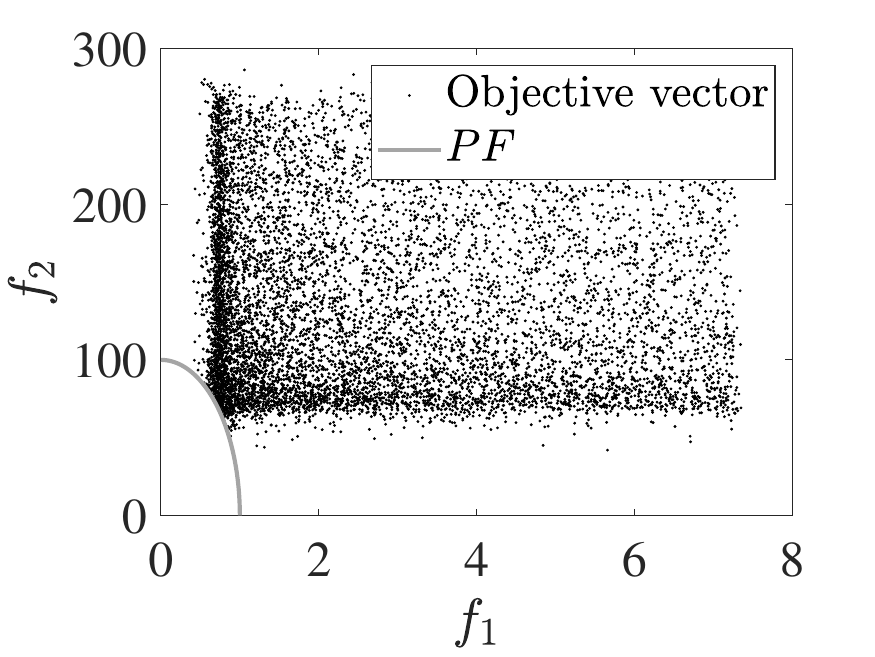}\label{fig:MOP1_sample}
}
\subfloat[MOP4]{
    \includegraphics[width = 0.24\linewidth]{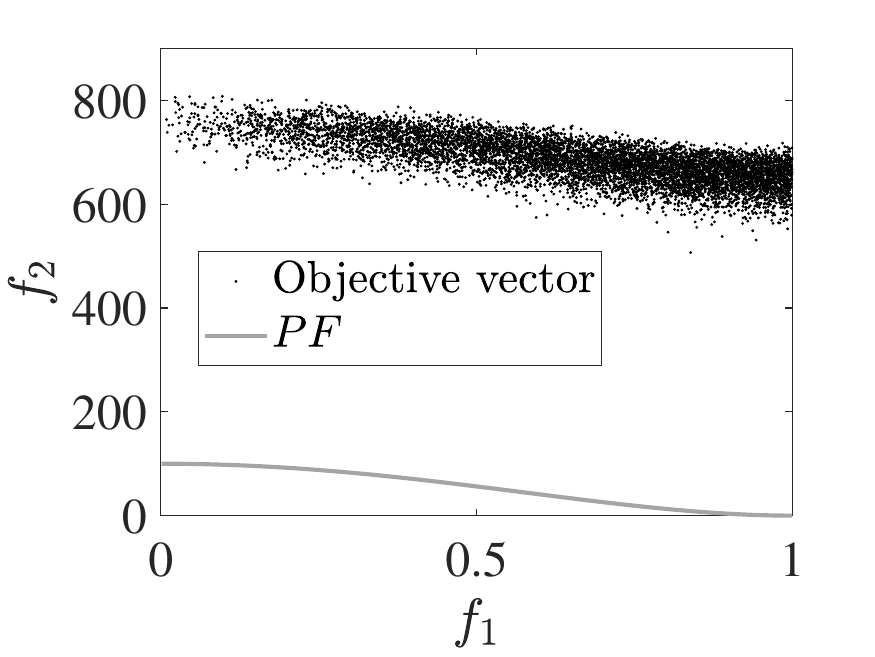}\label{fig:MOP4_sample}
}
\subfloat[MOP6]{
    \includegraphics[width = 0.24\linewidth]{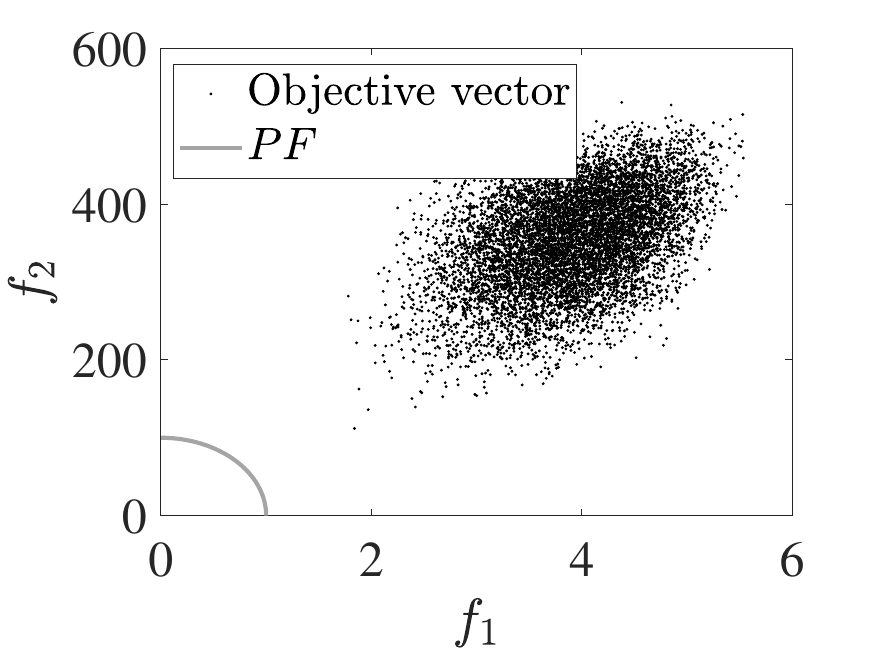}\label{fig:MOP5_sample}
}
\subfloat[MOP10]{
    \includegraphics[width = 0.24\linewidth]{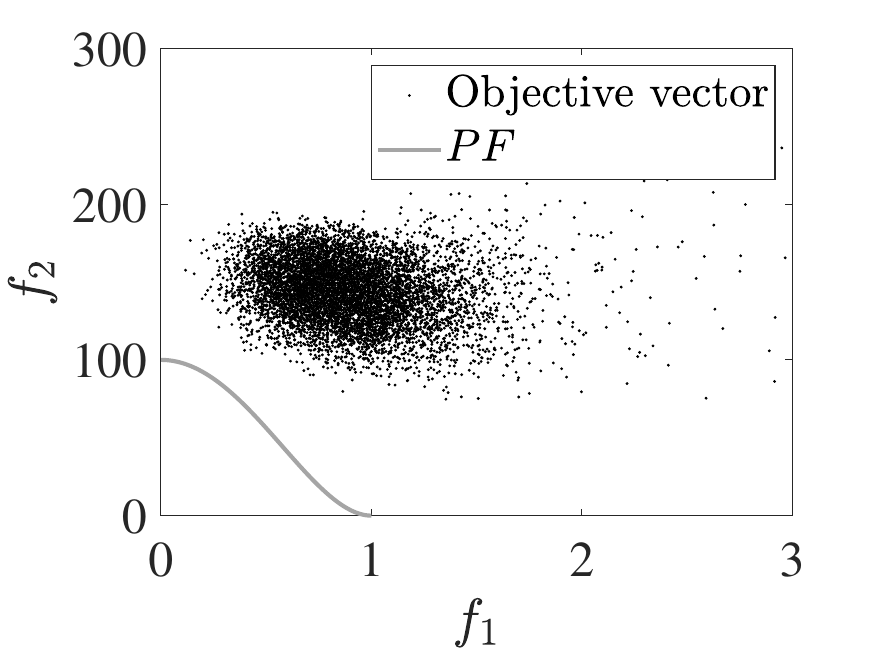}\label{fig:MOP10_sample}
}
\caption{Plots of 10000 random solutions on proposed MOPs.}
\label{fig:instance_sample}
\end{figure*}

A set of 16 new multi-objective test instances is generated and presented in Table~\ref{tab:instances}. Their corresponding parameter settings and features are also available in the table. 
%
Each instance has different ranges of the $PF$: $z_i^{ide}=0$ and $z_i^{nad}=10^{2(i-1)}$ for $i=1,\ldots,m$.
Among the 2-objective instances (\ie, MOP1--MOP10), MOP3 and MOP4 correspond to the difficulties in \figurename~\ref{fig:bias_dis}; MOP1, MOP2, and MOP7-MOP9 correspond to those in \figurename~\ref{fig:bias_pos}; MOP5, MOP6, and MOP10 correspond to those in \figurename~\ref{fig:bias_mixed}. The 3-objective instances (\ie, MOP11--MOP16) are modified from the 2-objective ones. Specifically, MOP11 combines MOP3 and MOP4; MOP12 combines MOP5 and MOP6; MOP13, MOP14, MOP15, and MOP16 corresponds to MOP7, MOP8, MOP9, and MOP10, respectively.
\figurename~\ref{fig:instance_sample} plots randomly generated solutions in the objective space, highlighting the biased features:
\begin{itemize}
    \item MOP2. The middle part of the $PF$ is obtained while the boundary part has no scatter.
    \item MOP4. Achieving the entire $PF$ is challenging. However, the feasible objective region allows the scatters to attain $z_1^{ide}$ without achieving $z_2^{ide}$.
    \item MOP6. The scatters lack both convergence and diversity.
    \item MOP10. It is similar to MOP6. But $z_2^{ide}$ is more challenging to obtain than $z_1^{ide}$ in MOP10.
\end{itemize}
In short, the test instances have various biased objective spaces, and estimating the ideal objective vector on them is more challenging than on existing test instances.

\section{Enhanced Ideal Objective Vector \texorpdfstring{\newline}{} Estimation (EIE)}\label{sec:algorithm}
The significance and challenges in estimating the ideal objective vector drive us to propose a new estimation method called EIE. EIE is algorithm-independent, and any MOEA can integrate EIE to enhance the ideal objective vector estimation.

\subsection{Algorithm Overview}
An MOEA mainly has two steps: reproduction and selection. The MOEA with EIE works as outlined in Algorithm~\ref{alg:framework}. The reproduction and selection in an MOEA are shown in lines~\ref{algl:reproduction} and~\ref{algl:selection}. EIE generates an offspring set $O_1$ before the reproduction and updates its parameters after the selection.

\begin{algorithm}[ht]
\caption{MOEA + EIE}
\label{alg:framework}
\textbf{Input}: An MOP, stopping criteria.\\
\textbf{Output}: $P$ (the population).
\begin{algorithmic}[1] 
\STATE Initialize the MOEA and EIE.
\STATE Generate an initial population $P$.

\WHILE{stopping criteria are not satisfied}\label{algl:iter_s}
    \IF{EIE does not terminate}\label{algl:EIE_term}
        \STATE\label{algl:eie_rep} Create an offspring set $O_1$ via EIE.
    \ENDIF
    \STATE\label{algl:reproduction} Obtain a mating pool and use reproduction operators to create an offspring set $O_2$.
    \STATE\label{algl:selection} Select the next generation of $P$ from $O_1\cup O_2\cup P$.
    \STATE\label{algl:upd_eie} Update the parameters of EIE.
\ENDWHILE\label{algl:iter_e}
\end{algorithmic}
\end{algorithm}

EIE assists an MOEA to emphasize the search for the ideal objective vector in line~\ref{algl:eie_rep}. It carefully utilizes limited computational resources and halts upon accomplishing the task as shown in line~\ref{algl:EIE_term}, which strikes a balance in computational resource allocation between the MOEA and EIE. EIE can use any suitable optimizer capable of perceiving the biased search space and performing efficient searches within confined areas. That is, the chosen optimizer can adaptively adjust the generated solution set to fit the biased region.
Moreover, EIE and the MOEA facilitate their searches by exchanging information. For example, EIE can generate superior solutions to guide the population~\cite{gong2023effects}, providing information about the biased region. Conversely, the population can provide diverse solutions, helping the optimizer of EIE converge faster and evade local optima~\cite{qian2013analysis,dang2024crossover}.
Last but not least, EIE aims to optimize $m$ EWS-based subproblems rather than $m$ objective functions. The details of the EWS method will be introduced in the subsequent section.
In line~\ref{algl:upd_eie}, the parameters of the EIE are updated, including the termination state and the parameters of the optimizers for EWS-based subproblems.

\subsection{Alignment of Ideal Objective Vector via Extreme Weighted Sum (EWS) Method}\label{sec:EWS}
The ideal objective vector can be obtained by optimizing objective functions separately. However, the obtained optimal solution may be weakly Pareto-optimal rather than Pareto-optimal if its objective vector is located on the boundary of the feasible objective region.
Such a solution is regarded as dominance-resistant~\cite{ikeda2001failure,wang2019scalable,liu2021solving}. Integrating it into the population could degrade the performance of the MOEA.
Our proposed EWS method is a special case of the weighted sum method~\cite{miettinen1998nonlinear}, and it defines the subproblem for the $i$-th objective as
\begin{equation}\label{eqn:subp}
    \begin{array}{ll}
        \text{min.} &\!\! g_i^{ews}(\mathbf{x}|\alpha_i) \!=\! (1 \!-\! \alpha_i)f_i(\mathbf{x}) \!+\! \frac{\alpha_i}{m-1}\sum\limits_{\substack{j=1 \\ \wedge j\neq i}}^m f_j(\mathbf{x}),
        \\
        \text{s.t.} &\!\! \mathbf{x}\in\Omega,
    \end{array}
\end{equation}
where $\alpha_i>0$ is a small value. The non-negative weight vector is controlled by $\alpha_i$: the $i$-th element is $(1-\alpha_i)$ and the rest are $\frac{\alpha_i}{m-1}$. 
If $\alpha_i$ is sufficiently small, the EWS method can define $m$ single-objective optimization subproblems to find the Pareto-optimal objective vectors aligned with the ideal objective vector. The upper bound of the estimation error is given by the following theorem. The proof is provided in Appendix~\ref{sec:proof}.

\begin{theorem}\label{the:error_bound}
Let $\mathbf{z}^i$ be any optimal objective vector of the $i$-th EWS-based subproblem and $\beta$ be a constant. Suppose that $\alpha_i\leq 0.5$ and $z_j^{nad}=z_j^{ide}+\beta$ for $j=1,\ldots,m$. Then
\begin{equation}\label{eqn:ine_error_bound}
    z_i^i-z_i^{ide} \leq \frac{\alpha_i\beta}{1-\alpha_i}.
\end{equation}
\end{theorem}

Theorem~\ref{the:error_bound} has two assumptions. The first one is the maximum value that $\alpha_i$ can take. The second one indicates the objective functions have the same range. In other words, the objective space is normalized. To better meet the second assumption, we normalize the objective space using the minimal objective function values and the maximum objective function values extracted from the current iteration's population.
$z_i^i-z_i^{ide}$ in Eq.~\eqref{eqn:ine_error_bound} represents a maximum error of the estimated optimal objective function value. In practical applications, setting a tolerance level for the error is more intuitive than configuring $\alpha_i$. Therefore, we let $\epsilon_i=\left(z_i^i-z_i^{ide}\right)/\left(z_i^{nad}-z_i^{ide}\right)$ be a normalized tolerance determined by the users, and then $\alpha_i$ can be computed by $\alpha_i=\frac{\epsilon_i}{\epsilon_i+1}$.

\subsection{Implementing EIE for Continuous MOPs}\label{sec:impEIE}
To evaluate EIE on the proposed MOP1-MOP16, we adopt a continuous problem optimizer called PSA-CMA-ES~\cite{nishida2018psa}.
PSA-CMA-ES generally contains two parts: covariance matrix adaptation evolution strategy (CMA-ES) and the population size adaptation (PSA) strategy. We choose it for three reasons:
\begin{itemize}
    \item CMA-ES can well exploit a small region since it adaptively adjusts the spread and shape of the generated solution set.
    \item The PSA strategy can adaptively adjust the number of generated solutions (\ie, the population size), conforming to the principle of careful computational resource usage.
    \item Each iteration of PSA-CMA-ES exhibits linear time complexity with respect to its population size. Consequently, integrating $m$ PSA-CMA-ES procedures into the MOEA does not increase the overall computational costs.
\end{itemize}
At each iteration, $O_1$ are generated by $m$ independent PSA-CMA-ES procedures (Algorithm~\ref{alg:framework} line~\ref{algl:eie_rep}). For each EWS-based subproblem, PSA-CMA-ES generates the candidate solutions as $\mathbf{x}^i \sim N(\bar{\mathbf{x}},\sigma^2 C)$ for $i=1,\ldots,\lambda$, where $N(\bar{\mathbf{x}},\sigma^2 C)$ is the normal distribution with mean $\bar{\mathbf{x}}$ and covariance matrix $\sigma^2 C$; and $\lambda$ indicates the population size of PSA-CMA-ES. The PSA strategy requires a range of alternative population sizes. Let $\lambda_{def}$ be the default population size of CMA-ES~\cite{hansen2016cma}. We set $\lambda\in[\lambda_{def},8\lambda_{def}]$, where the lower bound is the default value in~\cite{nishida2018psa} and the upper one is finite as suggested in~\cite{nishida2018benchmarking}. Other parameter settings are the same as those in~\cite{hansen2016cma}. The warm starting strategy~\cite{nomura2021warm} is employed to initialize $\bar{\mathbf{x}}$, $\sigma$, and $C$, whose input data is the population of the MOEA.

\begin{algorithm}[ht]
\caption{Update EIE (for the $i$-th EWS-based subproblem)}
\label{alg:EIE}
\textbf{Input}: stopping criteria, $\bar{\mathbf{x}}$, $\sigma$, $C$, $\lambda$, $P$, $O_{es}$ (the set of solutions generated by the $i$-th PSA-CMA-ES procedure), $O_1\cup O_2$ (the set of all new solutions).\\
\textbf{Output}: $\bar{\mathbf{x}}$, $\sigma$, $C$, $\lambda$.

\begin{algorithmic}[1] 
\IF{$\lambda\leq\lambda_{def}$}\label{algl:inj_s}
    \STATE $O_{in}\leftarrow O_{es}\cup O_1\cup O_2$.
\ELSE
    \STATE $O_{in}\leftarrow O_{es}$.
\ENDIF\label{algl:inj_e}
\STATE\label{algl:comp_fit} Compute the subproblem function values of $O_{in}$.
\STATE\label{algl:sel_lam} Select $\lambda$ best solutions from $O_{in}$.
\STATE Recompute parameters depending on $\lambda$.
\STATE Update $\bar{\mathbf{x}},\sigma,C$ and other variables of CMA-ES.
\STATE\label{algl:upd_psa} Update $\lambda$ and other variables of the PSA strategy.
\IF{Exceptional stopping criteria are satisfied}\label{algl:stop_cri_s}
    \STATE Warm-start the $i$-th PSA-CMA-ES procedure according to $P$~\cite{nomura2021warm}.
\ELSIF{Conventional stopping criteria are satisfied}
    \STATE Stop the $i$-th PSA-CMA-ES procedure.
\ENDIF\label{algl:stop_cri_e}
\end{algorithmic}
\end{algorithm}

Algorithm~\ref{alg:EIE} outlines the procedure of updating EIE for each EWS-based subproblem, representing the implementation of Algorithm~\ref{alg:framework} line~\ref{algl:upd_eie}. According to~\cite{nishida2018psa}, a small population size indicates stable updates of $\bar{\mathbf{x}}$, $\sigma$, and $C$. In this case, the search distribution might be concentrated at the basin of a local minimum. We inject 
good new solutions to cope with this issue (lines~\ref{algl:inj_s}-\ref{algl:inj_e}). Thereafter, the variables of PSA-CMA-ES are updated (lines~\ref{algl:sel_lam}-\ref{algl:upd_psa}). Note that the solutions in $O_1\cup O_2\setminus O_{es}$ are injected solutions for the $i$-th PSA-CMA-ES procedure. The injected solutions should obey the injection rule~\cite{hansen2011injecting,zapotecas2015injecting}.
To save computational resources and prevent stagnation, the procedure of each subproblem finally checks the following four criteria taken from~\cite{hansen2016cma}:
\begin{itemize}
    \item \texttt{NoEffectAxis}. 
    Let $d_i$ be an eigenvalue of $C$ and $\mathbf{b}_i$ be the corresponding eigenvector. Stop if $\forall i=1,\ldots,n, \bar{\mathbf{x}}=\bar{\mathbf{x}}+0.1\sigma\sqrt{d_i}\mathbf{b}_i$.
    \item \texttt{NoEffectCoord}. Stop if $\forall i=1,\ldots,n, \bar{x}_i=\bar{x}_i+0.2\sigma\sqrt{C_{ii}}$.
    \item \texttt{TolFun$\wedge$TolX}. 
    Stop if 1) the range of the best objective function values of the last $10+\left\lceil30\frac{n}{\lambda}\right\rceil$ generations and all function values of the recent generation is below $10^{-3}$
    and 2) $\forall i=1,\ldots,n, (\sigma\sqrt{C_{ii}}<10^{-6}\sigma^\prime) \wedge (\sigma (p_c)_i<10^{-6}\sigma^\prime)$.
    \item \texttt{TolXUp}. Stop if $\exists i=1,\ldots,n, \sigma\sqrt{d_i}>10^4\sigma^\prime\sqrt{d_i^\prime}$ where $\sigma^\prime$ and $d_i^\prime$ is the initial values of $\sigma$ and $d_i$.
\end{itemize}
When one of the stopping criteria is met, the corresponding PSA-CMA-ES procedure terminates or restarts (lines~\ref{algl:stop_cri_s}-\ref{algl:stop_cri_e}). We consider \texttt{NoEffectAxis}, \texttt{NoEffectCoord}, and \texttt{TolX$\wedge$TolFun} as the conventional stopping criteria and \texttt{TolXUp} as the exceptional stopping criterion.

\section{Experiments}\label{sec:experiments}
In this section, we show the effectiveness of EIE. More experiments are conducted in Appendix~\ref{sec:discussions} to demonstrate:
\begin{itemize}
    \item the superiority of EIE over other ideal objective vector estimation methods;
    \item the positive impact of the EWS method on MOEAs employing EIE;
    \item the competitive results of MOEAs with EIE on variants of MOP11--MOP16 with inverted $PF$s and the \texttt{bbob-biobj} test suite~\cite{brockhoff2022using}.
\end{itemize}
The source code for all experiments is available at \url{https://github.com/EricZheng1024/EIE}.

\begin{table*}[t]
\footnotesize
  \centering
  \caption{Comparisons of $\operatorname{E}$ metric values between the MOEA and the MOEA with EIE.}
    \renewcommand{\arraystretch}{1.2}
    \scalebox{1}{
	\setlength{\tabcolsep}{0.8mm}{
	\begin{tabular}{cc||ccc|ccc|ccc}
	\toprule
    Instance & $\operatorname{E}$    & PMEA  & +EIE  & $\Delta$ & gMOEA/D-GGR & +EIE  & $\Delta$ & HVCTR & +EIE  & $\Delta$ \\
    \midrule
    \multirow{2}[1]{*}{MOP1} & mean  & 0.18089(2)- & \cellcolor[rgb]{ .651,  .651,  .651}0.0091884(1) & -0.1717 & 0.054191(2)- & \cellcolor[rgb]{ .651,  .651,  .651}0.0015222(1) & -0.052669 & 0.15868(2)- & \cellcolor[rgb]{ .651,  .651,  .651}0.010025(1) & -0.14865 \\
          & std.  & 0.065871 & 0.022818 &       & 0.010661 & 0.0048056 &       & 0.021744 & 0.02137 &  \\
    \multirow{2}[0]{*}{MOP2} & mean  & 0.22895(2)- & \cellcolor[rgb]{ .651,  .651,  .651}0.00081583(1) & -0.22813 & 0.061795(2)- & \cellcolor[rgb]{ .651,  .651,  .651}0.00027052(1) & -0.061525 & 0.23259(2)- & \cellcolor[rgb]{ .651,  .651,  .651}0.00058399(1) & -0.232 \\
          & std.  & 0.049788 & 0.0010416 &       & 0.011325 & 0.00060602 &       & 0.026187 & 0.00071678 &  \\
    \multirow{2}[0]{*}{MOP3} & mean  & 0.044149(2)- & \cellcolor[rgb]{ .651,  .651,  .651}0.0002752(1) & -0.043874 & 0.026881(2)- & \cellcolor[rgb]{ .651,  .651,  .651}0.0022258(1) & -0.024655 & \cellcolor[rgb]{ .651,  .651,  .651}3.2968e-05(1)+ & 0.00010581(2) & 7.2841e-05 \\
          & std.  & 0.063828 & 0.00048072 &       & 0.10336 & 0.0021959 &       & 2.5449e-05 & 8.1196e-05 &  \\
    \multirow{2}[0]{*}{MOP4} & mean  & 0.23985(2)- & \cellcolor[rgb]{ .651,  .651,  .651}0.0066605(1) & -0.23319 & 0.037047(2)- & \cellcolor[rgb]{ .651,  .651,  .651}0.0022646(1) & -0.034782 & 0.010127(2)- & \cellcolor[rgb]{ .651,  .651,  .651}0.0012552(1) & -0.0088721 \\
          & std.  & 0.039378 & 0.0054399 &       & 0.034039 & 0.0017182 &       & 0.039914 & 0.0019377 &  \\
    \multirow{2}[0]{*}{MOP5} & mean  & \cellcolor[rgb]{ .651,  .651,  .651}0.085634(1)= & 0.089466(2) & 0.0038316 & 0.019387(2)- & \cellcolor[rgb]{ .651,  .651,  .651}0.0032029(1) & -0.016184 & 0.024967(2)- & \cellcolor[rgb]{ .651,  .651,  .651}0.0033346(1) & -0.021633 \\
          & std.  & 0.01483 & 0.015991 &       & 0.0027861 & 0.0036188 &       & 0.0038873 & 0.0031756 &  \\
    \multirow{2}[0]{*}{MOP6} & mean  & 0.26944(2)- & \cellcolor[rgb]{ .651,  .651,  .651}0.037092(1) & -0.23235 & 0.080559(2)- & \cellcolor[rgb]{ .651,  .651,  .651}0.017454(1) & -0.063105 & 0.22356(2)- & \cellcolor[rgb]{ .651,  .651,  .651}0.038634(1) & -0.18493 \\
          & std.  & 0.021423 & 0.053028 &       & 0.010875 & 0.023984 &       & 0.029224 & 0.042473 &  \\
    \multirow{2}[0]{*}{MOP7} & mean  & 0.095539(2)- & \cellcolor[rgb]{ .651,  .651,  .651}0.074193(1) & -0.021346 & 0.040644(2)- & \cellcolor[rgb]{ .651,  .651,  .651}0.002601(1) & -0.038043 & 0.053428(2)- & \cellcolor[rgb]{ .651,  .651,  .651}0.0028248(1) & -0.050603 \\
          & std.  & 0.0057233 & 0.0063049 &       & 0.004193 & 0.00020939 &       & 0.0050568 & 0.00012199 &  \\
    \multirow{2}[0]{*}{MOP8} & mean  & 0.042751(2)- & \cellcolor[rgb]{ .651,  .651,  .651}0.00022782(1) & -0.042524 & 0.016882(2)- & \cellcolor[rgb]{ .651,  .651,  .651}3.464e-05(1) & -0.016847 & 0.15582(2)- & \cellcolor[rgb]{ .651,  .651,  .651}0.00020177(1) & -0.15562 \\
          & std.  & 0.067985 & 0.00062329 &       & 0.014622 & 7.3183e-06 &       & 0.074577 & 0.00058243 &  \\
    \multirow{2}[0]{*}{MOP9} & mean  & 0.15186(2)- & \cellcolor[rgb]{ .651,  .651,  .651}0.038038(1) & -0.11382 & 0.082225(2)- & \cellcolor[rgb]{ .651,  .651,  .651}0.0043003(1) & -0.077925 & 0.12753(2)- & \cellcolor[rgb]{ .651,  .651,  .651}0.0040902(1) & -0.12344 \\
          & std.  & 0.0096829 & 0.047876 &       & 0.0065422 & 0.0069067 &       & 0.0074688 & 0.00028397 &  \\
    \multirow{2}[0]{*}{MOP10} & mean  & 0.27629(2)- & \cellcolor[rgb]{ .651,  .651,  .651}0.046936(1) & -0.22935 & 0.16661(2)- & \cellcolor[rgb]{ .651,  .651,  .651}0.049306(1) & -0.11731 & 0.26893(2)- & \cellcolor[rgb]{ .651,  .651,  .651}0.064343(1) & -0.20459 \\
          & std.  & 0.040841 & 0.0064742 &       & 0.036631 & 0.0077489 &       & 0.052671 & 0.0077825 &  \\
    \multirow{2}[0]{*}{MOP11} & mean  & 2.6722(2)- & \cellcolor[rgb]{ .651,  .651,  .651}0.011802(1) & -2.6604 & 1.5813(2)- & \cellcolor[rgb]{ .651,  .651,  .651}0.0049623(1) & -1.5764 & 0.13548(2)- & \cellcolor[rgb]{ .651,  .651,  .651}0.002938(1) & -0.13255 \\
          & std.  & 0.23174 & 0.023672 &       & 0.43108 & 0.0024799 &       & 0.11036 & 0.0034786 &  \\
    \multirow{2}[0]{*}{MOP12} & mean  & 0.26396(2)- & \cellcolor[rgb]{ .651,  .651,  .651}0.019599(1) & -0.24436 & 0.13363(2)- & \cellcolor[rgb]{ .651,  .651,  .651}0.0084918(1) & -0.12514 & 0.12493(2)- & \cellcolor[rgb]{ .651,  .651,  .651}0.0084181(1) & -0.11652 \\
          & std.  & 0.029921 & 0.046111 &       & 0.018476 & 0.015861 &       & 0.020548 & 0.02085 &  \\
    \multirow{2}[0]{*}{MOP13} & mean  & 0.024109(2)- & \cellcolor[rgb]{ .651,  .651,  .651}0.0010698(1) & -0.023039 & 0.0088682(2)- & \cellcolor[rgb]{ .651,  .651,  .651}0.00015638(1) & -0.0087118 & 0.011704(2)- & \cellcolor[rgb]{ .651,  .651,  .651}0.00055261(1) & -0.011151 \\
          & std.  & 0.002398 & 0.0011775 &       & 0.0012902 & 5.5381e-05 &       & 0.0010953 & 0.00010071 &  \\
    \multirow{2}[0]{*}{MOP14} & mean  & 0.10888(2)- & \cellcolor[rgb]{ .651,  .651,  .651}8.1619e-06(1) & -0.10887 & 0.062963(2)- & \cellcolor[rgb]{ .651,  .651,  .651}1.0645e-05(1) & -0.062953 & 0.029319(2)- & \cellcolor[rgb]{ .651,  .651,  .651}4.8064e-06(1) & -0.029314 \\
          & std.  & 0.036245 & 1.1272e-05 &       & 0.026478 & 1.0792e-05 &       & 0.011521 & 4.6016e-06 &  \\
    \multirow{2}[0]{*}{MOP15} & mean  & 0.06199(2)- & \cellcolor[rgb]{ .651,  .651,  .651}0.0038598(1) & -0.05813 & 0.029477(2)- & \cellcolor[rgb]{ .651,  .651,  .651}0.00062477(1) & -0.028852 & 0.037928(2)- & \cellcolor[rgb]{ .651,  .651,  .651}0.0036141(1) & -0.034314 \\
          & std.  & 0.0039023 & 0.0074118 &       & 0.0039464 & 0.00023242 &       & 0.0022613 & 0.00058919 &  \\
    \multirow{2}[1]{*}{MOP16} & mean  & 0.18366(2)- & \cellcolor[rgb]{ .651,  .651,  .651}0.037962(1) & -0.1457 & 0.13099(2)- & \cellcolor[rgb]{ .651,  .651,  .651}0.027427(1) & -0.10356 & 0.079291(2)- & \cellcolor[rgb]{ .651,  .651,  .651}0.035449(1) & -0.043841 \\
          & std.  & 0.027365 & 0.0044556 &       & 0.029654 & 0.0039304 &       & 0.015319 & 0.0036123 &  \\
    \midrule
    \multicolumn{2}{c}{Total +/=/-} & 0/1/15 & \textbackslash{} &       & 0/0/16 & \textbackslash{} &       & 1/0/15 & \textbackslash{} &  \\
    \multicolumn{2}{c}{Average rank} & 1.9375(2) & 1.0625(1) &       & 2(2)  & 1(1)  &       & 1.9375(2) & 1.0625(1) &  \\
    \bottomrule
    \end{tabular}%
    }
    }
  \label{tab:baseline_e}%
\end{table*}%
\begin{table*}[t]
\footnotesize
  \centering
  \caption{Comparisons of $\operatorname{HV}$ metric values between the MOEA and the MOEA with EIE.}
    \renewcommand{\arraystretch}{1.2}
    \scalebox{1}{
	\setlength{\tabcolsep}{0.8mm}{
	\begin{tabular}{cc||ccc|ccc|ccc}
	\toprule
    Instance & $\operatorname{HV}$    & PMEA  & +EIE  & $\Delta$ & gMOEA/D-GGR & +EIE  & $\Delta$ & HVCTR & +EIE  & $\Delta$ \\
    \midrule
    \multirow{2}[1]{*}{MOP1} & mean  & 0.662847(2)- & \cellcolor[rgb]{ .651,  .651,  .651}0.70003(1) & 0.0371832 & 0.694146(2)- & \cellcolor[rgb]{ .651,  .651,  .651}0.703357(1) & 0.00921148 & 0.674416(2)- & \cellcolor[rgb]{ .651,  .651,  .651}0.702499(1) & 0.0280833 \\
          & std.  & 0.010497 & 0.00499976 &       & 0.00143603 & 0.00101947 &       & 0.005282 & 0.00311904 &  \\
    \multirow{2}[0]{*}{MOP2} & mean  & 0.38963(2)- & \cellcolor[rgb]{ .651,  .651,  .651}0.419003(1) & 0.0293733 & 0.410776(2)- & \cellcolor[rgb]{ .651,  .651,  .651}0.419059(1) & 0.00828313 & 0.387927(2)- & \cellcolor[rgb]{ .651,  .651,  .651}0.420289(1) & 0.032362 \\
          & std.  & 0.0102756 & 0.000967767 &       & 0.00142906 & 0.000718643 &       & 0.00557535 & 0.000924814 &  \\
    \multirow{2}[0]{*}{MOP3} & mean  & 0.64718(2)- & \cellcolor[rgb]{ .651,  .651,  .651}0.704328(1) & 0.0571484 & 0.670473(2)- & \cellcolor[rgb]{ .651,  .651,  .651}0.698207(1) & 0.0277337 & \cellcolor[rgb]{ .651,  .651,  .651}0.704931(1)+ & 0.704918(2) & -1.2213e-05 \\
          & std.  & 0.0824221 & 0.000691952 &       & 0.114741 & 0.00242507 &       & 6.23006e-06 & 1.04308e-05 &  \\
    \multirow{2}[0]{*}{MOP4} & mean  & 0.434233(2)- & \cellcolor[rgb]{ .651,  .651,  .651}0.666302(1) & 0.232069 & 0.632297(2)- & \cellcolor[rgb]{ .651,  .651,  .651}0.66452(1) & 0.032223 & 0.664067(2)= & \cellcolor[rgb]{ .651,  .651,  .651}0.671117(1) & 0.00704989 \\
          & std.  & 0.0301284 & 0.00935514 &       & 0.0294533 & 0.00419556 &       & 0.0405197 & 0.00102382 &  \\
    \multirow{2}[0]{*}{MOP5} & mean  & 0.993616(2)= & \cellcolor[rgb]{ .651,  .651,  .651}0.995271(1) & 0.0016544 & 1.0291(2)- & \cellcolor[rgb]{ .651,  .651,  .651}1.03585(1) & 0.00674584 & 1.02895(2)- & \cellcolor[rgb]{ .651,  .651,  .651}1.03697(1) & 0.00802155 \\
          & std.  & 0.00905147 & 0.00880108 &       & 0.00130447 & 0.00218626 &       & 0.00159396 & 0.00138483 &  \\
    \multirow{2}[0]{*}{MOP6} & mean  & 0.381333(2)- & \cellcolor[rgb]{ .651,  .651,  .651}0.411897(1) & 0.0305635 & 0.389398(2)- & \cellcolor[rgb]{ .651,  .651,  .651}0.400834(1) & 0.0114356 & 0.389591(2)- & \cellcolor[rgb]{ .651,  .651,  .651}0.415089(1) & 0.0254987 \\
          & std.  & 0.00323642 & 0.00514564 &       & 0.00295437 & 0.00398384 &       & 0.00411072 & 0.00425518 &  \\
    \multirow{2}[0]{*}{MOP7} & mean  & 0.987983(2)- & \cellcolor[rgb]{ .651,  .651,  .651}1.00484(1) & 0.0168588 & 1.02164(2)- & \cellcolor[rgb]{ .651,  .651,  .651}1.03195(1) & 0.0103061 & 1.01582(2)- & \cellcolor[rgb]{ .651,  .651,  .651}1.03945(1) & 0.0236262 \\
          & std.  & 0.00384932 & 0.00425867 &       & 0.0020964 & 0.00170743 &       & 0.00308745 & 3.88267e-05 &  \\
    \multirow{2}[0]{*}{MOP8} & mean  & 0.646003(2)- & \cellcolor[rgb]{ .651,  .651,  .651}0.665932(1) & 0.0199293 & 0.658923(2)- & \cellcolor[rgb]{ .651,  .651,  .651}0.660511(1) & 0.00158834 & 0.630566(2)- & \cellcolor[rgb]{ .651,  .651,  .651}0.667899(1) & 0.0373327 \\
          & std.  & 0.0153143 & 0.00709498 &       & 0.00181245 & 0.00117616 &       & 0.0172147 & 0.00681087 &  \\
    \multirow{2}[0]{*}{MOP9} & mean  & 0.938312(2)- & \cellcolor[rgb]{ .651,  .651,  .651}1.0153(1) & 0.0769918 & 0.993875(2)- & \cellcolor[rgb]{ .651,  .651,  .651}1.02712(1) & 0.033243 & 0.960201(2)- & \cellcolor[rgb]{ .651,  .651,  .651}1.03876(1) & 0.0785564 \\
          & std.  & 0.00870241 & 0.0309005 &       & 0.00454164 & 0.0034667 &       & 0.00630759 & 7.86262e-05 &  \\
    \multirow{2}[0]{*}{MOP10} & mean  & 0.5738(2)- & \cellcolor[rgb]{ .651,  .651,  .651}0.633353(1) & 0.0595533 & 0.616293(2)- & \cellcolor[rgb]{ .651,  .651,  .651}0.640441(1) & 0.0241484 & 0.583795(2)- & \cellcolor[rgb]{ .651,  .651,  .651}0.642445(1) & 0.0586496 \\
          & std.  & 0.0126322 & 0.00674937 &       & 0.00959871 & 0.00247657 &       & 0.0173097 & 0.00527604 &  \\
    \multirow{2}[0]{*}{MOP11} & mean  & 0(2)- & \cellcolor[rgb]{ .651,  .651,  .651}1.11203(1) & 1.11203 & 0.00927957(2)- & \cellcolor[rgb]{ .651,  .651,  .651}1.2288(1) & 1.21952 & 0.970397(2)- & \cellcolor[rgb]{ .651,  .651,  .651}1.24029(1) & 0.269893 \\
          & std.  & 0     & 0.14117 &       & 0.018189 & 0.00608648 &       & 0.198769 & 0.0210573 &  \\
    \multirow{2}[0]{*}{MOP12} & mean  & 0.535195(2)- & \cellcolor[rgb]{ .651,  .651,  .651}0.676807(1) & 0.141612 & 0.582742(2)- & \cellcolor[rgb]{ .651,  .651,  .651}0.667576(1) & 0.0848338 & 0.634423(2)- & \cellcolor[rgb]{ .651,  .651,  .651}0.743591(1) & 0.109168 \\
          & std.  & 0.0152833 & 0.0181651 &       & 0.0113595 & 0.0336248 &       & 0.0175199 & 0.0140262 &  \\
    \multirow{2}[0]{*}{MOP13} & mean  & \cellcolor[rgb]{ .651,  .651,  .651}1.23026(1)+ & 1.21589(2) & -0.0143707 & 1.29636(2)- & \cellcolor[rgb]{ .651,  .651,  .651}1.3121(1) & 0.0157404 & 1.29504(2)- & \cellcolor[rgb]{ .651,  .651,  .651}1.31467(1) & 0.0196314 \\
          & std.  & 0.00316603 & 0.00341321 &       & 0.00162777 & 0.000843473 &       & 0.0018436 & 0.000158645 &  \\
    \multirow{2}[0]{*}{MOP14} & mean  & 0.964158(2)- & \cellcolor[rgb]{ .651,  .651,  .651}1.00933(1) & 0.0451733 & 0.99019(2)- & \cellcolor[rgb]{ .651,  .651,  .651}1.01344(1) & 0.0232551 & 1.02248(2)- & \cellcolor[rgb]{ .651,  .651,  .651}1.03323(1) & 0.0107547 \\
          & std.  & 0.0187333 & 0.00482035 &       & 0.00948318 & 0.00311727 &       & 0.00496676 & 0.00216258 &  \\
    \multirow{2}[0]{*}{MOP15} & mean  & 1.1936(2)- & \cellcolor[rgb]{ .651,  .651,  .651}1.27232(1) & 0.0787266 & 1.25915(2)- & \cellcolor[rgb]{ .651,  .651,  .651}1.30693(1) & 0.0477822 & 1.2458(2)- & \cellcolor[rgb]{ .651,  .651,  .651}1.3061(1) & 0.0603069 \\
          & std.  & 0.00671164 & 0.00560318 &       & 0.00654064 & 0.0028307 &       & 0.00413541 & 0.00110377 &  \\
    \multirow{2}[1]{*}{MOP16} & mean  & 0.882233(2)- & \cellcolor[rgb]{ .651,  .651,  .651}0.925115(1) & 0.0428828 & 0.896244(2)- & \cellcolor[rgb]{ .651,  .651,  .651}0.924206(1) & 0.0279612 & 0.932359(2)- & \cellcolor[rgb]{ .651,  .651,  .651}0.960696(1) & 0.0283369 \\
          & std.  & 0.00884846 & 0.00758356 &       & 0.00825143 & 0.00990376 &       & 0.00769505 & 0.00682911 &  \\
    \midrule
    \multicolumn{2}{c||}{Total +/=/-} & 1/1/14 & \textbackslash{} &       & 0/0/16 & \textbackslash{} &       & 1/1/14 & \textbackslash{} &  \\
    \multicolumn{2}{c||}{Average rank} & 1.9375(2) & 1.0625(1) &       & 2(2)  & 1(1)  &       & 1.9375(2) & 1.0625(1) &  \\
    \bottomrule
    \end{tabular}%
    }
    }
  \label{tab:baseline_hv}%
\end{table*}%

\subsection{Experimental Setup}\label{sec:exp_set}
\subsubsection{General Settings}
The population size in our experiments is set to 100 for 2-objective instances and 210 for 3-objective instances. The maximum number of function evaluations is 200,000 for 2-objective instances and 400,000 for 3-objective instances. Each algorithm is executed 30 times on each instance. All experiments are implemented on the PlatEMO platform~\cite{tian2017platemo}.

In the result tables, the mean and standard deviation values of the metrics are recorded. The rank of each algorithm on each instance is provided after the mean of the metric value. The Wilcoxon rank-sum test with a 0.05 significance level is employed to statistically analyze the algorithms' performance on each instance. The symbols ``+'', ``='', and ``-'' indicate that the performance of the corresponding algorithm is statistically better than, comparable to, or worse than that of the rightmost algorithm. The best mean metric values are also emphasized. Besides, we present the metric value gap between the MOEA with EIE and the original one (denoted as $\Delta$).

\subsubsection{Algorithm Settings}
In EIE, $\epsilon_i$ is set to $5\%$ for $i=1,\ldots,m$. Three recently published MOEAs with representative frameworks are selected as baselines, and their parameters are set according to the corresponding references.
\begin{itemize}
    \item PMEA~\cite{liu2021solving} (dominance-based). The algorithm can eliminate dominance-resistant solutions and utilize metrics to strike a balance between convergence and diversity. The parameter $r$ is set to $1.5$.
    \item gMOEA/D-GGR~\cite{zheng2023generalized} (decomposition-based). MOEA/D-GGR is an improved version of MOEA/D-GR~\cite{wang2016adaptive} that enables the use of the $L_p$ scalarization. gMOEA/D-GGR is the generational version of MOEA/D-GGR, having better robustness. The parameters are set as follows: $p=1$, $T_m=0.1N$, and $T_r=1$.
    \item HVCTR~\cite{pang2024hypervolume} (indicator-based). It is an improved version of SMS-EMOA~\cite{beume2007sms} that simultaneously uses two reference points. Two reference points are set to $(1,\ldots,1)^\intercal$ and $(100,\ldots,100)^\intercal$ respectively.
\end{itemize}
PMEA and HVCTR have their own normalization scheme, while gMOEA/D-GGR adopts the same normalization method as EIE.
Algorithms adopt the differential evolution (DE) operator and the polynomial mutation (PM) operator, as previously described and utilized in~\cite{li2009multiobjective}. For the DE operator, we set $F = 0.5$ and $CR = 0.9$. For the PM operator, the distribution index is assigned a value of 50, and the mutation probability is established at $1/n$.

\subsubsection{Performance Metrics}
The error of the estimated ideal objective vector is computed by
\begin{equation}
    \operatorname{E}\left(\mathbf{z}^e | \mathbf{z}^{ide}, \mathbf{z}^{nad}\right) = \sqrt{\sum_{i=1}^m \left(\frac{z_i^e-z_i^{ide}}{z_i^{nad}-z_i^{ide}}\right)},
\end{equation}
where $\mathbf{z}^e$ is the estimated ideal objective vector. Let $P$ be the final population. $\mathbf{z}^e$ is extracted by $z_i^{e}=\min_{\mathbf{x} \in P} f_i\left(\mathbf{x}\right) \text{ for } i=1,\ldots,m$.
Additionally, the hypervolume ($\operatorname{HV}$)~\cite{zitzler1999multiobjective} is employed to assess the overall quality of $P$. The $\operatorname{HV}$ metric is formulated as
\begin{equation}
    \operatorname{HV}\!\left(P | \mathbf{z}^r\right) \!=\! \operatorname{Leb} \! \left(\bigcup_{\mathbf{x} \in P}\left[f_1(\mathbf{x}), z_1^r\right] \! \times \! \cdots \! \times \! \left[f_m(\mathbf{x}), z_m^r\right]\!\!\right),
\end{equation}
where $\operatorname{Leb}(\cdot)$ denotes the Lebesgue measure and $\mathbf{z}^r=\left(z_1^r, z_2^r, \ldots, z_m^r\right)^\intercal$ is a reference point dominated by any objective vector on the $PF$. We normalize $\{f_i(\mathbf{x})|\mathbf{x}\in P\}$ for $i=1,\ldots,m$ based on the $PF$. Then $z_i^h$ is set to 1.1 for $i = 1,\ldots,m$. A higher value of the $\operatorname{HV}$ metric signifies superior performance on approximating the $PF$.

\begin{figure*}[ht]
\centering
\subfloat[MOP1]{\includegraphics[width=0.245\linewidth]{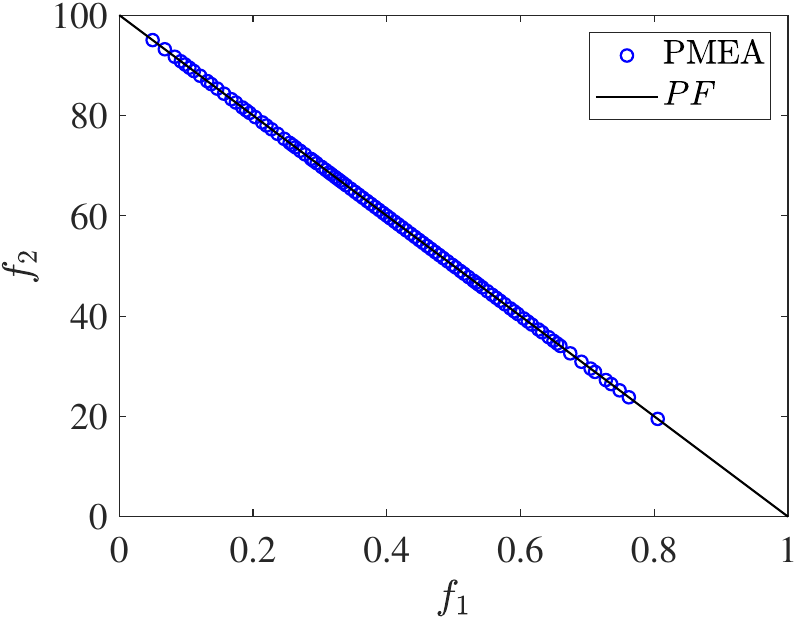}}
\subfloat[MOP1]{\includegraphics[width=0.245\linewidth]{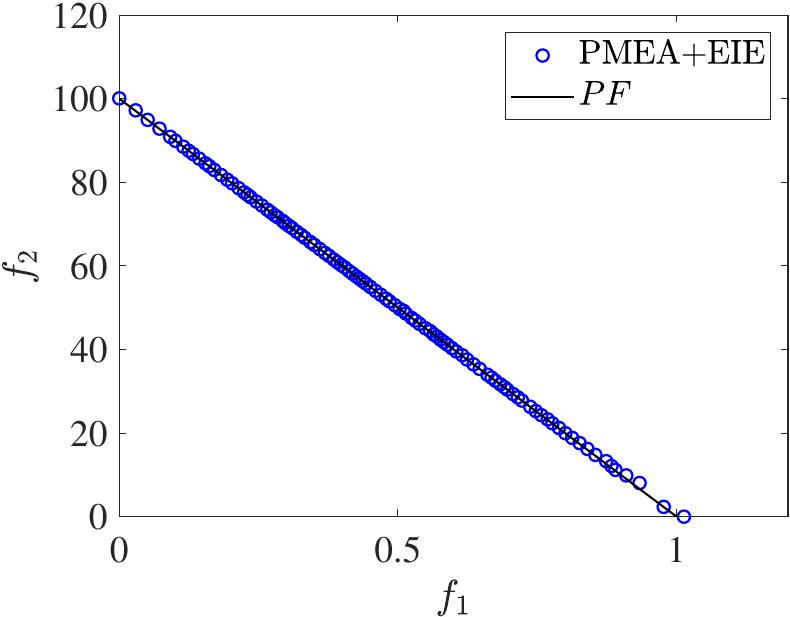}}
\subfloat[MOP11]{\includegraphics[width=0.245\linewidth]{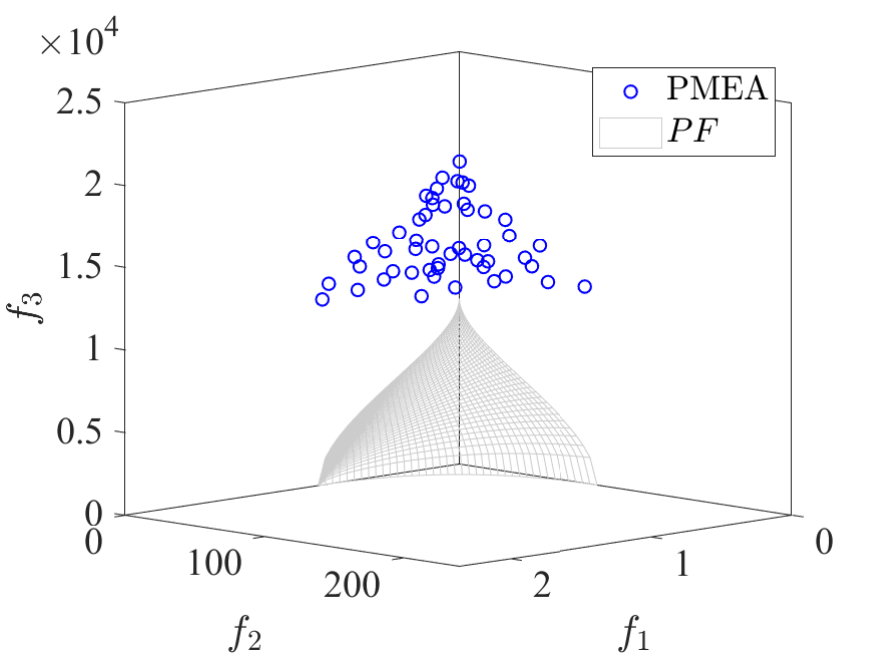}}
\subfloat[MOP11]{\includegraphics[width=0.245\linewidth]{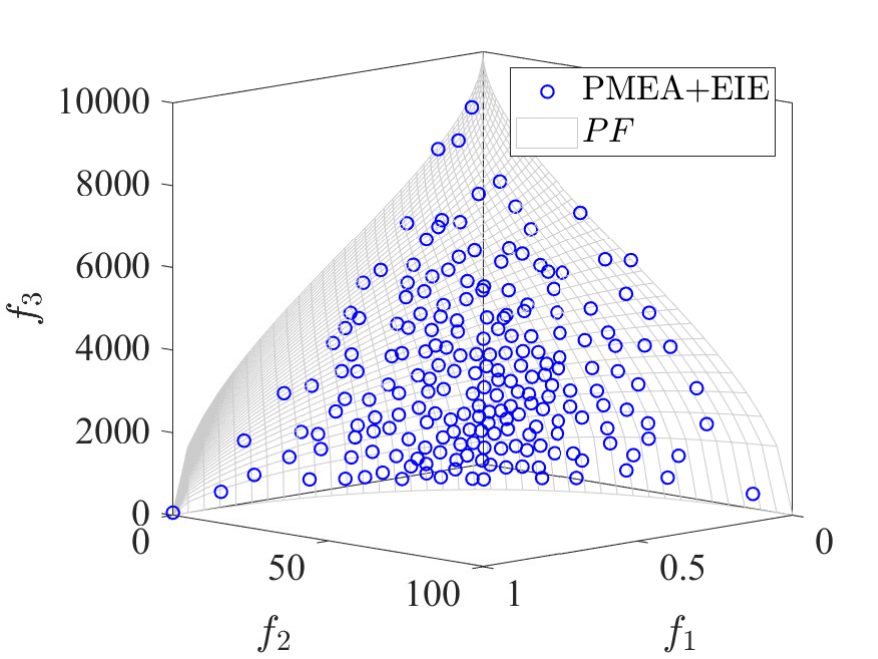}}
\hfil
\subfloat[MOP1]{\includegraphics[width=0.245\linewidth]{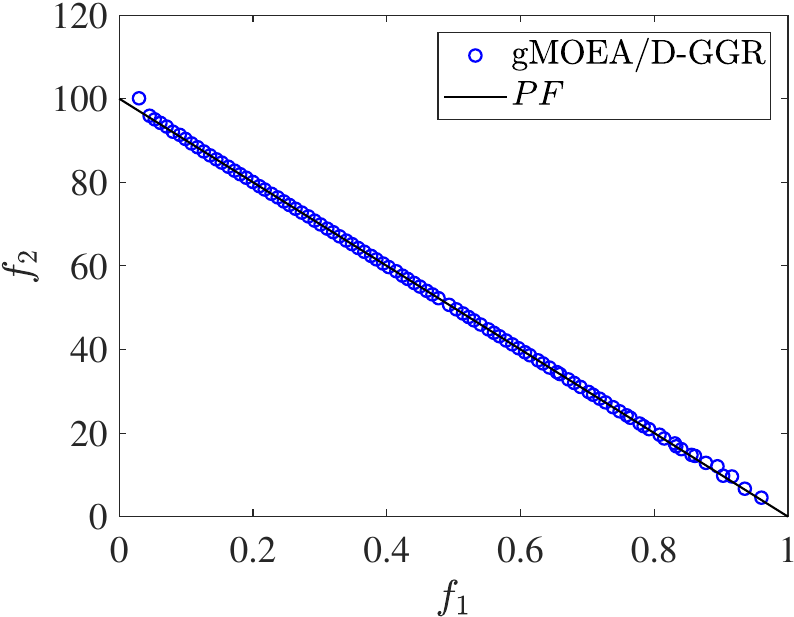}}
\subfloat[MOP1]{\includegraphics[width=0.245\linewidth]{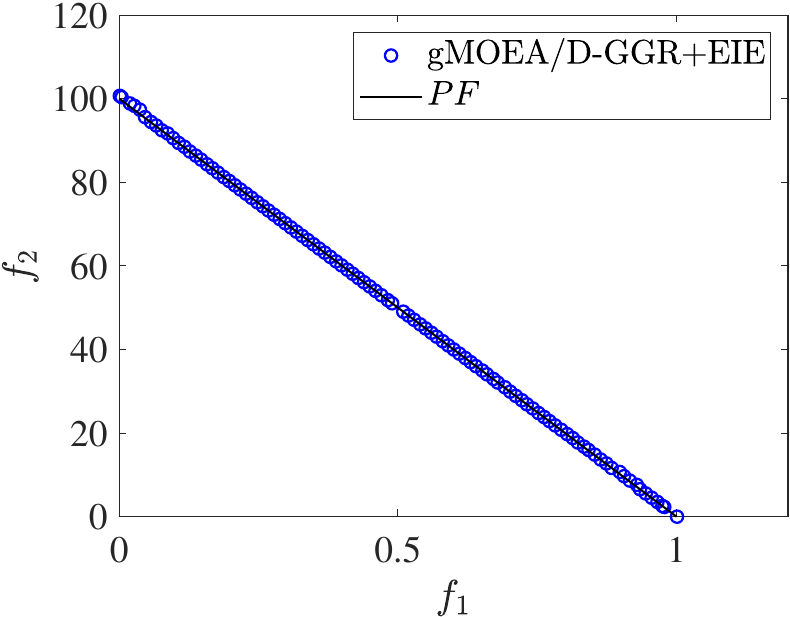}}
\subfloat[MOP11]{\includegraphics[width=0.245\linewidth]{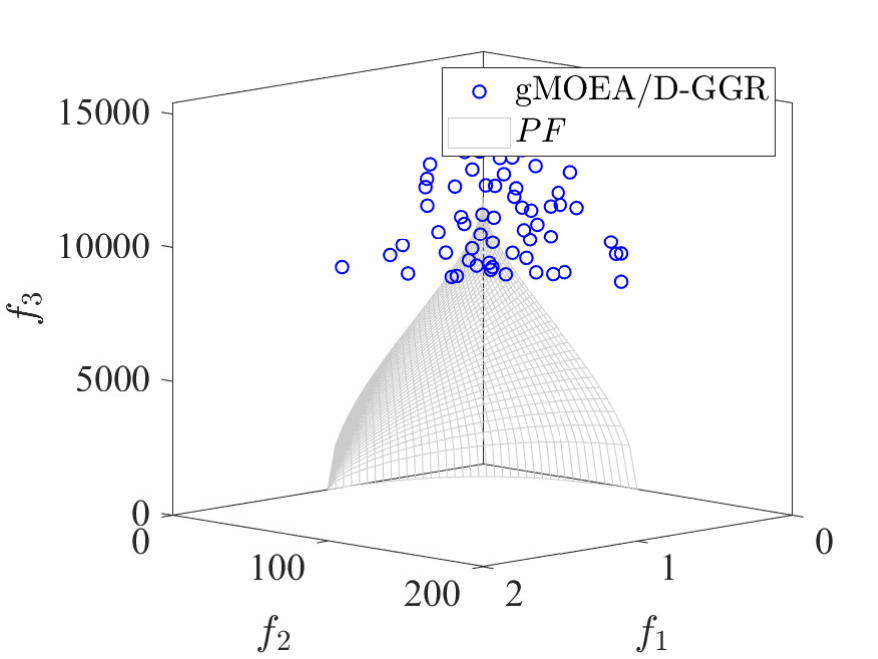}}
\subfloat[MOP11]{\includegraphics[width=0.245\linewidth]{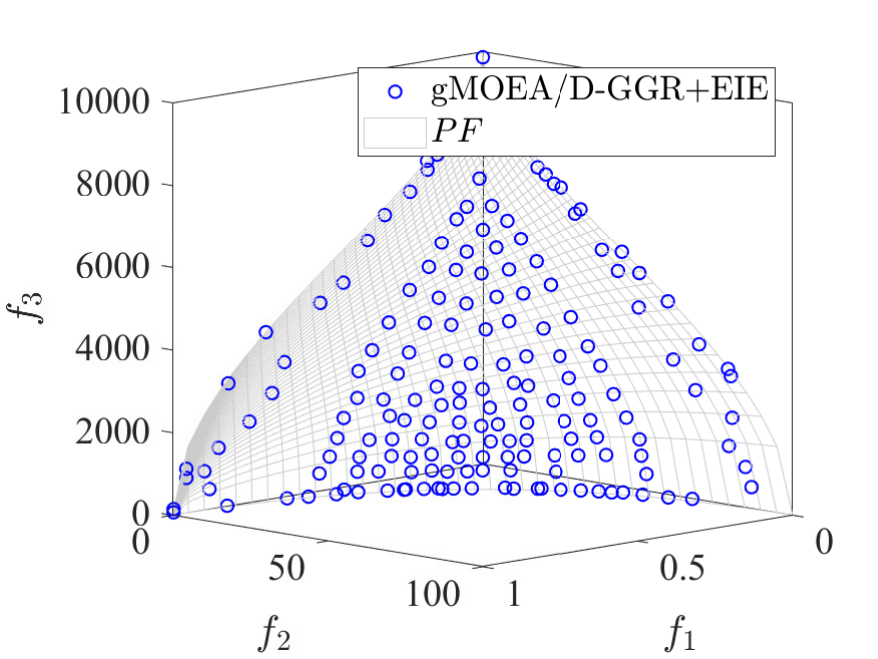}}
\hfil
\subfloat[MOP1]{\includegraphics[width=0.245\linewidth]{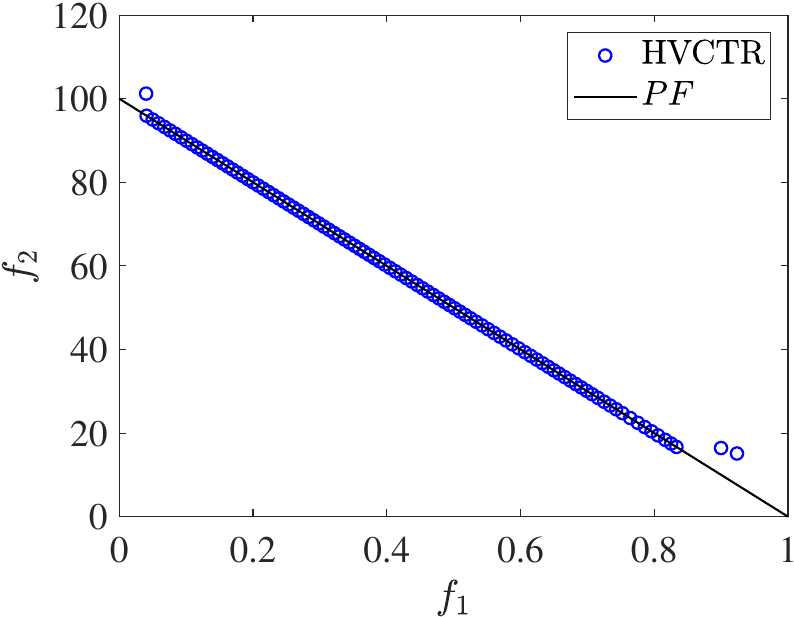}}
\subfloat[MOP1]{\includegraphics[width=0.245\linewidth]{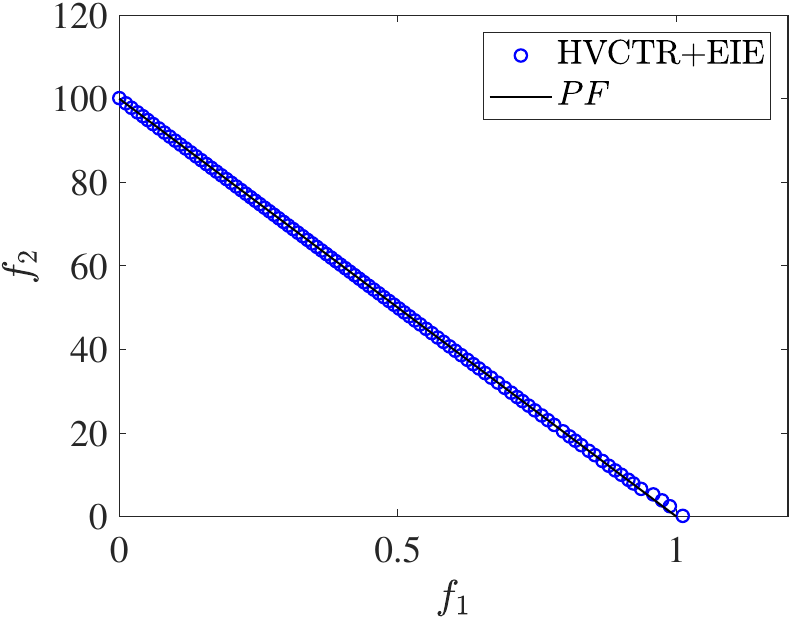}}
\subfloat[MOP11]{\includegraphics[width=0.245\linewidth]{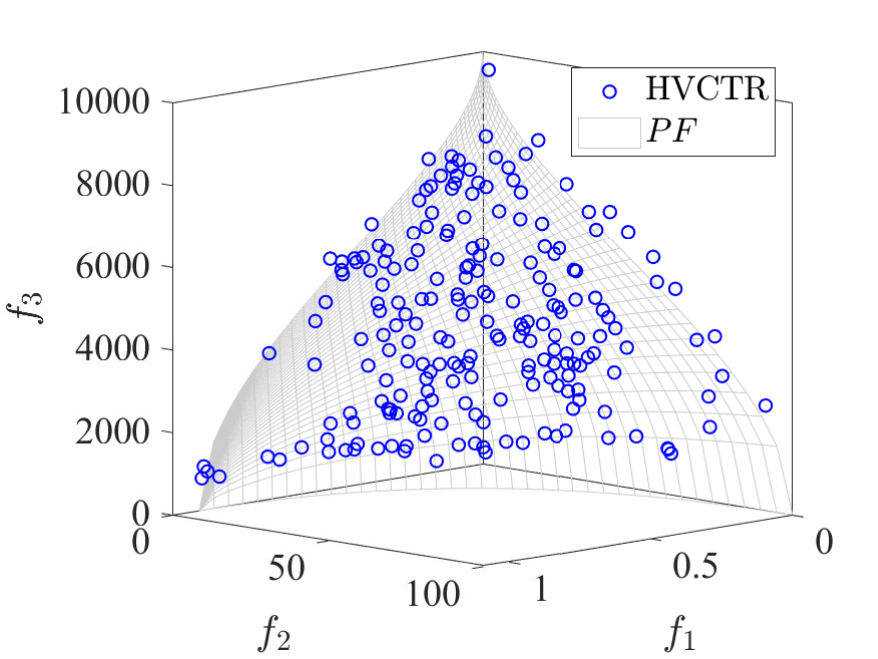}}
\subfloat[MOP11]{\includegraphics[width=0.245\linewidth]{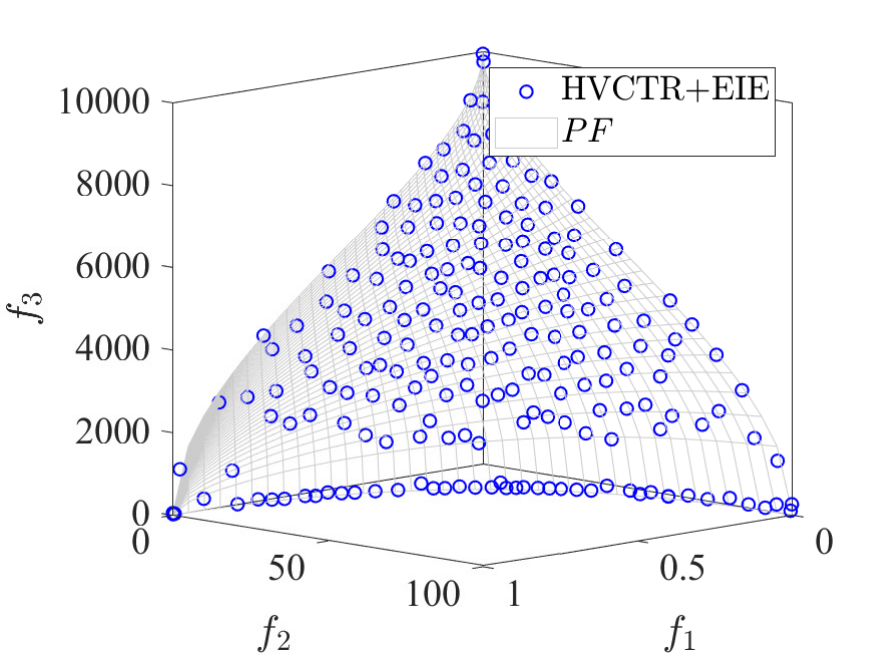}}
\caption{Plots of the non-dominated sets with the median $\operatorname{HV}$ metric values found by the MOEA and the MOEA with EIE.}
\label{fig:ndobjs_baseline}
\end{figure*}

\begin{figure*}[ht]
\centering
\subfloat[MOP1]{\includegraphics[width=0.245\linewidth]{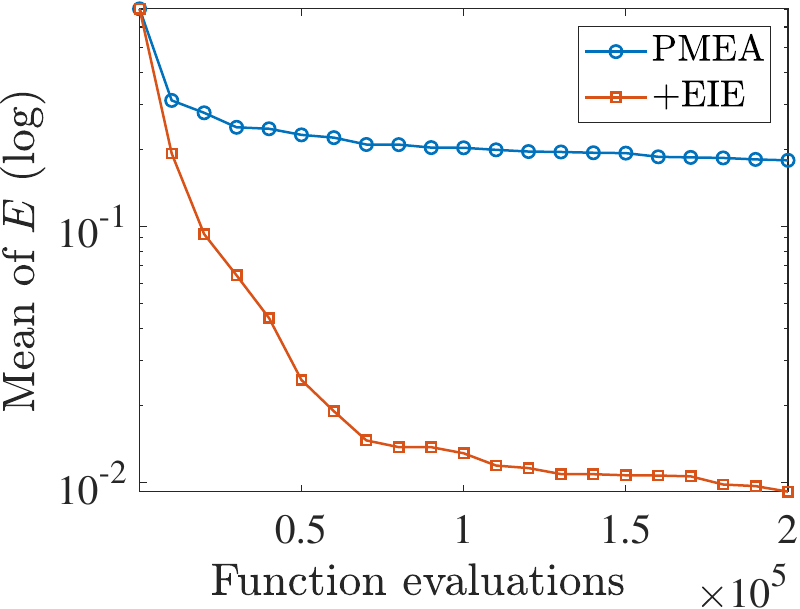}}
\subfloat[MOP1]{\includegraphics[width=0.245\linewidth]{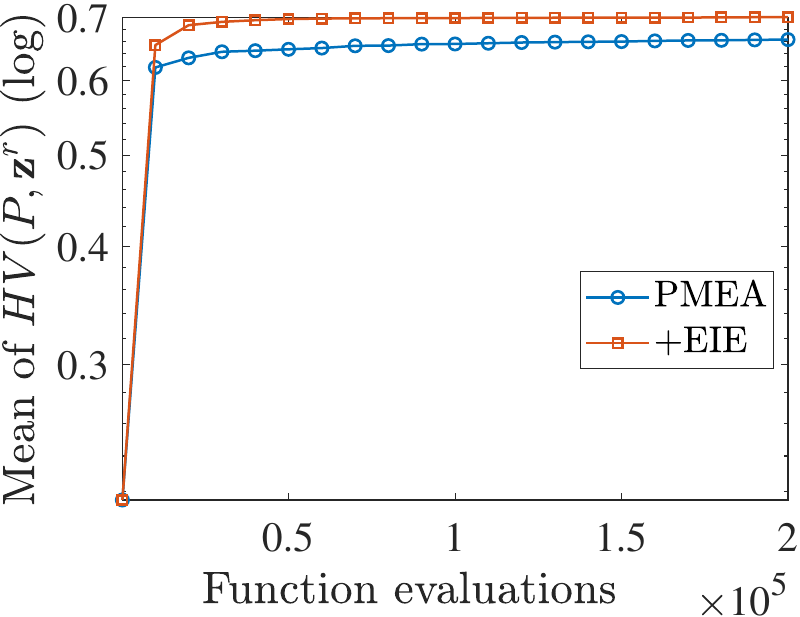}}
\subfloat[MOP11]{\includegraphics[width=0.245\linewidth]{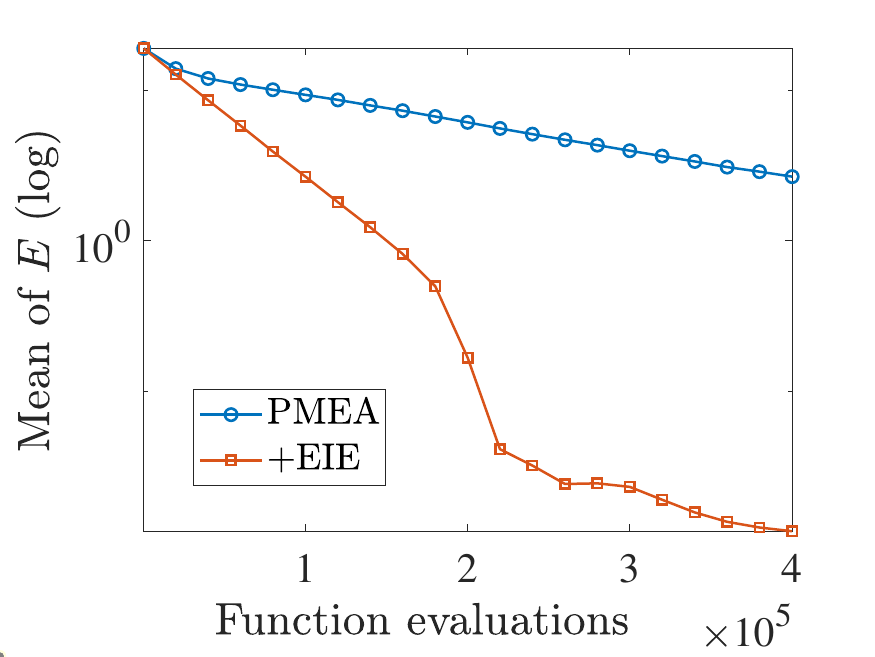}}
\subfloat[MOP11]{\includegraphics[width=0.245\linewidth]{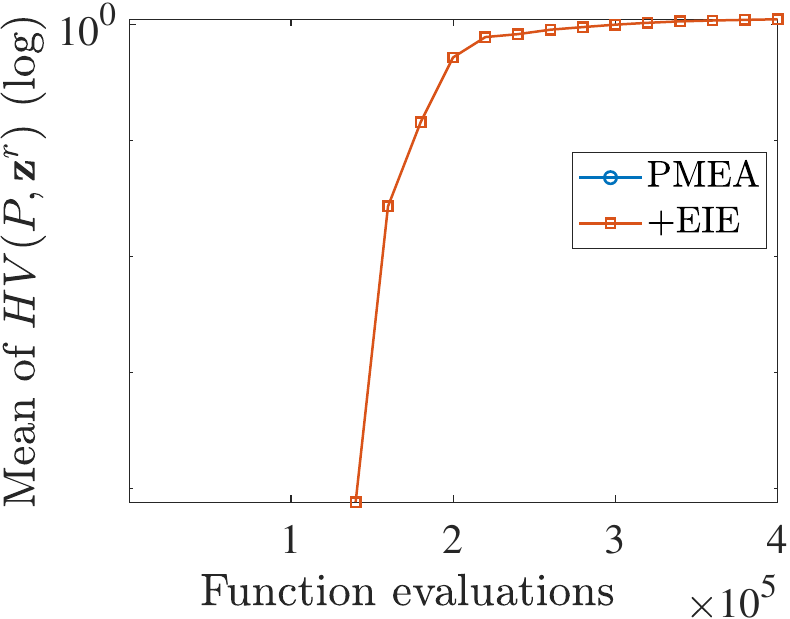}}
\hfil
\subfloat[MOP1]{\includegraphics[width=0.245\linewidth]{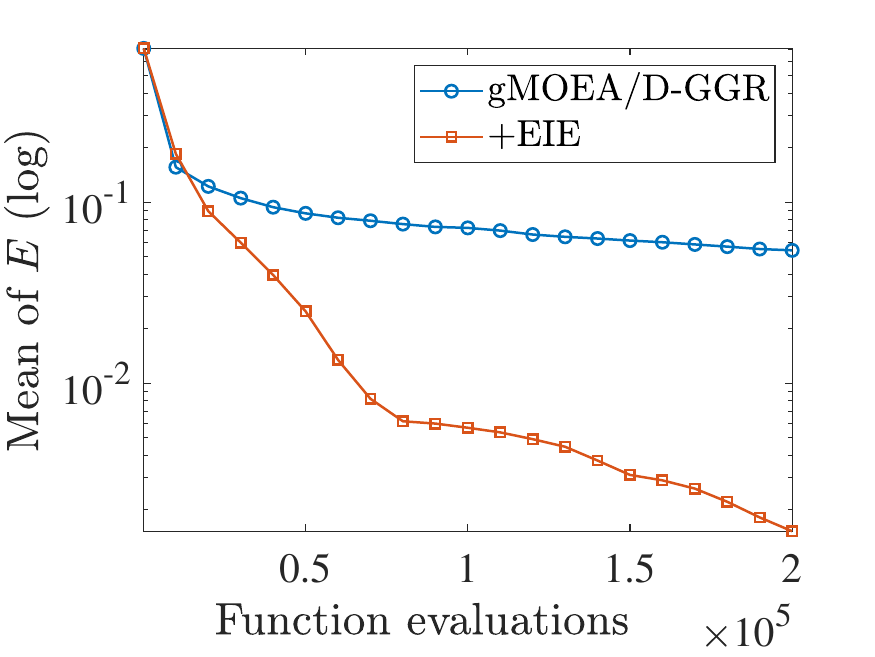}}
\subfloat[MOP1]{\includegraphics[width=0.245\linewidth]{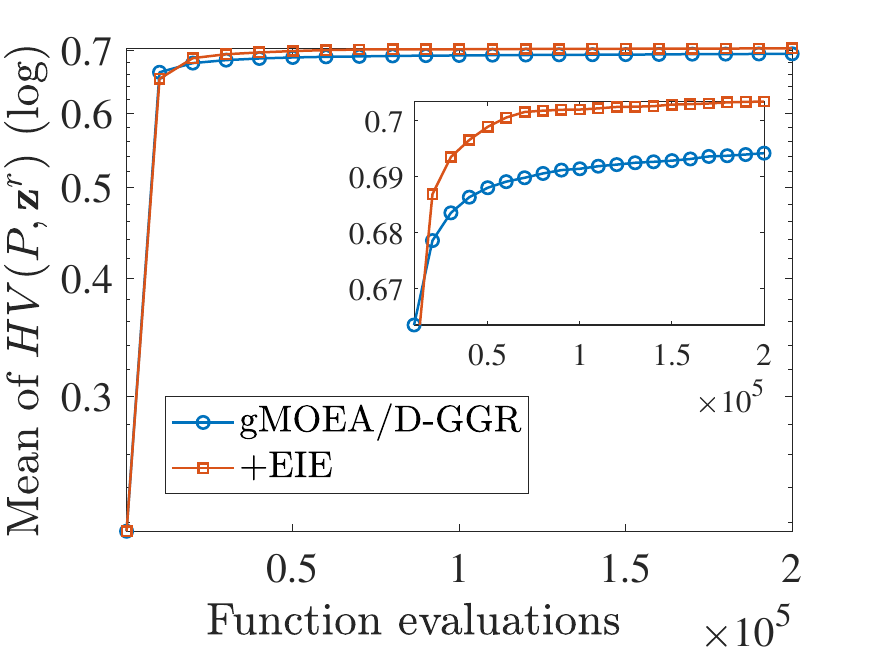}}
\subfloat[MOP11]{\includegraphics[width=0.245\linewidth]{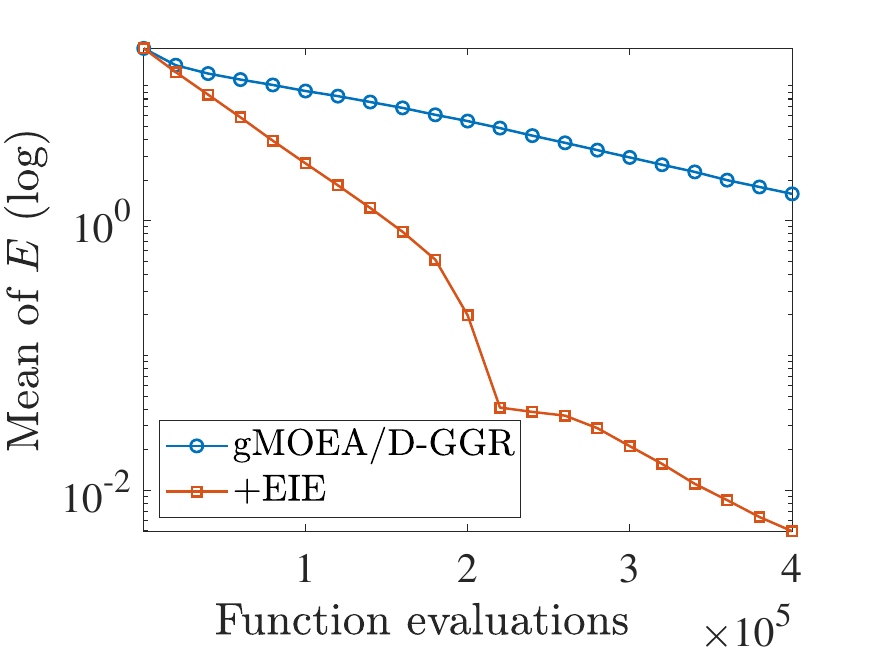}}
\subfloat[MOP11]{\includegraphics[width=0.245\linewidth]{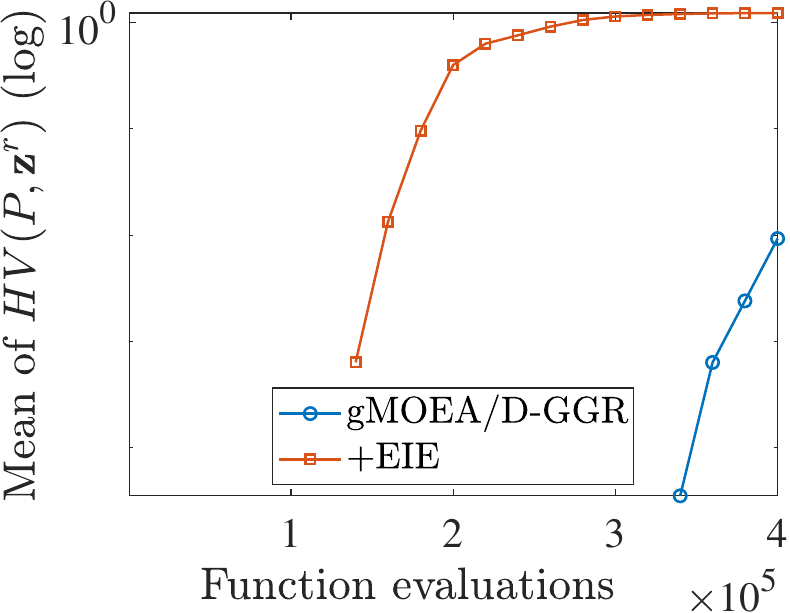}}
\hfil
\subfloat[MOP1]{\includegraphics[width=0.245\linewidth]{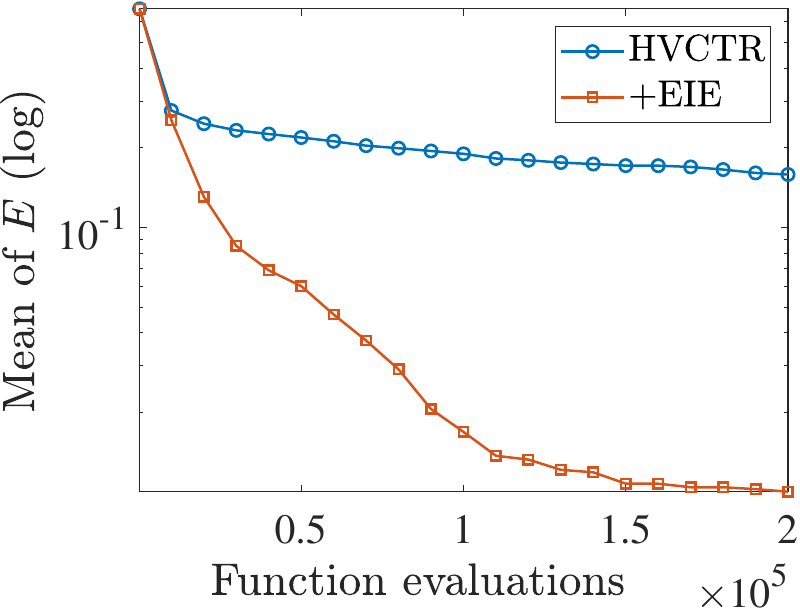}}
\subfloat[MOP1]{\includegraphics[width=0.245\linewidth]{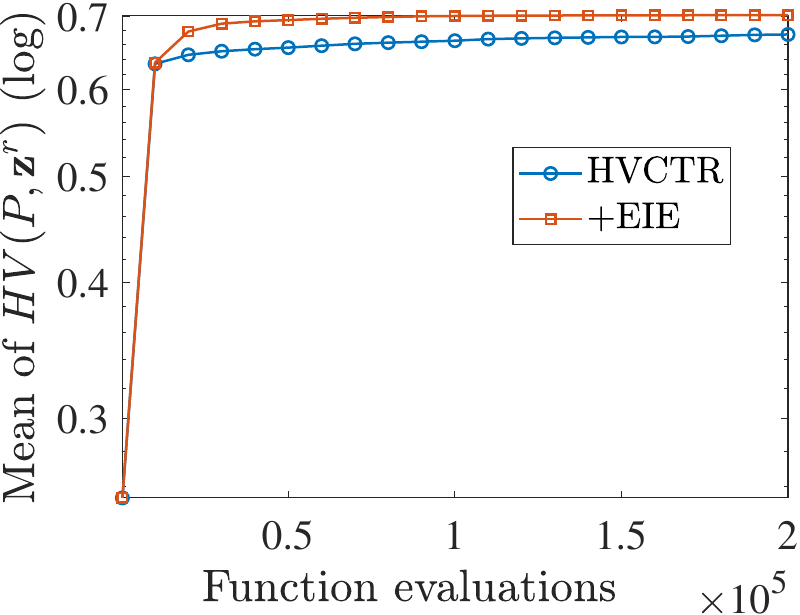}}
\subfloat[MOP11]{\includegraphics[width=0.245\linewidth]{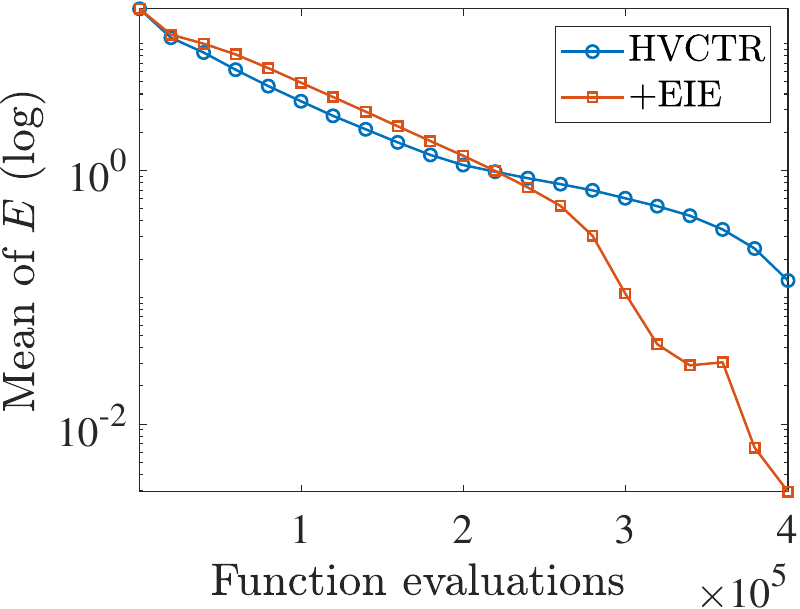}\label{fig:mc11_metric26_mops3_algs20}}
\subfloat[MOP11]{\includegraphics[width=0.245\linewidth]{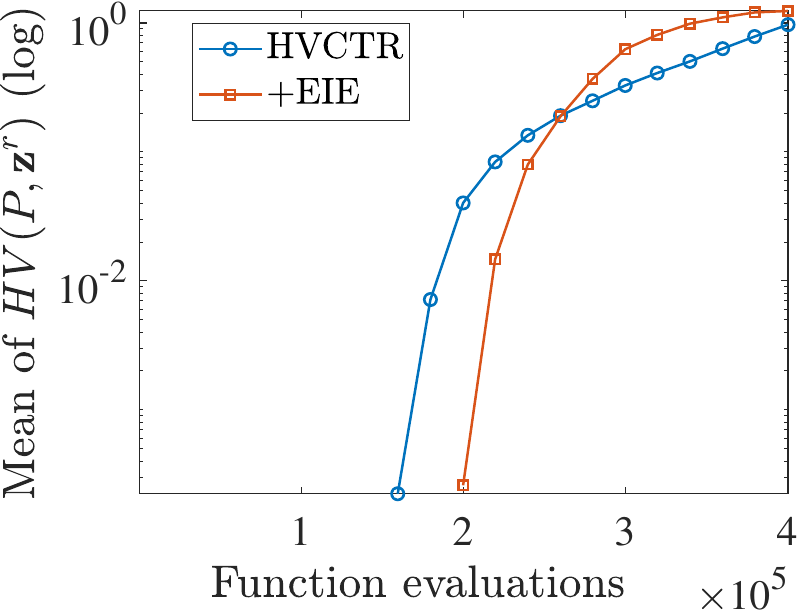}\label{fig:mc11_metric1_mops3_algs20}}
\caption{Curves of the metric value versus the number of function evaluations with the median metric values obtained by the MOEA and the MOEA with EIE. The missing data points of the curves on MOP11 indicate the corresponding $\operatorname{HV}$ metric values are zero. For example, the $\operatorname{HV}$ metric value of PMEA on MOP11 remains zero throughout the evolutionary process, and thus the corresponding curve is not shown in the plot.}
\label{fig:metric_curve_baseline}
\end{figure*}

\subsection{Effectiveness across Algorithms}\label{ssec:exp_eff}
We combine EIE with each of the three MOEAs individually. The $\operatorname{E}$ metric values obtained by the 6 algorithms on 16 test instances are given in Table~\ref{tab:baseline_e}. From the table, it can be found that the MOEAs with EIE have lower errors across almost all instances. HVCTR with EIE is outperformed by HVCTR in terms of the error on MOP3. Nevertheless, their error gap on MOP3 is very small ($|\Delta|<10^{-4}$). Moreover, enhanced versions of PMEA and gMOEA/D-GGR have lower errors than the original versions on MOP3. PMEA with EIE only has a statistically similar error to PMEA on MOP5. Enhanced gMOEA/D-GGR has significantly smaller errors than PMEA on all instances. Across all results, the mean estimated errors are no greater than $5\%$ except for: PMEA with EIE on MOP5 and MOP7; and HVCTR with EIE on MOP10.
We also report the mean and standard deviation of the number of function evaluations used by EIE in Table~\ref{tab:baseline_eiefes}. EIE requires different numbers of function evaluations when combined with different MOEAs, which suggests that the information exchange between the MOEA and EIE considerably affects the behavior of EIE. The results also demonstrate that EIE can dynamically determine the required computational resources according to different optimization problems.

\begin{table}[t]
\footnotesize
  \centering
  \caption{Percentage of function evaluations used by EIE (\%).}
    \renewcommand{\arraystretch}{1}
    \scalebox{1}{
	\setlength{\tabcolsep}{2mm}{
    \begin{tabular}{cccc}
    \toprule
    Instance & PMEA  & gMOEA/D-GGR & HVCTR \\
    \midrule
    MOP1  & 37.88±3.861 & 36.41±5.349 & 23.29±4.286 \\
    MOP2  & 41.62±3.769 & 36.81±4.856 & 22.62±2.358 \\
    MOP3  & 57.08±3.215 & 54.63±4.59 & 41.31±0.0589 \\
    MOP4  & 31.48±9.783 & 25.56±7.172 & 27.42±4.684 \\
    MOP5  & 22.16±7.157 & 25.35±9.419 & 23.84±7.746 \\
    MOP6  & 13.08±2.573 & 12.61±2.268 & 13.81±3.003 \\
    MOP7  & 19.85±0.8489 & 16.87±2.79 & 11.68±0.2611 \\
    MOP8  & 10.57±2.093 & 13.61±0.4067 & 10.2±0.9186 \\
    MOP9  & 20.24±7.995 & 19.28±5.502 & 11.17±0.3457 \\
    MOP10 & 15.91±0.773 & 15.35±1.184 & 12.15±0.4672 \\
    MOP11 & 35.29±0.5464 & 30.3±0.6057 & 35.55±0.5505 \\
    MOP12 & 22.58±3.633 & 16.98±3.91 & 22.4±2.755 \\
    MOP13 & 22.12±0.3088 & 19.1±2.817 & 14.25±0.3252 \\
    MOP14 & 14.66±1.64 & 17.25±3.825 & 18.71±3.715 \\
    MOP15 & 20.2±0.3249 & 20.04±0.3177 & 14.11±0.3151 \\
    MOP16 & 22.3±2.985 & 22.06±2.956 & 18.58±1.826 \\
    \bottomrule
    \end{tabular}%
    }
    }
  \label{tab:baseline_eiefes}%
\end{table}%

To show the relationship between the estimated error and the performance on approximating the $PF$, we present the $\operatorname{HV}$ metric values in Table~\ref{tab:baseline_hv}. We can only find that: PMEA achieves a better HV metric value than PMEA with EIE on MOP13; HVCTR outperforms the one with EIE on MOP3, but the performance gap is relatively small ($|\Delta|<10^{-5}$). In contrast, EIE can evidently improve the performance of MOEAs. For example, MOEAs perform very poorly on MOP11. Particularly, PMEA has a $\operatorname{HV}$ metric value of $0$. EIE notably enhances their performance on MOP11. Furthermore, we find that no single instance makes all the algorithms using EIE deteriorate. It manifests that EIE can effectively handle various biases.
To better demonstrate the effectiveness, the final non-dominated sets obtained by algorithms are shown in \figurename~\ref{fig:ndobjs_baseline}. On MOP1, the populations of PMEA and HVCTR only converge to certain parts of the $PF$, whereas gMOEA/D-GGR achieves sparse boundary objective vectors. On MOP11, PMEA and gMOEA/D-GGR fail to approximate the $PF$ effectively, and HVCTR results in a final population with poor diversity. However, the final results are substantially enhanced when EIE is applied to MOEAs. The boundaries of the $PF$ are identified, and the solutions are well distributed on the $PF$.
We also report the evolutionary trajectories of metrics in \figurename~\ref{fig:metric_curve_baseline}. MOEAs with EIE can generally have a faster convergence rate than MOEAs in \figurename~\ref{fig:metric_curve_baseline}, indicating better anytime performance in terms of both metrics. The convergence rate of HVCTR with EIE on MOP11 (\ie, \figurename~\ref{fig:mc11_metric26_mops3_algs20} and \figurename~\ref{fig:mc11_metric1_mops3_algs20}) further shows the relationship between the two metrics when using EIE. The convergence rate with respect to $\operatorname{E}$ is slower at the beginning but speeds up subsequently, which is consistent with the convergence trend observed for $\operatorname{HV}$.

In general, a better estimated ideal objective vector is consistent with a better approximation of the $PF$ when the MOEA uses EIE. EIE is compatible with MOEAs and truly enhances MOEAs' performance.

\section{Conclusion}\label{sec:conclusion}
This paper has systematically analyzed the challenges posed by biases in estimating the ideal objective vector. Three scenarios, that detrimentally affect the performance of population-based estimation methods, have been illustrated. A test problem generator has been proposed, which can generate various biased test instances with difficulties in approximating the ideal objective vector. A coping strategy called EIE has also been developed. EIE and MOEA execute simultaneously. In each iteration, EIE guides the population's search by providing solutions that are close to the ideal objective vector in the objective space. In EIE, solving $m$ EWS-based subproblems is involved. The search granularity and computational resources should be adaptively adjusted in each iteration of EIE.
Experimental studies have been conducted on the 16 test instances obtained by the proposed generator and 55 existing ones. EIE adopts PSA-CMA-ES as the solver, and is integrated into three representative MOEAs. The results underscore the significance of EIE and the inefficiency of population-based estimation methods.

This work validates the superiority of EIE using continuous MOPs as test instances. Future work will consider discrete and mixed MOPs. Moreover, we plan to conduct a more nuanced investigation of bias by analyzing the formulations of real-world MOPs.




\bibliographystyle{IEEEtran}
\bibliography{references}


 




\clearpage
\newpage
{
\appendices
This is supplementary material for ``Enhanced Ideal Objective Vector Estimation for Evolutionary Multi-Objective Optimization''.
\section{Proof of Theorem~\ref{the:error_bound}}\label{sec:proof}
We use \figurename~\ref{fig:proof_sketch_error_bound} to illustrate the proof. \figurename~\ref{fig:proof_sketch_error_bound} shows a 2-objective case. Since $(z_1^{nad},z_2^{ide})^\intercal$ must be the Pareto optimal objective vector, the upper bound of optimal subproblem function value is $c_1$. Any objective vector on the lowest level surface of the figure (\ie, $g_2^{ews}=c_1$) is optimal for the subproblem. Therefore, the upper bound of the estimation error is determined by the intersection of the lowest level surface and the $f_2$ axis. The formal proof is presented below.

\begin{figure}[ht]
    \centering
    \includegraphics[width = 0.7\linewidth]{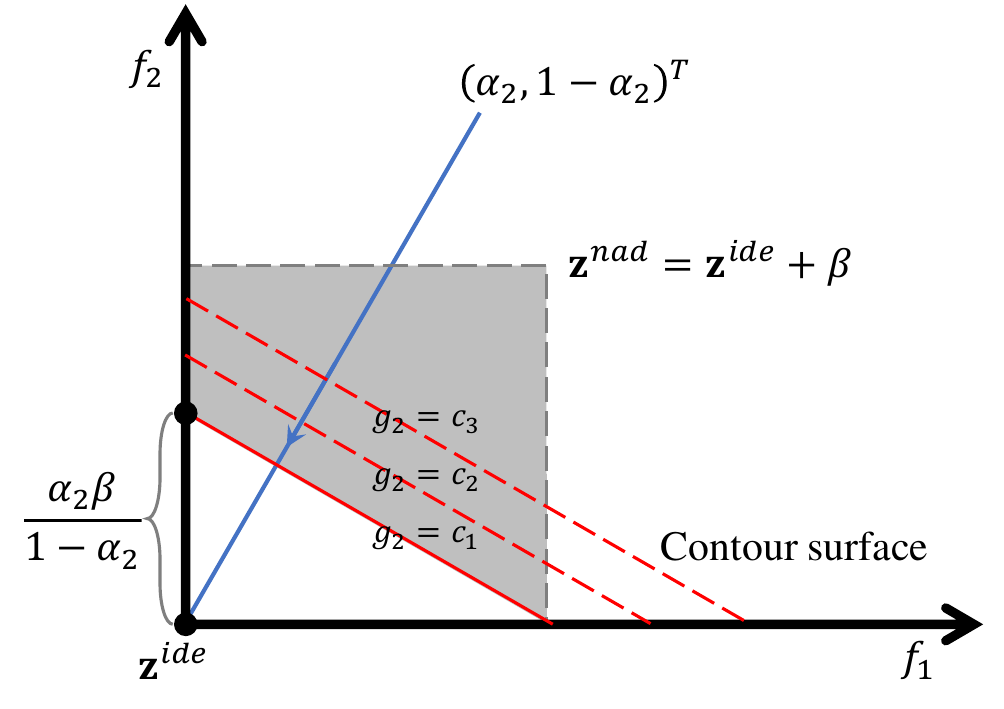}
    \caption{Illustration of the proof of Theorem~\ref{the:error_bound}.}
    \label{fig:proof_sketch_error_bound}
\end{figure}

\begin{proof}
Since $z_j^{nad}=z_j^{ide}+\beta$ for $j=1,\ldots,m$, we let $z_j^{ide}=0$ for $j=1,\ldots,m$ and $\beta=1$ without loss of generality. Then $z_j^{nad}=1$ for $j=1,\ldots,m$. Recall that $\mathbf{z}^i$ is the optimal objective vector of the $i$-th EWS-based subproblem. $z_i^i$ is bounded by the optimal value of the following problem
\begin{equation}\label{eqn:error_bound}
    \begin{array}{ll}
        \text{max.} & z_i,
        \\
        \text{s.t.} & (1-\alpha_i)z_i+\frac{\alpha_i}{m-1}\sum\limits_{\substack{j=1 \\ \wedge j\neq i}}^m z_j = c_i, \\
        & 0\leq z_j\leq 1 \text{ for } j=1,\ldots,m.
    \end{array}
\end{equation}
To achieve the largest optimal value for Problem~\eqref{eqn:error_bound}, $c_i$ should be the upper bound of the $i$-th EWS-based subproblem's optimal value, which is formulated by
\begin{equation}\label{eqn:subp_bound_raw}
    \sup_{Z}\min_{\mathbf{z}\in Z} \left( (1-\alpha_i)z_i+\frac{\alpha_i}{m-1}\sum\limits_{\substack{j=1 \\ \wedge j\neq i}}^m z_j\right),
\end{equation}
where $Z$ satisfies:
\begin{equation}
\begin{aligned}
    & Z\subset [0,1]^m; \\
    & \forall j\in\{1,\ldots,m\} \text{ and } \mathbf{z}^\prime\in Z, \\
    & \exists \mathbf{z} \!\in\! Z \text{ such that } z_j \!=\! 0 \text{ and } \mathbf{z} \text{ is not dominated by } \mathbf{z}^\prime; \\
    & \forall j\in\{1,\ldots,m\} \text{ and } \mathbf{z}^\prime\in Z, \\
    & \exists \mathbf{z} \!\in\! Z \text{ such that } z_j \!=\! 1 \text{ and } \mathbf{z} \text{ is not dominated by } \mathbf{z}^\prime. \\
\end{aligned}
\end{equation}
If we only consider $Z\subset [0,1]^m$, then Eq.~\eqref{eqn:subp_bound_raw} is equal to $1$ (\ie, $\mathbf{z}=(1,\ldots,1)^\intercal$). We have $\alpha_i\leq 0.5$, implying that $\alpha_i\leq 1-\alpha_i$. Then Eq.~\eqref{eqn:subp_bound_raw} is equivalent to the optimal value of the following problem
\begin{equation}\label{eqn:subp_bound}
    \begin{array}{ll}
        \text{min.} & (1-\alpha_i)z_i+\frac{\alpha_i}{m-1}\sum\limits_{\substack{j=1 \\ \wedge j\neq i}}^m z_j,
        \\
        \text{s.t.} & 0\leq z_i\leq 1, \\
        & z_j=1 \text{ for every } j\in\{1,\ldots,m\}\setminus\{i\}.
    \end{array}
\end{equation}
The optimal solution of Problem~\eqref{eqn:subp_bound} is 
\begin{equation}
    z_j=
    \begin{cases}
        0, & j=i, \\
        1, & j\neq i,
    \end{cases}
\end{equation}
which is straightforward. The corresponding optimal value is $\alpha_i$. That is, $c_i=\alpha_i$. To maximize $z_i$ in Problem~\eqref{eqn:error_bound}, $z_j=0$ for every $j\in\{1,\ldots,m\}\setminus\{i\}$. Thus, the optimal value of Problem~\eqref{eqn:error_bound} is $\frac{\alpha_i}{1-\alpha_i}$.
\end{proof}

\section{More Experiments}\label{sec:discussions}
In this section, we further investigate EIE empirically. Corresponding experimental settings are the same as Section~\ref{sec:exp_set}.

\subsection{Comparison with Other Population-Based Methods}
We further investigate EIE by comparing it with other population-based methods. Two representative methods are selected: UT~\cite{wang2019generator} and DRP~\cite{wang2017effect}. They generally estimate the ideal objective vector by
\begin{equation}
    z_i^{e} = z_i^\prime - \beta_i \text{ for } i=1,\ldots,m,
\end{equation}
where $z_i^\prime$ is calculated using Eq.~\eqref{eqn:pop_ide_iter}. Their differences lie in the specification of $\beta_i$. For convenience, we assume a normalized objective space. Then,
\begin{equation}
\begin{aligned}
    & \text{UT:} \quad \beta_i=0.1, \\
    & \text{DRP:} \quad \beta_i=(1-10^{-3})\frac{FE_{max}-FE}{FE_{max}}+10^{-3},
\end{aligned}
\end{equation}
where $FE_{max}$ is the maximum number of function evaluations and $FE$ is the number of evaluations currently consumed.

The results are displayed in Table~\ref{tab:vs_pop_e} and Table~\ref{tab:vs_pop_hv}. EIE significantly outperforms UT and DRP across most instances, regardless of the chosen algorithm. Specifically, PMEA with UT surpasses its counterpart utilizing EIE in terms of the HV metric on 2/16 of instances (\ie, MOP5 and MOP13). HVCTR with UT outperforms the one with EIE in terms of the HV metric on MOP3 only. DRP yields similar statistical results to UT when compared with EIE.
However, gMOEA/D-GGR with UT or DRP cannot have the best HV metric value on any test instance. Although gMOEA/D-GGR with UT achieves the smallest error on MOP3, gMOEA/D-GGR with EIE has a reasonably small error ($\operatorname{E}<10^{-3}$) and performs the best in terms of the HV metric. Besides, UT has better average ranks than DRP in terms of both metrics. We also discover that any instance still cannot show that EIE degrades the performance of all algorithms. A smaller error remains associated with a higher HV metric value.

Overall, this section confirms the superiority of EIE over population-based methods.

\begin{table*}[ht]
\footnotesize
  \centering
  \caption{Comparisons of $\operatorname{E}$ metric values between population-based methods and EIE.}
    \renewcommand{\arraystretch}{1.2}
    \scalebox{1}{
	\setlength{\tabcolsep}{0.8mm}{
    \begin{tabular}{cc||ccc|ccc|ccc}
    \toprule
    \multirow{2}[1]{*}{Instance} & \multirow{2}[1]{*}{$\operatorname{E}$} & \multicolumn{3}{c|}{PMEA} & \multicolumn{3}{c|}{gMOEA/D-GGR} & \multicolumn{3}{c}{HVCTR} \\
\cline{3-11}          &       & +UT   & +DRP  & +EIE  & +UT   & +DRP  & +EIE  & +UT   & +DRP  & +EIE \\
    \midrule
    \multirow{2}[1]{*}{MOP1} & mean  & 0.201(3)- & 0.1915(2)- & \cellcolor[rgb]{ .651,  .651,  .651}0.009188(1) & 0.006156(2)- & 0.0314(3)- & \cellcolor[rgb]{ .651,  .651,  .651}0.001522(1) & 0.1576(3)- & 0.1564(2)- & \cellcolor[rgb]{ .651,  .651,  .651}0.01002(1) \\
          & std.  & 0.03184 & 0.04747 & 0.02282 & 0.001461 & 0.00586 & 0.004806 & 0.02071 & 0.02185 & 0.02137 \\
    \multirow{2}[0]{*}{MOP2} & mean  & 0.2251(3)- & 0.2242(2)- & \cellcolor[rgb]{ .651,  .651,  .651}0.0008158(1) & \cellcolor[rgb]{ .651,  .651,  .651}1.85e-06(1)+ & 0.03705(3)- & 0.0002705(2) & 0.2306(2)- & 0.2324(3)- & \cellcolor[rgb]{ .651,  .651,  .651}0.000584(1) \\
          & std.  & 0.0724 & 0.05387 & 0.001042 & 3.655e-06 & 0.01483 & 0.000606 & 0.02822 & 0.02777 & 0.0007168 \\
    \multirow{2}[0]{*}{MOP3} & mean  & 0.07432(2)- & 0.07662(3)- & \cellcolor[rgb]{ .651,  .651,  .651}0.0002752(1) & 0.8334(2)- & 2.354(3)- & \cellcolor[rgb]{ .651,  .651,  .651}0.002226(1) & 4.328e-05(2)+ & \cellcolor[rgb]{ .651,  .651,  .651}3.416e-05(1)+ & 0.0001058(3) \\
          & std.  & 0.08676 & 0.0929 & 0.0004807 & 0.3271 & 0.3371 & 0.002196 & 3.895e-05 & 2.367e-05 & 8.12e-05 \\
    \multirow{2}[0]{*}{MOP4} & mean  & 0.3065(2)- & 0.363(3)- & \cellcolor[rgb]{ .651,  .651,  .651}0.00666(1) & 0.04167(2)- & 0.377(3)- & \cellcolor[rgb]{ .651,  .651,  .651}0.002265(1) & 0.009969(3)- & 0.009777(2)- & \cellcolor[rgb]{ .651,  .651,  .651}0.001255(1) \\
          & std.  & 0.04218 & 0.07187 & 0.00544 & 0.03122 & 0.1026 & 0.001718 & 0.03995 & 0.03998 & 0.001938 \\
    \multirow{2}[0]{*}{MOP5} & mean  & \cellcolor[rgb]{ .651,  .651,  .651}0.06431(1)+ & 0.06444(2)+ & 0.08947(3) & 0.06647(2)- & 0.3326(3)- & \cellcolor[rgb]{ .651,  .651,  .651}0.003203(1) & 0.02334(2)- & 0.02368(3)- & \cellcolor[rgb]{ .651,  .651,  .651}0.003335(1) \\
          & std.  & 0.00904 & 0.01042 & 0.01599 & 0.03274 & 0.08003 & 0.003619 & 0.004572 & 0.004444 & 0.003176 \\
    \multirow{2}[0]{*}{MOP6} & mean  & 0.244(2)- & 0.2619(3)- & \cellcolor[rgb]{ .651,  .651,  .651}0.03709(1) & 0.1443(2)- & 0.1905(3)- & \cellcolor[rgb]{ .651,  .651,  .651}0.01745(1) & 0.2258(2)- & 0.2259(3)- & \cellcolor[rgb]{ .651,  .651,  .651}0.03863(1) \\
          & std.  & 0.02836 & 0.03612 & 0.05303 & 0.01407 & 0.02397 & 0.02398 & 0.02924 & 0.02554 & 0.04247 \\
    \multirow{2}[0]{*}{MOP7} & mean  & \cellcolor[rgb]{ .651,  .651,  .651}0.07223(1)= & 0.08131(3)- & 0.07419(2) & 0.01436(2)- & 0.01994(3)- & \cellcolor[rgb]{ .651,  .651,  .651}0.002601(1) & 0.05363(2)- & 0.05363(3)- & \cellcolor[rgb]{ .651,  .651,  .651}0.002825(1) \\
          & std.  & 0.007664 & 0.01088 & 0.006305 & 0.008928 & 0.01146 & 0.0002094 & 0.00506 & 0.00506 & 0.000122 \\
    \multirow{2}[0]{*}{MOP8} & mean  & 0.07769(3)- & 0.03801(2)- & \cellcolor[rgb]{ .651,  .651,  .651}0.0002278(1) & 0.008353(2)- & 0.01076(3)- & \cellcolor[rgb]{ .651,  .651,  .651}3.464e-05(1) & 0.1596(3)- & 0.1539(2)- & \cellcolor[rgb]{ .651,  .651,  .651}0.0002018(1) \\
          & std.  & 0.08953 & 0.06428 & 0.0006233 & 0.008644 & 0.008328 & 7.318e-06 & 0.07512 & 0.0794 & 0.0005824 \\
    \multirow{2}[0]{*}{MOP9} & mean  & 0.1481(3)- & 0.1465(2)- & \cellcolor[rgb]{ .651,  .651,  .651}0.03804(1) & 0.04322(3)- & 0.04057(2)- & \cellcolor[rgb]{ .651,  .651,  .651}0.0043(1) & 0.1295(2)- & 0.1344(3)- & \cellcolor[rgb]{ .651,  .651,  .651}0.00409(1) \\
          & std.  & 0.01179 & 0.007871 & 0.04788 & 0.01162 & 0.009412 & 0.006907 & 0.007538 & 0.0252 & 0.000284 \\
    \multirow{2}[0]{*}{MOP10} & mean  & 0.2708(3)- & 0.2625(2)- & \cellcolor[rgb]{ .651,  .651,  .651}0.04694(1) & 0.1345(3)- & 0.1174(2)- & \cellcolor[rgb]{ .651,  .651,  .651}0.04931(1) & 0.2627(3)- & 0.2624(2)- & \cellcolor[rgb]{ .651,  .651,  .651}0.06434(1) \\
          & std.  & 0.03613 & 0.03798 & 0.006474 & 0.02447 & 0.04278 & 0.007749 & 0.04876 & 0.04866 & 0.007782 \\
    \multirow{2}[0]{*}{MOP11} & mean  & 2.815(2)- & 2.913(3)- & \cellcolor[rgb]{ .651,  .651,  .651}0.0118(1) & 8.719(2)- & 12.18(3)- & \cellcolor[rgb]{ .651,  .651,  .651}0.004962(1) & 0.07146(2)- & 0.1028(3)- & \cellcolor[rgb]{ .651,  .651,  .651}0.002938(1) \\
          & std.  & 0.1892 & 0.2082 & 0.02367 & 0.8978 & 0.7376 & 0.00248 & 0.07383 & 0.09318 & 0.003479 \\
    \multirow{2}[0]{*}{MOP12} & mean  & 0.2665(2)- & 0.2778(3)- & \cellcolor[rgb]{ .651,  .651,  .651}0.0196(1) & 0.7916(2)- & 1.315(3)- & \cellcolor[rgb]{ .651,  .651,  .651}0.008492(1) & 0.1197(2)- & 0.1205(3)- & \cellcolor[rgb]{ .651,  .651,  .651}0.008418(1) \\
          & std.  & 0.01754 & 0.02634 & 0.04611 & 0.09784 & 0.07243 & 0.01586 & 0.0227 & 0.0184 & 0.02085 \\
    \multirow{2}[0]{*}{MOP13} & mean  & 0.02204(2)- & 0.02387(3)- & \cellcolor[rgb]{ .651,  .651,  .651}0.00107(1) & 0.03991(2)- & 0.04107(3)- & \cellcolor[rgb]{ .651,  .651,  .651}0.0001564(1) & 0.01113(2)- & 0.01161(3)- & \cellcolor[rgb]{ .651,  .651,  .651}0.0005526(1) \\
          & std.  & 0.001932 & 0.002773 & 0.001177 & 0.006416 & 0.006323 & 5.538e-05 & 0.0009944 & 0.001121 & 0.0001007 \\
    \multirow{2}[0]{*}{MOP14} & mean  & 0.07866(3)- & 0.06842(2)- & \cellcolor[rgb]{ .651,  .651,  .651}8.162e-06(1) & 0.1011(2)- & 0.1772(3)- & \cellcolor[rgb]{ .651,  .651,  .651}1.065e-05(1) & 0.0289(2)- & 0.0338(3)- & \cellcolor[rgb]{ .651,  .651,  .651}4.806e-06(1) \\
          & std.  & 0.05063 & 0.0478 & 1.127e-05 & 0.05966 & 0.09907 & 1.079e-05 & 0.01527 & 0.01518 & 4.602e-06 \\
    \multirow{2}[0]{*}{MOP15} & mean  & 0.05853(3)- & 0.05497(2)- & \cellcolor[rgb]{ .651,  .651,  .651}0.00386(1) & 0.1122(3)- & 0.1012(2)- & \cellcolor[rgb]{ .651,  .651,  .651}0.0006248(1) & 0.03798(3)- & 0.03789(2)- & \cellcolor[rgb]{ .651,  .651,  .651}0.003614(1) \\
          & std.  & 0.004726 & 0.005569 & 0.007412 & 0.0284 & 0.007032 & 0.0002324 & 0.002433 & 0.002037 & 0.0005892 \\
    \multirow{2}[1]{*}{MOP16} & mean  & 0.1631(3)- & 0.1194(2)- & \cellcolor[rgb]{ .651,  .651,  .651}0.03796(1) & 0.1189(3)- & 0.03241(2)- & \cellcolor[rgb]{ .651,  .651,  .651}0.02743(1) & 0.07954(3)- & 0.07888(2)- & \cellcolor[rgb]{ .651,  .651,  .651}0.03545(1) \\
          & std.  & 0.02113 & 0.03188 & 0.004456 & 0.03808 & 0.03841 & 0.00393 & 0.01647 & 0.01543 & 0.003612 \\
    \midrule
    \multicolumn{2}{c|}{Total +/=/-} & 1/1/14 & 1/0/15 & \textbackslash{} & 1/0/15 & 0/0/16 & \textbackslash{} & 1/0/15 & 1/0/15 & \textbackslash{} \\
    \multicolumn{2}{c|}{Average rank} & 2.375(2) & 2.4375(3) & 1.1875(1) & 2.1875(2) & 2.75(3) & 1.0625(1) & 2.375(2) & 2.5(3) & 1.125(1) \\
    \bottomrule
    \end{tabular}%
    }
    }
  \label{tab:vs_pop_e}%
\end{table*}%

\begin{table*}[ht]
\footnotesize
  \centering
  \caption{Comparisons of $\operatorname{HV}$ metric values between population-based methods and EIE.}
    \renewcommand{\arraystretch}{1.2}
    \scalebox{1}{
	\setlength{\tabcolsep}{0.8mm}{
    \begin{tabular}{cc||ccc|ccc|ccc}
    \toprule
    \multirow{2}[1]{*}{Instance} & \multirow{2}[1]{*}{$\operatorname{HV}$} & \multicolumn{3}{c|}{PMEA} & \multicolumn{3}{c|}{gMOEA/D-GGR} & \multicolumn{3}{c}{HVCTR} \\
\cline{3-11}          &       & +UT   & +DRP  & +EIE  & +UT   & +DRP  & +EIE  & +UT   & +DRP  & +EIE \\
    \midrule
    \multirow{2}[1]{*}{MOP1} & mean  & 0.661338(3)- & 0.662155(2)- & \cellcolor[rgb]{ .651,  .651,  .651}0.70003(1) & 0.699331(2)- & 0.695083(3)- & \cellcolor[rgb]{ .651,  .651,  .651}0.703357(1) & 0.674638(2)- & 0.674533(3)- & \cellcolor[rgb]{ .651,  .651,  .651}0.702499(1) \\
          & std.  & 0.00813322 & 0.00648597 & 0.00499976 & 0.000359817 & 0.00128637 & 0.00101947 & 0.00503542 & 0.00541878 & 0.00311904 \\
    \multirow{2}[0]{*}{MOP2} & mean  & 0.388883(3)- & 0.390262(2)- & \cellcolor[rgb]{ .651,  .651,  .651}0.419003(1) & 0.414538(2)- & 0.408293(3)- & \cellcolor[rgb]{ .651,  .651,  .651}0.419059(1) & 0.388444(2)- & 0.388199(3)- & \cellcolor[rgb]{ .651,  .651,  .651}0.420289(1) \\
          & std.  & 0.0114188 & 0.0105765 & 0.000967767 & 0.00034979 & 0.00462227 & 0.000718643 & 0.00621261 & 0.00618729 & 0.000924814 \\
    \multirow{2}[0]{*}{MOP3} & mean  & 0.612541(2)- & 0.606856(3)- & \cellcolor[rgb]{ .651,  .651,  .651}0.704328(1) & 0.0727428(2)- & 0(3)- & \cellcolor[rgb]{ .651,  .651,  .651}0.698207(1) & \cellcolor[rgb]{ .651,  .651,  .651}0.704931(1)+ & 0.704929(2)+ & 0.704918(3) \\
          & std.  & 0.104971 & 0.117036 & 0.000691952 & 0.107161 & 0     & 0.00242507 & 7.44589e-06 & 8.32888e-06 & 1.04308e-05 \\
    \multirow{2}[0]{*}{MOP4} & mean  & 0.381076(2)- & 0.340012(3)- & \cellcolor[rgb]{ .651,  .651,  .651}0.666302(1) & 0.635175(2)- & 0.333056(3)- & \cellcolor[rgb]{ .651,  .651,  .651}0.66452(1) & 0.664027(2)= & 0.663964(3)= & \cellcolor[rgb]{ .651,  .651,  .651}0.671117(1) \\
          & std.  & 0.0303812 & 0.0501843 & 0.00935514 & 0.0208787 & 0.0764859 & 0.00419556 & 0.040519 & 0.0405182 & 0.00102382 \\
    \multirow{2}[0]{*}{MOP5} & mean  & \cellcolor[rgb]{ .651,  .651,  .651}1.00561(1)+ & 1.00387(2)+ & 0.995271(3) & 0.983841(2)- & 0.656023(3)- & \cellcolor[rgb]{ .651,  .651,  .651}1.03585(1) & 1.02959(2)- & 1.02957(3)- & \cellcolor[rgb]{ .651,  .651,  .651}1.03697(1) \\
          & std.  & 0.00516227 & 0.00562675 & 0.00880108 & 0.046519 & 0.0819235 & 0.00218626 & 0.00190153 & 0.00184059 & 0.00138483 \\
    \multirow{2}[0]{*}{MOP6} & mean  & 0.384575(2)- & 0.382319(3)- & \cellcolor[rgb]{ .651,  .651,  .651}0.411897(1) & 0.392023(2)- & 0.373748(3)- & \cellcolor[rgb]{ .651,  .651,  .651}0.400834(1) & 0.389321(2)- & 0.389134(3)- & \cellcolor[rgb]{ .651,  .651,  .651}0.415089(1) \\
          & std.  & 0.00415776 & 0.00515698 & 0.00514564 & 0.00170185 & 0.00463307 & 0.00398384 & 0.00416917 & 0.00335016 & 0.00425518 \\
    \multirow{2}[0]{*}{MOP7} & mean  & 1.003(2)= & 0.997273(3)- & \cellcolor[rgb]{ .651,  .651,  .651}1.00484(1) & 1.02458(2)- & 1.02004(3)- & \cellcolor[rgb]{ .651,  .651,  .651}1.03195(1) & 1.01569(2)- & 1.01569(3)- & \cellcolor[rgb]{ .651,  .651,  .651}1.03945(1) \\
          & std.  & 0.00472392 & 0.00705354 & 0.00425867 & 0.00423842 & 0.00419266 & 0.00170743 & 0.00309338 & 0.00309338 & 3.88267e-05 \\
    \multirow{2}[0]{*}{MOP8} & mean  & 0.639221(3)- & 0.646782(2)- & \cellcolor[rgb]{ .651,  .651,  .651}0.665932(1) & 0.617719(3)- & 0.622416(2)- & \cellcolor[rgb]{ .651,  .651,  .651}0.660511(1) & 0.629534(3)- & 0.630466(2)- & \cellcolor[rgb]{ .651,  .651,  .651}0.667899(1) \\
          & std.  & 0.0187704 & 0.0138775 & 0.00709498 & 0.0111342 & 0.0146644 & 0.00117616 & 0.0175679 & 0.0180245 & 0.00681087 \\
    \multirow{2}[0]{*}{MOP9} & mean  & 0.941823(3)- & 0.943203(2)- & \cellcolor[rgb]{ .651,  .651,  .651}1.0153(1) & 1.00994(2)- & 1.00858(3)- & \cellcolor[rgb]{ .651,  .651,  .651}1.02712(1) & 0.958525(2)- & 0.954476(3)- & \cellcolor[rgb]{ .651,  .651,  .651}1.03876(1) \\
          & std.  & 0.0105386 & 0.0070373 & 0.0309005 & 0.0046247 & 0.00572818 & 0.0034667 & 0.00643967 & 0.0209611 & 7.86262e-05 \\
    \multirow{2}[0]{*}{MOP10} & mean  & 0.574786(3)- & 0.579423(2)- & \cellcolor[rgb]{ .651,  .651,  .651}0.633353(1) & 0.628987(2)- & 0.621131(3)- & \cellcolor[rgb]{ .651,  .651,  .651}0.640441(1) & 0.585433(3)- & 0.585601(2)- & \cellcolor[rgb]{ .651,  .651,  .651}0.642445(1) \\
          & std.  & 0.0106643 & 0.0115775 & 0.00674937 & 0.00582353 & 0.00648545 & 0.00247657 & 0.0160573 & 0.0160362 & 0.00527604 \\
    \multirow{2}[0]{*}{MOP11} & mean  & 0(2)- & 0(3)- & \cellcolor[rgb]{ .651,  .651,  .651}1.11203(1) & 0(2)- & 0(3)- & \cellcolor[rgb]{ .651,  .651,  .651}1.2288(1) & 1.09419(2)- & 1.03151(3)- & \cellcolor[rgb]{ .651,  .651,  .651}1.24029(1) \\
          & std.  & 0     & 0     & 0.14117 & 0     & 0     & 0.00608648 & 0.149089 & 0.181372 & 0.0210573 \\
    \multirow{2}[0]{*}{MOP12} & mean  & 0.529451(2)- & 0.512875(3)- & \cellcolor[rgb]{ .651,  .651,  .651}0.676807(1) & 0.128397(2)- & 0.015963(3)- & \cellcolor[rgb]{ .651,  .651,  .651}0.667576(1) & 0.641745(2)- & 0.639676(3)- & \cellcolor[rgb]{ .651,  .651,  .651}0.743591(1) \\
          & std.  & 0.00923809 & 0.0184806 & 0.0181651 & 0.0386494 & 0.00749498 & 0.0336248 & 0.0191667 & 0.0192837 & 0.0140262 \\
    \multirow{2}[0]{*}{MOP13} & mean  & \cellcolor[rgb]{ .651,  .651,  .651}1.2455(1)+ & 1.23142(2)+ & 1.21589(3) & 1.22652(3)- & 1.23504(2)- & \cellcolor[rgb]{ .651,  .651,  .651}1.3121(1) & 1.2959(2)- & 1.29516(3)- & \cellcolor[rgb]{ .651,  .651,  .651}1.31467(1) \\
          & std.  & 0.00236516 & 0.0039056 & 0.00341321 & 0.0135579 & 0.0127939 & 0.000843473 & 0.0016324 & 0.00179619 & 0.000158645 \\
    \multirow{2}[0]{*}{MOP14} & mean  & 0.979906(3)- & 0.985026(2)- & \cellcolor[rgb]{ .651,  .651,  .651}1.00933(1) & 0.731809(3)- & 0.750485(2)- & \cellcolor[rgb]{ .651,  .651,  .651}1.01344(1) & 1.02228(2)- & 1.02043(3)- & \cellcolor[rgb]{ .651,  .651,  .651}1.03323(1) \\
          & std.  & 0.024034 & 0.022033 & 0.00482035 & 0.029066 & 0.0330712 & 0.00311727 & 0.00664365 & 0.00746083 & 0.00216258 \\
    \multirow{2}[0]{*}{MOP15} & mean  & 1.20199(3)- & 1.20675(2)- & \cellcolor[rgb]{ .651,  .651,  .651}1.27232(1) & 1.08087(3)- & 1.1232(2)- & \cellcolor[rgb]{ .651,  .651,  .651}1.30693(1) & 1.24572(3)- & 1.24574(2)- & \cellcolor[rgb]{ .651,  .651,  .651}1.3061(1) \\
          & std.  & 0.00846114 & 0.00970293 & 0.00560318 & 0.0458696 & 0.0127984 & 0.0028307 & 0.00436022 & 0.00364543 & 0.00110377 \\
    \multirow{2}[1]{*}{MOP16} & mean  & 0.889734(2)- & 0.869224(3)- & \cellcolor[rgb]{ .651,  .651,  .651}0.925115(1) & 0.861506(2)- & 0.764427(3)- & \cellcolor[rgb]{ .651,  .651,  .651}0.924206(1) & 0.934658(3)- & 0.936389(2)- & \cellcolor[rgb]{ .651,  .651,  .651}0.960696(1) \\
          & std.  & 0.00717994 & 0.0194271 & 0.00758356 & 0.0178004 & 0.0238714 & 0.00990376 & 0.00711057 & 0.00898316 & 0.00682911 \\
    \midrule
    \multicolumn{2}{c||}{Total +/=/-} & 2/1/13 & 2/0/14 & \textbackslash{} & 0/0/16 & 0/0/16 & \textbackslash{} & 1/1/14 & 1/1/14 & \textbackslash{} \\
    \multicolumn{2}{c||}{Average rank} & 2.3125(2) & 2.4375(3) & 1.25(1) & 2.25(2) & 2.75(3) & 1(1)  & 2.1875(2) & 2.6875(3) & 1.125(1) \\
    \bottomrule
    \end{tabular}%
    }
    }
  \label{tab:vs_pop_hv}%
\end{table*}%

\begin{table*}[ht]
\footnotesize
  \centering
  \caption{Comparisons of $\operatorname{HV}$ metric values between EIE$^*$ and EIE.}
    \renewcommand{\arraystretch}{1.2}
    \scalebox{1}{
	\setlength{\tabcolsep}{0.8mm}{
    \begin{tabular}{cc||cccc|ccccc|cccc}
    \toprule
    \multirow{2}[1]{*}{Instance} & \multirow{2}[1]{*}{$\operatorname{HV}$} & \multicolumn{3}{c}{PMEA} &       &       & \multicolumn{3}{c}{gMOEA/D-GGR} &       &       & \multicolumn{3}{c}{HVCTR} \\
\cline{3-5}\cline{8-10}\cline{13-15}          &       & +EIE$^*$ & +EIE  & $\Delta$ &       &       & +EIE$^*$ & +EIE  & $\Delta$ &       &       & +EIE$^*$ & +EIE  & $\Delta$ \\
    \midrule
    \multirow{2}[1]{*}{MOP1} & mean  & 0.6681(2)- & \cellcolor[rgb]{ .651,  .651,  .651}0.70003(1) & 0.0319295 &       &       & 0.696709(2)- & \cellcolor[rgb]{ .651,  .651,  .651}0.703357(1) & 0.00664799 &       &       & 0.686139(2)- & \cellcolor[rgb]{ .651,  .651,  .651}0.702499(1) & 0.0163607 \\
          & std.  & 0.00864565 & 0.00499976 &       &       &       & 0.0019781 & 0.00101947 &       &       &       & 0.00753357 & 0.00311904 &  \\
    \multirow{2}[0]{*}{MOP2} & mean  & 0.394785(2)- & \cellcolor[rgb]{ .651,  .651,  .651}0.419003(1) & 0.0242185 &       &       & 0.41599(2)- & \cellcolor[rgb]{ .651,  .651,  .651}0.419059(1) & 0.00306934 &       &       & 0.407611(2)- & \cellcolor[rgb]{ .651,  .651,  .651}0.420289(1) & 0.0126782 \\
          & std.  & 0.00962395 & 0.000967767 &       &       &       & 0.00252522 & 0.000718643 &       &       &       & 0.00739978 & 0.000924814 &  \\
    \multirow{2}[0]{*}{MOP4} & mean  & 0.613836(2)- & \cellcolor[rgb]{ .651,  .651,  .651}0.666302(1) & 0.0524656 &       &       & 0.652296(2)- & \cellcolor[rgb]{ .651,  .651,  .651}0.66452(1) & 0.0122231 &       &       & 0.671046(2)= & \cellcolor[rgb]{ .651,  .651,  .651}0.671117(1) & 7.07239e-05 \\
          & std.  & 0.0629681 & 0.00935514 &       &       &       & 0.0210578 & 0.00419556 &       &       &       & 0.000682803 & 0.00102382 &  \\
    \multirow{2}[0]{*}{MOP10} & mean  & 0.599761(2)- & \cellcolor[rgb]{ .651,  .651,  .651}0.633353(1) & 0.033592 &       &       & 0.637651(2)- & \cellcolor[rgb]{ .651,  .651,  .651}0.640441(1) & 0.00278987 &       &       & 0.609034(2)- & \cellcolor[rgb]{ .651,  .651,  .651}0.642445(1) & 0.033411 \\
          & std.  & 0.0108764 & 0.00674937 &       &       &       & 0.00450336 & 0.00247657 &       &       &       & 0.012924 & 0.00527604 &  \\
    \multirow{2}[1]{*}{MOP16} & mean  & 0.858169(2)- & \cellcolor[rgb]{ .651,  .651,  .651}0.925115(1) & 0.066946 &       &       & 0.876338(2)- & \cellcolor[rgb]{ .651,  .651,  .651}0.924206(1) & 0.0478676 &       &       & 0.858495(2)- & \cellcolor[rgb]{ .651,  .651,  .651}0.960696(1) & 0.1022 \\
          & std.  & 0.01174 & 0.00758356 &       &       &       & 0.0131022 & 0.00990376 &       &       &       & 0.0148161 & 0.00682911 &  \\
    \midrule
    \multicolumn{2}{c||}{Total +/=/-} & 0/0/5 & \textbackslash{} &       &       &       & 0/0/5 & \textbackslash{} &       &       &       & 0/1/4 & \textbackslash{} &  \\
    \multicolumn{2}{c||}{Average rank} & 2(2)  & 1(1)  &       &       &       & 2(2)  & 1(1)  &       &       &       & 2(2)  & 1(1)  &  \\
    \midrule
    \multirow{2}[1]{*}{MOP3} & mean  & \cellcolor[rgb]{ .651,  .651,  .651}0.704454(1)= & 0.704328(2) & -0.000126149 &       &       & \cellcolor[rgb]{ .651,  .651,  .651}0.698419(1)= & 0.698207(2) & -0.000212194 &       &       & 0.704918(2)= & \cellcolor[rgb]{ .651,  .651,  .651}0.704918(1) & 1.36437e-07 \\
          & std.  & 0.000129225 & 0.000691952 &       &       &       & 0.0019649 & 0.00242507 &       &       &       & 1.20586e-05 & 1.04308e-05 &  \\
    \multirow{2}[0]{*}{MOP5} & mean  & \cellcolor[rgb]{ .651,  .651,  .651}0.998153(1)= & 0.995271(2) & -0.00288241 &       &       & 1.03527(2)= & \cellcolor[rgb]{ .651,  .651,  .651}1.03585(1) & 0.000571874 &       &       & 1.03621(2)= & \cellcolor[rgb]{ .651,  .651,  .651}1.03697(1) & 0.000762594 \\
          & std.  & 0.0106517 & 0.00880108 &       &       &       & 0.00207042 & 0.00218626 &       &       &       & 0.00301269 & 0.00138483 &  \\
    \multirow{2}[0]{*}{MOP6} & mean  & 0.41071(2)= & \cellcolor[rgb]{ .651,  .651,  .651}0.411897(1) & 0.00118671 &       &       & \cellcolor[rgb]{ .651,  .651,  .651}0.401069(1)= & 0.400834(2) & -0.000235164 &       &       & 0.414851(2)= & \cellcolor[rgb]{ .651,  .651,  .651}0.415089(1) & 0.000238186 \\
          & std.  & 0.00532505 & 0.00514564 &       &       &       & 0.00445304 & 0.00398384 &       &       &       & 0.00437633 & 0.00425518 &  \\
    \multirow{2}[0]{*}{MOP7} & mean  & \cellcolor[rgb]{ .651,  .651,  .651}1.00796(1)= & 1.00484(2) & -0.00311635 &       &       & 1.02908(2)- & \cellcolor[rgb]{ .651,  .651,  .651}1.03195(1) & 0.00286874 &       &       & \cellcolor[rgb]{ .651,  .651,  .651}1.0395(1)+ & 1.03945(2) & -5.85275e-05 \\
          & std.  & 0.0107312 & 0.00425867 &       &       &       & 0.00171227 & 0.00170743 &       &       &       & 0.000732674 & 3.88267e-05 &  \\
    \multirow{2}[0]{*}{MOP8} & mean  & 0.662865(2)= & \cellcolor[rgb]{ .651,  .651,  .651}0.665932(1) & 0.00306729 &       &       & \cellcolor[rgb]{ .651,  .651,  .651}0.660869(1)= & 0.660511(2) & -0.000358236 &       &       & 0.666475(2)- & \cellcolor[rgb]{ .651,  .651,  .651}0.667899(1) & 0.00142462 \\
          & std.  & 0.0108196 & 0.00709498 &       &       &       & 0.00123665 & 0.00117616 &       &       &       & 0.00607564 & 0.00681087 &  \\
    \multirow{2}[0]{*}{MOP9} & mean  & \cellcolor[rgb]{ .651,  .651,  .651}1.02766(1)+ & 1.0153(2) & -0.0123606 &       &       & 1.02325(2)- & \cellcolor[rgb]{ .651,  .651,  .651}1.02712(1) & 0.00386501 &       &       & \cellcolor[rgb]{ .651,  .651,  .651}1.0391(1)+ & 1.03876(2) & -0.000338182 \\
          & std.  & 0.00289698 & 0.0309005 &       &       &       & 0.00579739 & 0.0034667 &       &       &       & 9.59875e-05 & 7.86262e-05 &  \\
    \multirow{2}[0]{*}{MOP11} & mean  & \cellcolor[rgb]{ .651,  .651,  .651}1.12861(1)= & 1.11203(2) & -0.0165841 &       &       & \cellcolor[rgb]{ .651,  .651,  .651}1.22997(1)+ & 1.2288(2) & -0.00116992 &       &       & \cellcolor[rgb]{ .651,  .651,  .651}1.24405(1)= & 1.24029(2) & -0.00376436 \\
          & std.  & 0.145639 & 0.14117 &       &       &       & 0.0509118 & 0.00608648 &       &       &       & 0.00806247 & 0.0210573 &  \\
    \multirow{2}[0]{*}{MOP12} & mean  & \cellcolor[rgb]{ .651,  .651,  .651}0.679736(1)= & 0.676807(2) & -0.00292854 &       &       & \cellcolor[rgb]{ .651,  .651,  .651}0.676005(1)= & 0.667576(2) & -0.00842958 &       &       & 0.742191(2)= & \cellcolor[rgb]{ .651,  .651,  .651}0.743591(1) & 0.00139938 \\
          & std.  & 0.0205778 & 0.0181651 &       &       &       & 0.0325222 & 0.0336248 &       &       &       & 0.0157187 & 0.0140262 &  \\
    \multirow{2}[0]{*}{MOP13} & mean  & \cellcolor[rgb]{ .651,  .651,  .651}1.27232(1)+ & 1.21589(2) & -0.0564329 &       &       & 1.31144(2)- & \cellcolor[rgb]{ .651,  .651,  .651}1.3121(1) & 0.000655162 &       &       & \cellcolor[rgb]{ .651,  .651,  .651}1.31521(1)+ & 1.31467(2) & -0.000540711 \\
          & std.  & 0.0149524 & 0.00341321 &       &       &       & 0.00114624 & 0.000843473 &       &       &       & 0.000310166 & 0.000158645 &  \\
    \multirow{2}[0]{*}{MOP14} & mean  & \cellcolor[rgb]{ .651,  .651,  .651}1.01182(1)+ & 1.00933(2) & -0.00249291 &       &       & 1.0133(2)= & \cellcolor[rgb]{ .651,  .651,  .651}1.01344(1) & 0.000140617 &       &       & \cellcolor[rgb]{ .651,  .651,  .651}1.03413(1)= & 1.03323(2) & -0.000892536 \\
          & std.  & 0.00417508 & 0.00482035 &       &       &       & 0.00272425 & 0.00311727 &       &       &       & 0.00145563 & 0.00216258 &  \\
    \multirow{2}[1]{*}{MOP15} & mean  & \cellcolor[rgb]{ .651,  .651,  .651}1.27928(1)+ & 1.27232(2) & -0.00695646 &       &       & 1.30509(2)- & \cellcolor[rgb]{ .651,  .651,  .651}1.30693(1) & 0.00183797 &       &       & \cellcolor[rgb]{ .651,  .651,  .651}1.307(1)+ & 1.3061(2) & -0.000899129 \\
          & std.  & 0.00783145 & 0.00560318 &       &       &       & 0.00253987 & 0.0028307 &       &       &       & 0.00174716 & 0.00110377 &  \\
    \midrule
    \multicolumn{2}{c||}{Total +/=/-} & 4/7/0 & \textbackslash{} &       &       &       & 1/6/4 & \textbackslash{} &       &       &       & 4/6/1 & \textbackslash{} &  \\
    \multicolumn{2}{c||}{Average rank} & 1.1818(1) & 1.8182(2) &       &       &       & 1.5455(2) & 1.4545(1) &       &       &       & 1.4545(1) & 1.5455(2) &  \\
    \bottomrule
    \end{tabular}%
    }
    }
  \label{tab:ews_hv}%
\end{table*}%

\subsection{Impact of Extreme Weighted Sum (EWS) Method}
We investigate the use of the EWS method by comparing it with optimizing the objectives separately. This variant is denoted as EIE$^*$.
According to the shape of the feasible objective region (controlled by $\Theta$), we divide the test instances into two parts: MOP1, MOP2, MOP4, MOP10, and MOP16 (\ie, multiple optimal solutions for some objectives); MOP3, MOP6--MOP9, and MOP11--MOP15 (\ie, a unique optimal solution for any objective). EIE$^*$ is more likely to contribute dominance-resistant solutions to the population when facing the first group.

The results are displayed in Table~\ref{tab:ews_hv}. In the first group of instances, EIE significantly outperforms EIE$^*$. Solutions generated by EIE or EIE$^*$ can significantly impact the performance of MOEAs. Only suitable solutions should be given higher preference for contributing to the population. \figurename~\ref{fig:ndobjs_ews} depicts the non-dominated set of the final population. We can see that MOEAs with EIE$^*$ all maintain dominance-resistant solutions. For example, PMEA with EIE$^*$ has solutions that have inferior values of $f_1$.
In the second group of instances, the performance of EIE is only slightly worse than that of EIE$^*$ when applied to PMEA and HVCTR. EIE exhibits statistically comparable performance to EIE$^*$ on more than half the instances. Besides, we count the performance gaps greater than $10^{-2}$ ($\Delta>10^{-2}$) in the first group: $5/5$ for PMEA; $2/5$ for gMOEA/D-GGR; $4/5$ for HVCTR. In contrast, the number of instances satisfying $\Delta<-10^{-2}$ in the second group is $3/11$ for PMEA, $0/11$ for gMOEA/D-GGR, and $0/11$ for HVCTR. The slight performance decline in the second group brings about a significant performance boost in the first group.

We utilize \figurename~\ref{fig:ndobjs_ews_second} that displays the final non-dominated sets to explain the performance degradation. We can find that the population spread achieved by MOEA with EIE is marginally inferior to that obtained by MOEA with EIE$^*$. In other words, MOEA with EIE may miss some boundary objective vectors. 
Nevertheless, we can employ a smaller value of $\epsilon_i$ to preserve more boundary objective vectors, thereby mitigating performance degradation on the instance of the second group, as validated in Table~\ref{tab:ews_hv_small_eps_2}. Specifically, a smaller value of $\epsilon_i$ makes PMEA with EIE have better performance in the second group. gMOEA/D-GGR with EIE (0.05) statistically outperforms that with EIE$^*$ on more instances but has a worse average rank. HVCTR with EIE ($<$0.05) generally outperforms that with EIE$^*$, while HVCTR with EIE (0.05) does not. On the downside, a smaller value of $\epsilon_i$ results in consistent performance deterioration in the first group, as shown in Table~\ref{tab:ews_hv_small_eps_1}. It is technically possible to tune $\epsilon_i$ dynamically~\cite{li2024survey}, which is an interesting future work.

This section confirms the reasonableness and robustness of using the EWS method. The negative effect of dominance-resistant solutions is also validated.

\begin{table*}[ht]
\footnotesize
  \centering
  \caption{Comparisons of $\operatorname{HV}$ metric values between EIE$^*$ and EIE with different settings of $\epsilon_i$ (first group). $\epsilon_i=0.05$ for $i=1,\ldots,m$ are the original setting of EIE.}
    \renewcommand{\arraystretch}{1.2}
    \scalebox{1}{
	\setlength{\tabcolsep}{2.2mm}{
    \begin{tabular}{c||ccccc|cccccc|ccccc}
    \toprule
    \multirow{2}[1]{*}{$\operatorname{HV}$ (first group)} & \multicolumn{4}{c}{PMEA}      &       &       & \multicolumn{4}{c}{gMOEA/D-GGR} &       &       & \multicolumn{4}{c}{HVCTR} \\
\cline{2-5}\cline{8-11}\cline{14-17}          & +EIE$^*$ & 0.005 & 0.01  & 0.05  &       &       & +EIE$^*$ & 0.005 & 0.01  & 0.05  &       &       & +EIE$^*$ & 0.005 & 0.01  & 0.05 \\
    \midrule
    \multirow{4}[2]{*}{Total +/=/-} & \textbackslash{} & 5/0/0 & 5/0/0 & 5/0/0 &       &       & \textbackslash{} & 5/0/0 & 5/0/0 & 5/0/0 &       &       & \textbackslash{} & 4/1/0 & 4/1/0 & 4/1/0 \\
          & 0/0/5 & \textbackslash{} & 2/3/0 & 5/0/0 &       &       & 0/0/5 & \textbackslash{} & 1/4/0 & 4/0/1 &       &       & 0/1/4 & \textbackslash{} & 4/1/0 & 4/1/0 \\
          & 0/0/5 & 0/3/2 & \textbackslash{} & 3/2/0 &       &       & 0/0/5 & 0/4/1 & \textbackslash{} & 1/3/1 &       &       & 0/1/4 & 0/1/4 & \textbackslash{} & 4/1/0 \\
          & 0/0/5 & 0/0/5 & 0/2/3 & \textbackslash{} &       &       & 0/0/5 & 1/0/4 & 1/3/1 & \textbackslash{} &       &       & 0/1/4 & 0/1/4 & 0/1/4 & \textbackslash{} \\
    \midrule
    Average rank & 4(4)  & 3(3)  & \cellcolor[rgb]{ .749,  .749,  .749}1.8(2) & \cellcolor[rgb]{ .502,  .502,  .502}1.2(1) &       &       & 4(4)  & 2.6(3) & \cellcolor[rgb]{ .749,  .749,  .749}2(2) & \cellcolor[rgb]{ .502,  .502,  .502}1.4(1) &       &       & 3.6(4) & 3(3)  & \cellcolor[rgb]{ .749,  .749,  .749}2.4(2) & \cellcolor[rgb]{ .502,  .502,  .502}1(1) \\
    \bottomrule
    \end{tabular}%
    }
    }
  \label{tab:ews_hv_small_eps_1}%
\end{table*}%

\begin{table*}[ht]
\footnotesize
  \centering
  \caption{Comparisons of $\operatorname{HV}$ metric values between EIE$^*$ and EIE with different settings of $\epsilon_i$ (second group). $\epsilon_i=0.05$ for $i=1,\ldots,m$ are the original setting of EIE.}
    \renewcommand{\arraystretch}{1.2}
    \scalebox{1}{
	\setlength{\tabcolsep}{1.6mm}{
    \begin{tabular}{c||ccccc|cccccc|ccccc}
    \toprule
    \multirow{2}[1]{*}{$\operatorname{HV}$ (second group)} & \multicolumn{4}{c}{PMEA}      &       &       & \multicolumn{4}{c}{gMOEA/D-GGR} &       &       & \multicolumn{4}{c}{HVCTR} \\
\cline{2-5}\cline{8-11}\cline{14-17}          & +EIE$^*$ & 0.005 & 0.01  & 0.05  &       &       & +EIE$^*$ & 0.005 & 0.01  & 0.05  &       &       & +EIE$^*$ & 0.005 & 0.01  & 0.05 \\
    \midrule
    \multirow{4}[2]{*}{Total +/=/-} & \textbackslash{} & 0/10/1 & 0/9/2 & 0/7/4 &       &       & \textbackslash{} & 2/8/1 & 1/9/1 & 4/6/1 &       &       & \textbackslash{} & 2/9/0 & 2/9/0 & 1/6/4 \\
          & 1/10/0 & \textbackslash{} & 0/11/0 & 2/7/2 &       &       & 1/8/2 & \textbackslash{} & 1/9/1 & 3/7/1 &       &       & 0/9/2 & \textbackslash{} & 2/8/1 & 1/5/5 \\
          & 2/9/0 & 0/11/0 & \textbackslash{} & 1/7/3 &       &       & 1/9/1 & 1/9/1 & \textbackslash{} & 3/7/1 &       &       & 0/9/2 & 1/8/2 & \textbackslash{} & 0/7/4 \\
          & 4/7/0 & 2/7/2 & 3/7/1 & \textbackslash{} &       &       & 1/6/4 & 1/7/3 & 1/7/3 & \textbackslash{} &       &       & 4/6/1 & 5/5/1 & 4/7/0 & \textbackslash{} \\
    \midrule
    Average rank & \cellcolor[rgb]{ .502,  .502,  .502}2(1) & \cellcolor[rgb]{ .749,  .749,  .749}2.45(2) & 2.55(3) & 3(4)  &       &       & 2.8182(4) & \cellcolor[rgb]{ .502,  .502,  .502}2.27(1) & \cellcolor[rgb]{ .502,  .502,  .502}2.27(1) & \cellcolor[rgb]{ .749,  .749,  .749}2.64(3) &       &       & 2.91(3) & \cellcolor[rgb]{ .749,  .749,  .749}2.18(2) & \cellcolor[rgb]{ .502,  .502,  .502}1.73(1) & 3.18(4) \\
    \bottomrule
    \end{tabular}%
    }
    }
  \label{tab:ews_hv_small_eps_2}%
\end{table*}%

\subsection{Results on Test Problems with Inverted \texorpdfstring{$PF$s}{PFs}}
The MOP11--MOP16 have the $PF$s similar to a regular $2$-dimensional simplex. We invert the $PF$s of these test problems for a more realistic configuration~\cite{ishibuchi2017performance}. Specifically, we slightly modify Eq.~\ref{eqn:pos_fun} and obtain a new position function as follows
\begin{equation}
    h_i^\prime(\mathbf{x}_{\urone} | \mathbf{p},\mathbf{c}^{pos},\gamma) = 1 - y_i(\hat{\mathbf{x}}(\mathbf{x}_{\urone}|\mathbf{c}^{pos},\gamma))^{p_i}.
\end{equation}
We denote these variants of MOP11--MOP16 as MOP11$^{-1}$--MOP16$^{-1}$. To cope with the irregular $PF$, we incorporate a similar weight vector adjustment strategy stated in~\cite{li2020weights} into gMOEA/D-GGR. The new version is referred to as gMOEA/D-GGRAW\footnote{The source code can be found at \url{https://github.com/EricZheng1024/MOEA-D-GGR}.}. The other two algorithms remain unaltered, as they inherently accommodate the irregular $PF$.

In alignment with Section~\ref{ssec:exp_eff}, three MOEAs are compared with their enhanced versions, respectively. The $\operatorname{E}$ metric values are provided in Table~\ref{tab:baseline_e_inv}. From the table, it can be found that the MOEAs with EIE have lower errors across all instances. The statistical results presented in the table indicate that MOEAs with EIE consistently exhibit reduced error rates across all instances. Moreover, the MOEA with EIE reduced the estimation error of the ideal objective vector by more than $10\%$ on most instances. The mean estimated errors are no greater than $5\%$ except for: PMEA with EIE on MOP12$^{-1}$, MOP15$^{-1}$, and MOP16$^{-1}$; gMOEA/D-GGRAW with EIE on MOP16$^{-1}$; HVCTR with EIE on MOP15$^{-1}$ and MOP16$^{-1}$. All algorithms can not perform well on MOP16$^{-1}$ in terms of the $\operatorname{E}$ metric, because MOP16$^{-1}$ has the hardly-dominated boundary~\cite{wang2019scalable}.

We now examine the relationship between the $\operatorname{E}$ metric and the $\operatorname{HV}$ metric. The $\operatorname{HV}$ metric values are presented in Table~\ref{tab:baseline_hv_inv}. Performance degradation is observed only in the result of gMOEA/D-GGRAW with EIE on MOP13$^{-1}$. Nevertheless, the corresponding performance gap is comparatively small ($|\Delta| \approx 10^{-3}$). For all other results, EIE still demonstrates a clear advantage in improving the performance of MOEAs. EIE exhibits a robust capacity to mitigate the impacts of diverse biases.
We also show the final non-dominated sets obtained by algorithms in \figurename~\ref{fig:ndobjs_baseline_inv}. On MOP11$^{-1}$, the populations of PMEA and HVCTR only converge to certain parts of the $PF$, whereas gMOEA/D-GGR achieves sparse boundary objective vectors. On MOP11$^{-1}$, the final populations of PMEA and gMOEA/D-GGR are far away from the $PF$, while that of HVCTR has poor diversity. On MOP15$^{-1}$, the original MOEAs cannot find the boundary of the $PF$. The final population achieves better coverage for the $PF$ as long as EIE is employed.

In summary, the findings align with those concluded in Section~\ref{ssec:exp_eff}: EIE enhances the performance of the MOEA and provides an accurate estimation of the ideal objective vector.

\subsection{Results on \texttt{bbob-biobj} Test Suite}
To evaluate the robustness of EIE across a broader range of problems, we assess its performance using the \texttt{bbob-biobj} test suite~\cite{brockhoff2022using}. This test suite comprises 55 test problems, each characterized by 15 different parameter configurations and 6 alternative numbers of variables. In this experiment, we select the first parameter configuration for each test problem and fix the number of variables at 10. The number of function evaluations is set to 100,000. Tables~\ref{tab:baseline_e_bbob} and~\ref{tab:baseline_hv_bbob} shows summary results. The complete results are available in our code repository. The results indicate that the MOEA augmented with EIE consistently outperforms its original counterpart across most instances.

\newpage
\clearpage

\begin{table*}[t]
\footnotesize
  \centering
  \caption{Comparisons of $\operatorname{E}$ metric values between the MOEA and the MOEA with EIE (MOP11$^{-1}$-MOP16$^{-1}$).}
    \renewcommand{\arraystretch}{1}
	\setlength{\tabcolsep}{1mm}{
    \begin{tabular}{cc||ccc|ccc|ccc}
    \toprule
    Instance & $\operatorname{E}$    & PMEA  & +EIE  & $\Delta$ & gMOEA/D-GGRAW & +EIE  & $\Delta$ & HVCTR & +EIE  & $\Delta$ \\
    \midrule
    \multirow{2}[1]{*}{MOP11$^{-1}$} & mean  & 2.8433(2)- & \cellcolor[rgb]{ .651,  .651,  .651}0.019445(1) & -2.8238 & 1.5101(2)- & \cellcolor[rgb]{ .651,  .651,  .651}0.0025932(1) & -1.5075 & 0.24008(2)- & \cellcolor[rgb]{ .651,  .651,  .651}0.016706(1) & -0.22337 \\
          & std.  & 0.19459 & 0.011482 &       & 0.75146 & 0.0017677 &       & 0.088242 & 0.0092438 &  \\
    \multirow{2}[0]{*}{MOP12$^{-1}$} & mean  & 0.18549(2)- & \cellcolor[rgb]{ .651,  .651,  .651}0.12967(1) & -0.055825 & 0.068803(2)- & \cellcolor[rgb]{ .651,  .651,  .651}0.016922(1) & -0.051881 & 0.074502(2)- & \cellcolor[rgb]{ .651,  .651,  .651}0.014433(1) & -0.060069 \\
          & std.  & 0.019391 & 0.031568 &       & 0.0092312 & 0.019857 &       & 0.010545 & 0.019553 &  \\
    \multirow{2}[0]{*}{MOP13$^{-1}$} & mean  & 0.26441(2)- & \cellcolor[rgb]{ .651,  .651,  .651}1.399e-12(1) & -0.26441 & 0.0030115(2)- & \cellcolor[rgb]{ .651,  .651,  .651}8.2734e-06(1) & -0.0030032 & 0.21219(2)- & \cellcolor[rgb]{ .651,  .651,  .651}3.5194e-05(1) & -0.21215 \\
          & std.  & 0.23988 & 1.2555e-12 &       & 0.010292 & 4.5315e-05 &       & 0.091149 & 2.5847e-05 &  \\
    \multirow{2}[0]{*}{MOP14$^{-1}$} & mean  & 0.30291(2)- & \cellcolor[rgb]{ .651,  .651,  .651}0.014376(1) & -0.28854 & 0.2176(2)- & \cellcolor[rgb]{ .651,  .651,  .651}0.0005033(1) & -0.21709 & 0.11546(2)- & \cellcolor[rgb]{ .651,  .651,  .651}1.0876e-05(1) & -0.11545 \\
          & std.  & 0.035908 & 0.009275 &       & 0.048407 & 0.00085986 &       & 0.057434 & 1.8657e-05 &  \\
    \multirow{2}[0]{*}{MOP15$^{-1}$} & mean  & 0.91199(2)- & \cellcolor[rgb]{ .651,  .651,  .651}0.057673(1) & -0.85431 & 0.73007(2)- & \cellcolor[rgb]{ .651,  .651,  .651}0.045099(1) & -0.68497 & 0.65186(2)- & \cellcolor[rgb]{ .651,  .651,  .651}0.099688(1) & -0.55217 \\
          & std.  & 0.028084 & 0.011911 &       & 0.065423 & 0.010497 &       & 0.03128 & 0.014762 &  \\
    \multirow{2}[1]{*}{MOP16$^{-1}$} & mean  & 0.48979(2)- & \cellcolor[rgb]{ .651,  .651,  .651}0.20477(1) & -0.28502 & 0.32889(2)- & \cellcolor[rgb]{ .651,  .651,  .651}0.19358(1) & -0.13531 & 0.3579(2)- & \cellcolor[rgb]{ .651,  .651,  .651}0.21233(1) & -0.14557 \\
          & std.  & 0.042789 & 0.011908 &       & 0.016811 & 0.0061387 &       & 0.022691 & 0.0056511 &  \\
    \midrule
    \multicolumn{2}{c|}{Total +/=/-} & 0/0/6 & \textbackslash{} &       & 0/0/6 & \textbackslash{} &       & 0/0/6 & \textbackslash{} &  \\
    \multicolumn{2}{c|}{Average rank} & 2(2)  & 1(1)  &       & 2(2)  & 1(1)  &       & 2(2)  & 1(1)  &  \\
    \bottomrule
    \end{tabular}%
    }
  \label{tab:baseline_e_inv}%
\end{table*}%

\begin{table*}[t]
\footnotesize
  \centering
  \caption{Comparisons of $\operatorname{HV}$ metric values between the MOEA and the MOEA with EIE (MOP11$^{-1}$-MOP16$^{-1}$).}
    \renewcommand{\arraystretch}{1}
	\setlength{\tabcolsep}{1mm}{
    \begin{tabular}{cc||ccc|ccc|ccc}
    \toprule
    Instance & $\operatorname{HV}$    & PMEA  & +EIE  & $\Delta$ & gMOEA/D-GGRAW & +EIE  & $\Delta$ & HVCTR & +EIE  & $\Delta$ \\
    \midrule
    \multirow{2}[1]{*}{MOP11$^{-1}$} & mean  & 0(2)- & \cellcolor[rgb]{ .651,  .651,  .651}0.19705(1) & 0.19705 & 0.0064225(2)- & \cellcolor[rgb]{ .651,  .651,  .651}0.19508(1) & 0.18865 & 0.064173(2)- & \cellcolor[rgb]{ .651,  .651,  .651}0.20021(1) & 0.13604 \\
          & std.  & 0     & 0.006386 &       & 0.034706 & 0.0016559 &       & 0.04153 & 0.0026595 &  \\
    \multirow{2}[0]{*}{MOP12$^{-1}$} & mean  & 0.67448(2)- & \cellcolor[rgb]{ .651,  .651,  .651}0.68953(1) & 0.015055 & 0.72816(2)- & \cellcolor[rgb]{ .651,  .651,  .651}0.73479(1) & 0.0066328 & 0.71147(2)- & \cellcolor[rgb]{ .651,  .651,  .651}0.73549(1) & 0.024012 \\
          & std.  & 0.004974 & 0.0071879 &       & 0.0021222 & 0.0022114 &       & 0.0031134 & 0.0030445 &  \\
    \multirow{2}[0]{*}{MOP13$^{-1}$} & mean  & 0.071639(2)- & \cellcolor[rgb]{ .651,  .651,  .651}0.083581(1) & 0.011942 & \cellcolor[rgb]{ .651,  .651,  .651}0.082873(1)+ & 0.081796(2) & -0.0010764 & 0.07548(2)- & \cellcolor[rgb]{ .651,  .651,  .651}0.084959(1) & 0.0094789 \\
          & std.  & 0.010731 & 0.00015229 &       & 0.0012688 & 0.0013598 &       & 0.0039726 & 0.0001472 &  \\
    \multirow{2}[0]{*}{MOP14$^{-1}$} & mean  & 0.39168(2)- & \cellcolor[rgb]{ .651,  .651,  .651}0.42254(1) & 0.030861 & 0.407(2)- & \cellcolor[rgb]{ .651,  .651,  .651}0.42988(1) & 0.022882 & 0.4216(2)- & \cellcolor[rgb]{ .651,  .651,  .651}0.43319(1) & 0.011587 \\
          & std.  & 0.0046547 & 0.0012519 &       & 0.0044818 & 0.0012184 &       & 0.0050412 & 0.0006686 &  \\
    \multirow{2}[0]{*}{MOP15$^{-1}$} & mean  & 0.044023(2)- & \cellcolor[rgb]{ .651,  .651,  .651}0.067133(1) & 0.023109 & 0.050579(2)- & \cellcolor[rgb]{ .651,  .651,  .651}0.065663(1) & 0.015084 & 0.056574(2)- & \cellcolor[rgb]{ .651,  .651,  .651}0.073506(1) & 0.016932 \\
          & std.  & 0.0010718 & 0.00235 &       & 0.0022723 & 0.0028852 &       & 0.0010711 & 0.0018868 &  \\
    \multirow{2}[1]{*}{MOP16$^{-1}$} & mean  & 0.32818(2)- & \cellcolor[rgb]{ .651,  .651,  .651}0.34725(1) & 0.019072 & 0.36909(2)- & \cellcolor[rgb]{ .651,  .651,  .651}0.3767(1) & 0.0076078 & 0.3233(2)- & \cellcolor[rgb]{ .651,  .651,  .651}0.34184(1) & 0.018541 \\
          & std.  & 0.010945 & 0.0056831 &       & 0.0025198 & 0.0031809 &       & 0.008314 & 0.0061374 &  \\
    \midrule
    \multicolumn{2}{c|}{Total +/=/-} & 0/0/6 & \textbackslash{} &       & 1/0/5 & \textbackslash{} &       & 0/0/6 & \textbackslash{} &  \\
    \multicolumn{2}{c|}{Average rank} & 2(2)  & 1(1)  &       & 1.8333(2) & 1.1667(1) &       & 2(2)  & 1(1)  &  \\
    \bottomrule
    \end{tabular}%
    }
  \label{tab:baseline_hv_inv}%
\end{table*}%

\begin{table*}[ht]
\footnotesize
\centering
    \centering  
    \caption{Comparisons of $\operatorname{E}$ metric values between the MOEA and the MOEA with EIE (\texttt{bbob-biobj}).}
    \renewcommand{\arraystretch}{1}
    \setlength{\tabcolsep}{3mm}{
    \begin{tabular}{c|cc|cc|cc}
    \toprule
    $\operatorname{E}$     & PMEA  & +EIE  & \makecell{gMOEA/D-GGR} & +EIE  & HVCTR & +EIE \\
    \midrule
    \makecell{Total +/=/-} & 13/17/25 & \textbackslash{} & 2/9/44 & \textbackslash{} & 3/19/33 & \textbackslash{} \\
    \makecell{Average rank} & 1.5636(2) & 1.4364(1) & 1.8727(2) & 1.1273(1) & 1.7455(2) & 1.2545(1) \\
    \bottomrule
    \end{tabular}%
    }
    \label{tab:baseline_e_bbob}%
\end{table*}%

\begin{table*}[ht]
\footnotesize
    \centering
    \caption{Comparisons of $\operatorname{HV}$ metric values between the MOEA and the MOEA with EIE (\texttt{bbob-biobj}).}
    \renewcommand{\arraystretch}{1}
    \setlength{\tabcolsep}{3mm}{
    \begin{tabular}{c|cc|cc|cc}
    \toprule
    $\operatorname{HV}$     & PMEA  & +EIE  & \makecell{gMOEA/D-GGR} & +EIE  & HVCTR & +EIE \\
    \midrule
    \makecell{Total +/=/-} & 11/20/24 & \textbackslash{} & 0/6/49 & \textbackslash{} & 3/20/32 & \textbackslash{} \\
    \makecell{Average rank} & 1.5455(2) & 1.4545(1) & 1.9818(2) & 1.0182(1) & 1.7455(2) & 1.2545(1) \\
    \bottomrule
    \end{tabular}%
    }
    \label{tab:baseline_hv_bbob}%
\end{table*}%

\begin{figure*}[ht]
\centering
\subfloat[MOP1]{\includegraphics[width=0.245\linewidth]{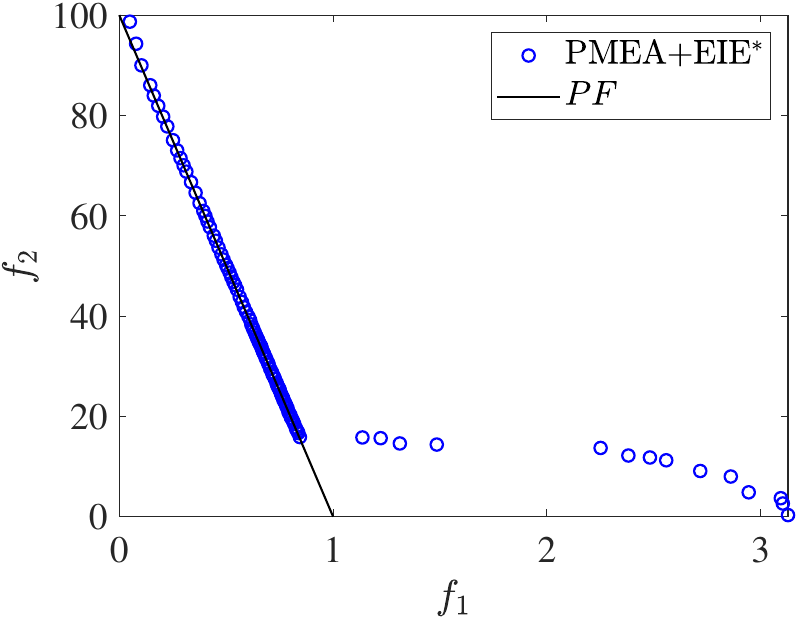}}
\subfloat[MOP1]{\includegraphics[width=0.245\linewidth]{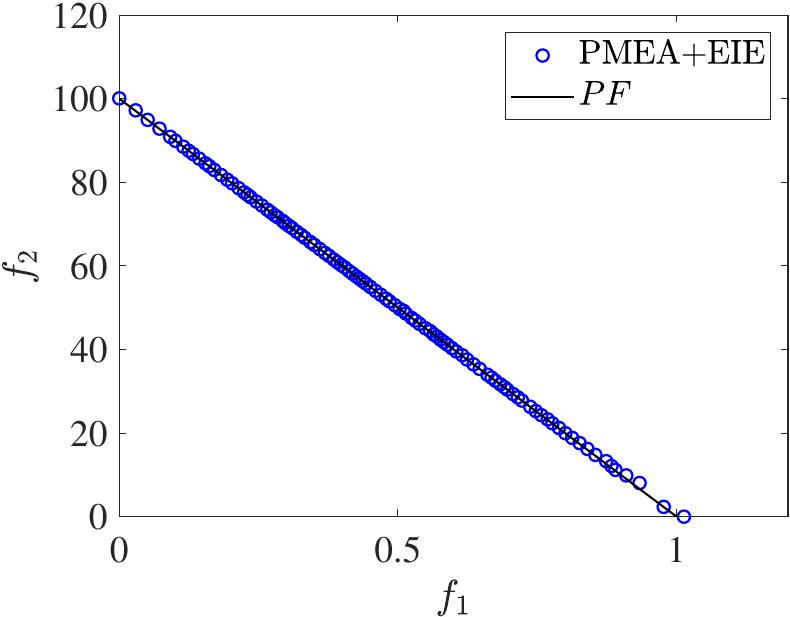}}
\subfloat[MOP10]{\includegraphics[width=0.245\linewidth]{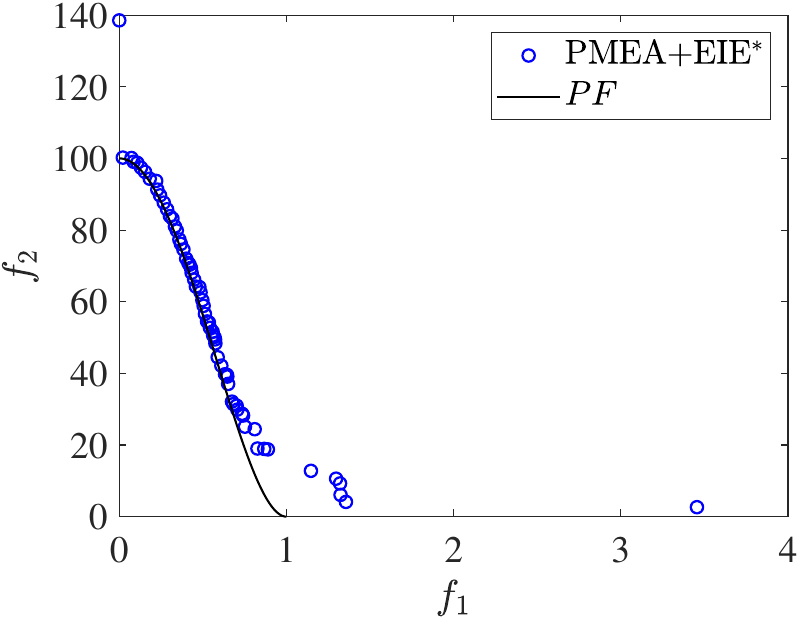}}
\subfloat[MOP10]{\includegraphics[width=0.245\linewidth]{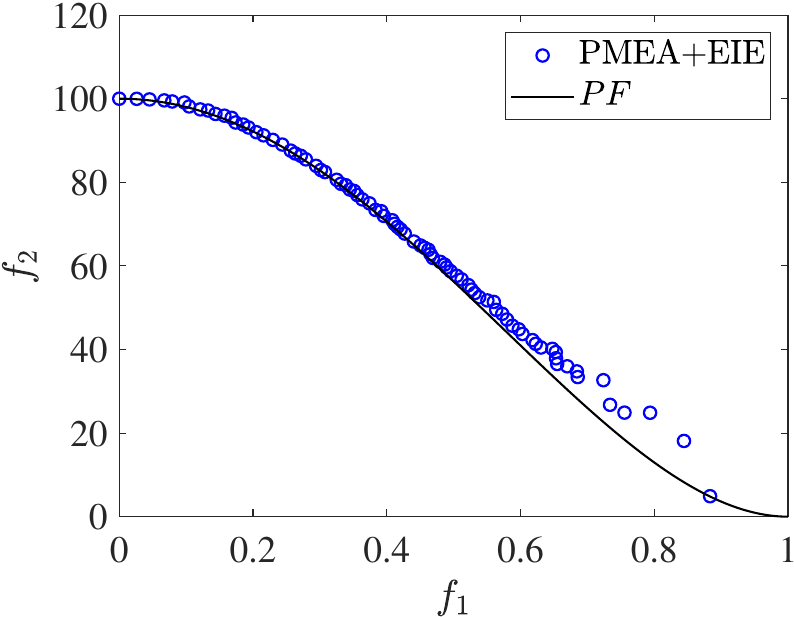}}
\hfil
\subfloat[MOP16]{\includegraphics[width=0.245\linewidth]{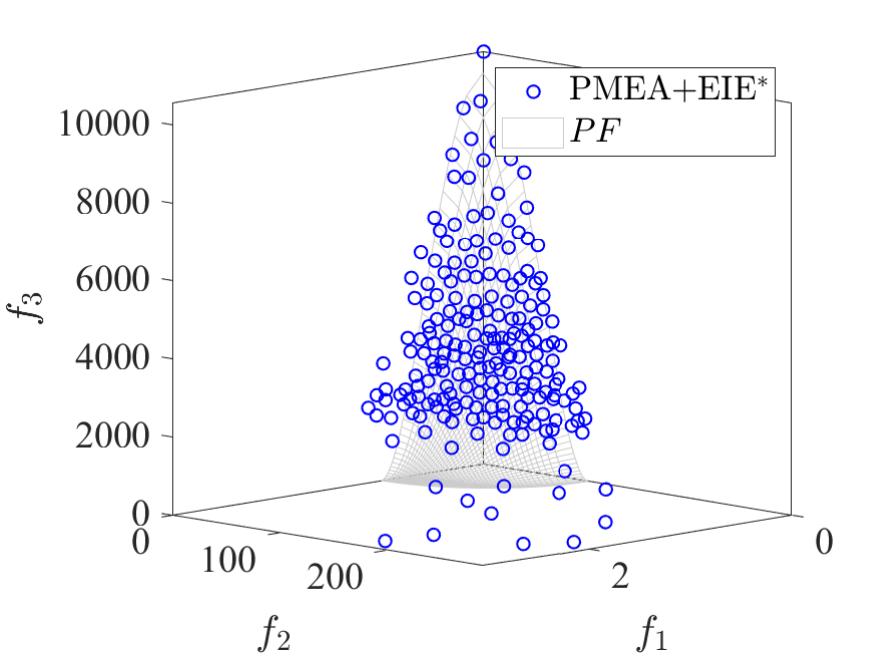}}
\subfloat[MOP16]{\includegraphics[width=0.245\linewidth]{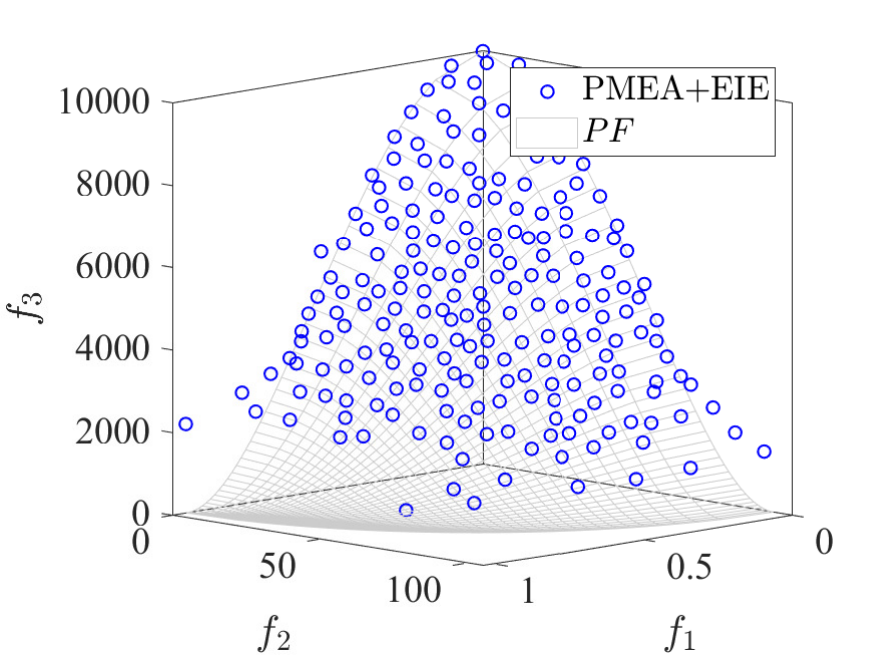}}
\subfloat[MOP1]{\includegraphics[width=0.245\linewidth]{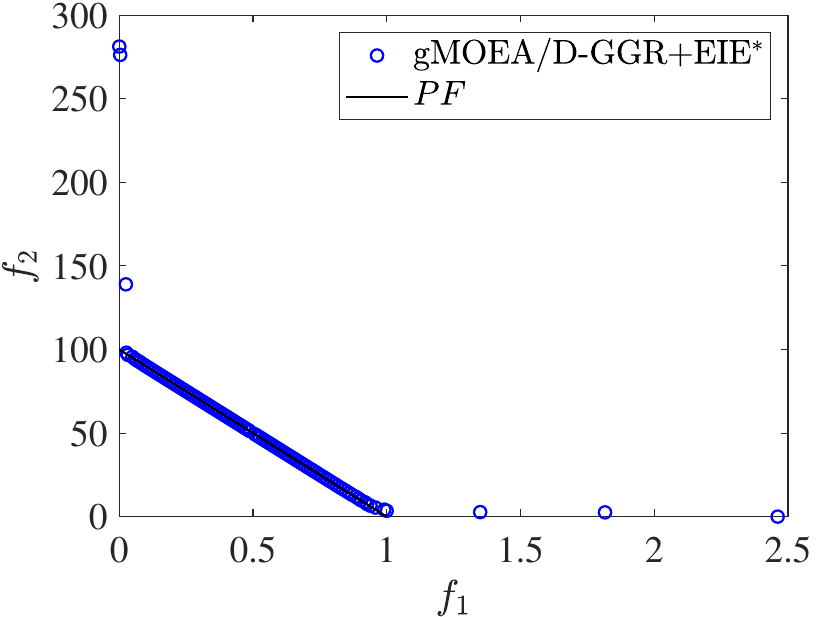}}
\subfloat[MOP1]{\includegraphics[width=0.245\linewidth]{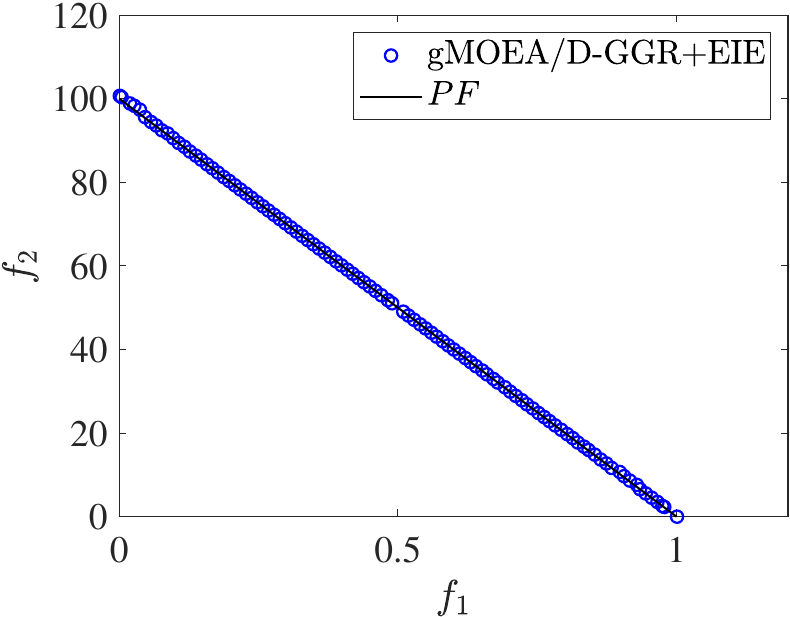}}
\hfil
\subfloat[MOP10]{\includegraphics[width=0.245\linewidth]{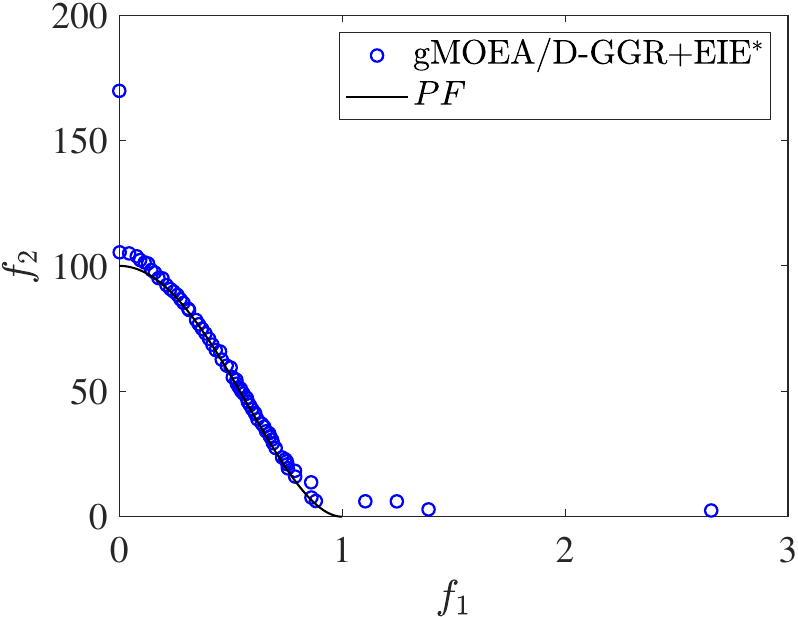}}
\subfloat[MOP10]{\includegraphics[width=0.245\linewidth]{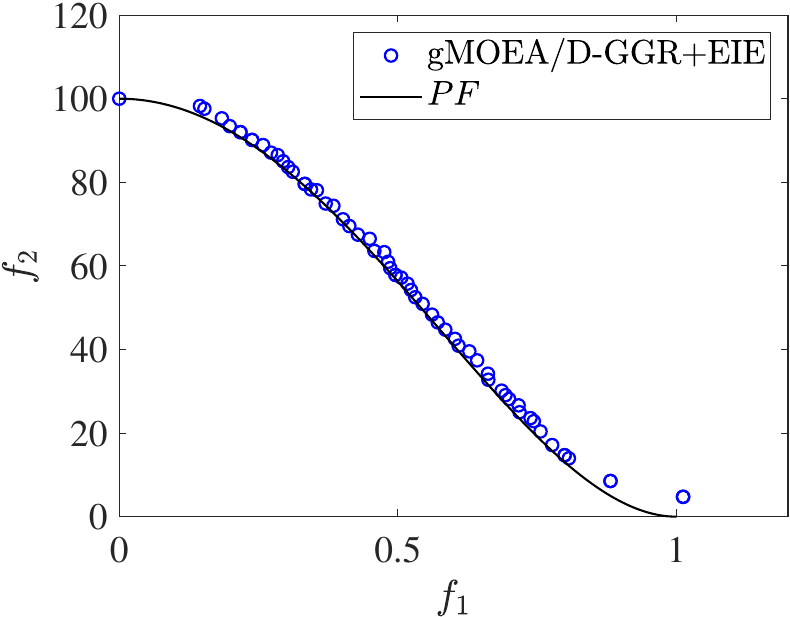}}
\subfloat[MOP16]{\includegraphics[width=0.245\linewidth]{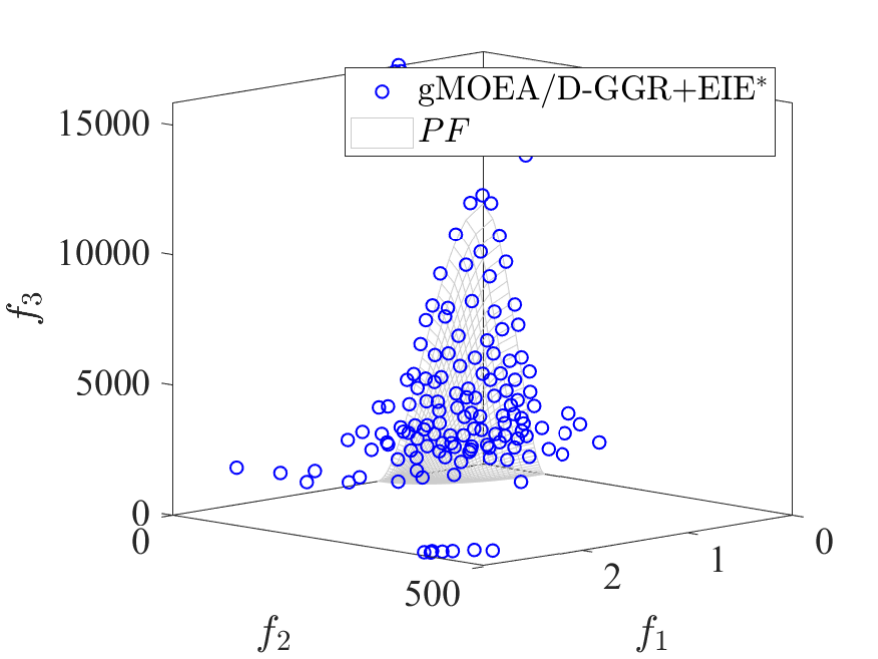}}
\subfloat[MOP16]{\includegraphics[width=0.245\linewidth]{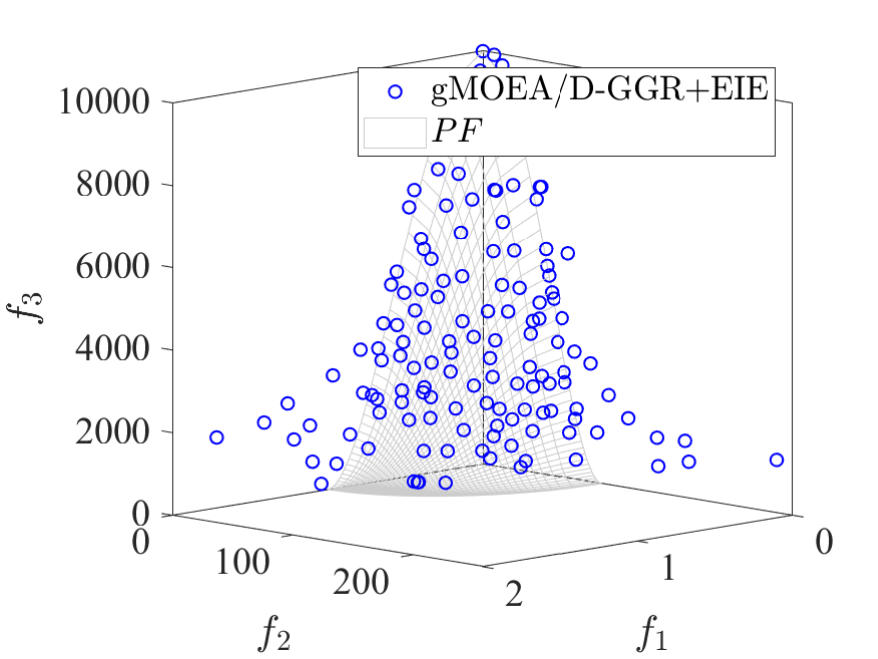}}
\hfil
\subfloat[MOP1]{\includegraphics[width=0.245\linewidth]{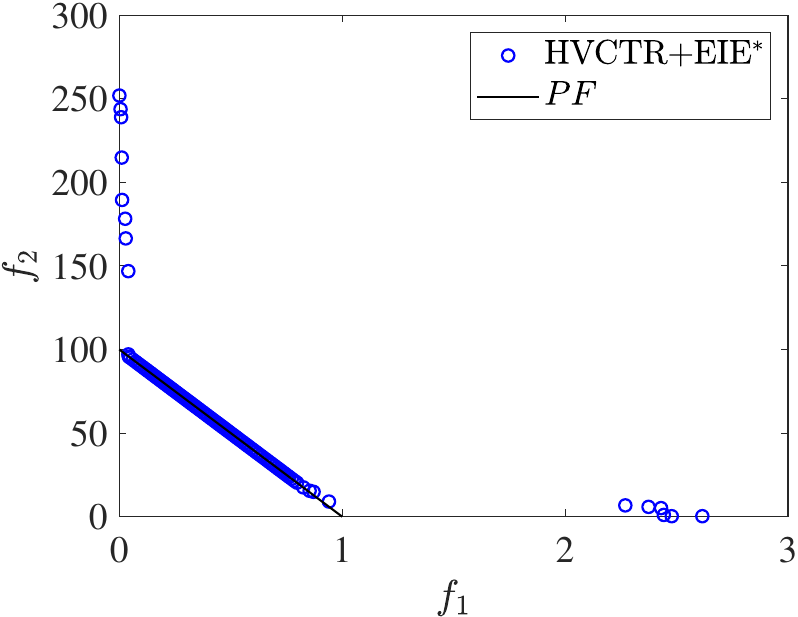}}
\subfloat[MOP1]{\includegraphics[width=0.245\linewidth]{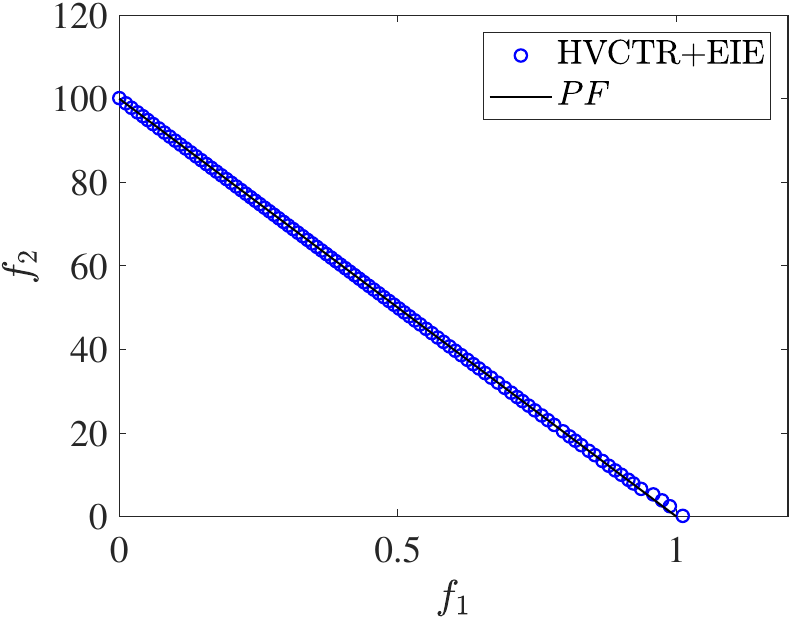}}
\subfloat[MOP2]{\includegraphics[width=0.245\linewidth]{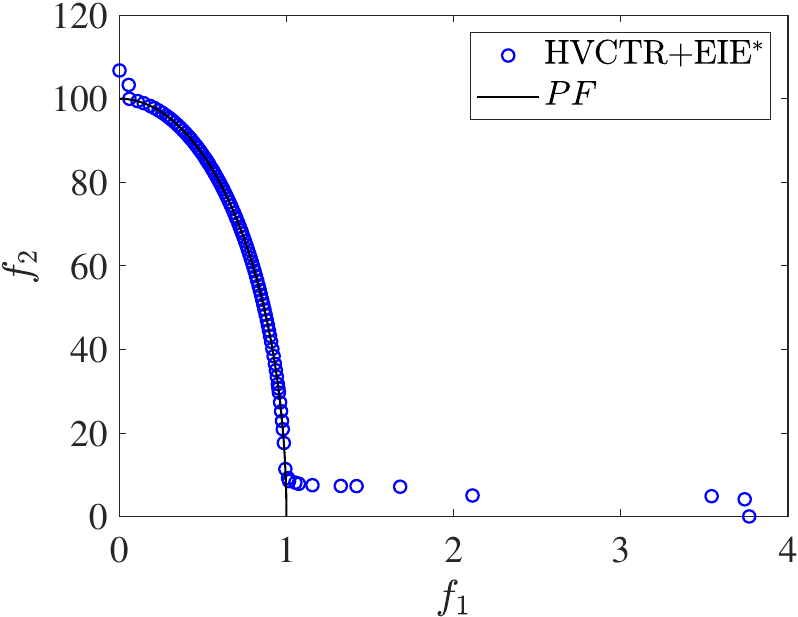}}
\subfloat[MOP2]{\includegraphics[width=0.245\linewidth]{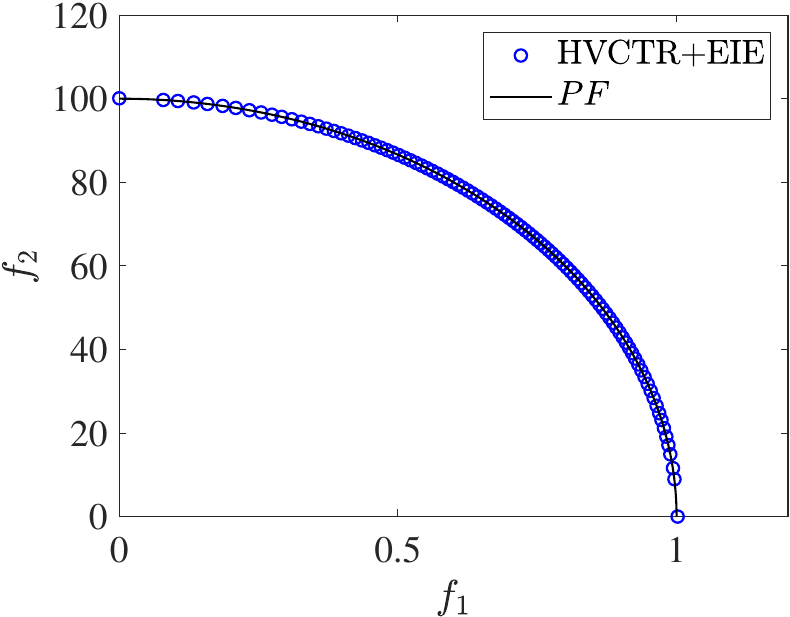}}
\hfil
\subfloat[MOP10]{\includegraphics[width=0.245\linewidth]{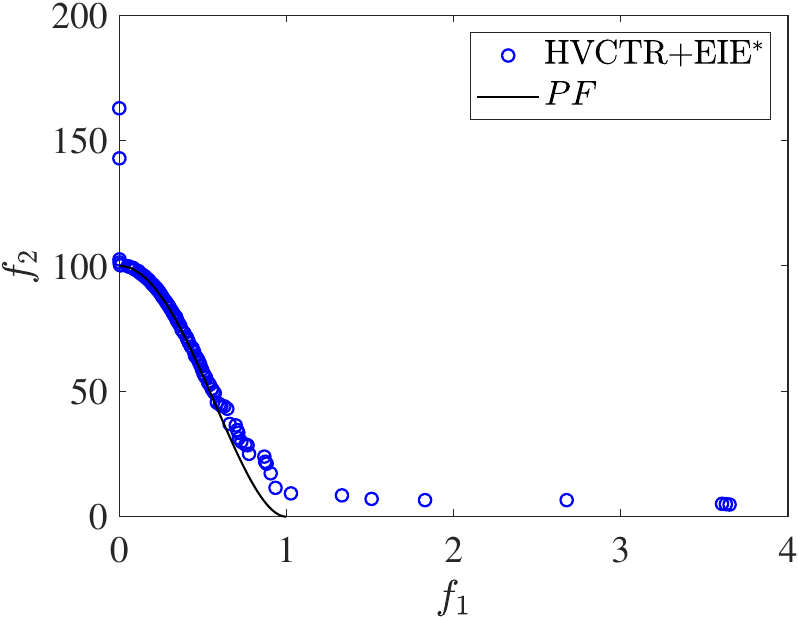}}
\subfloat[MOP10]{\includegraphics[width=0.245\linewidth]{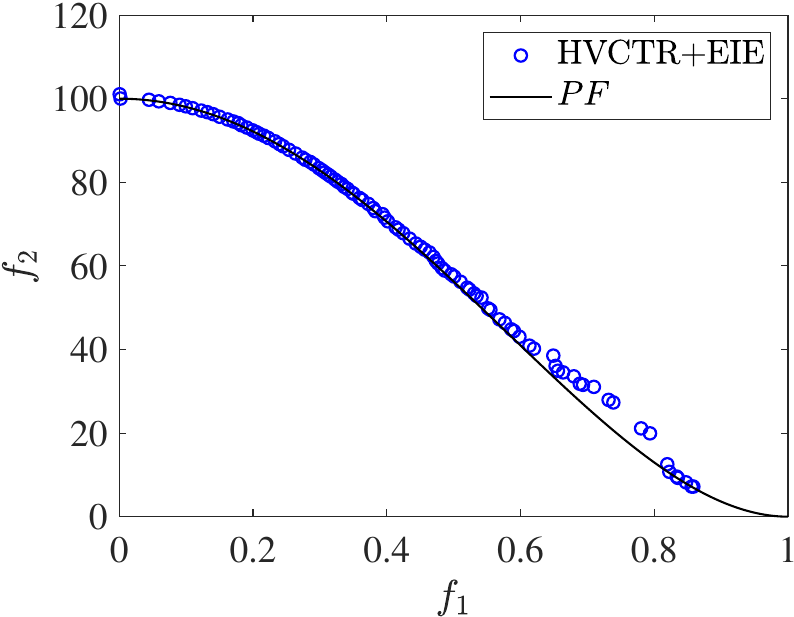}}
\subfloat[MOP16]{\includegraphics[width=0.245\linewidth]{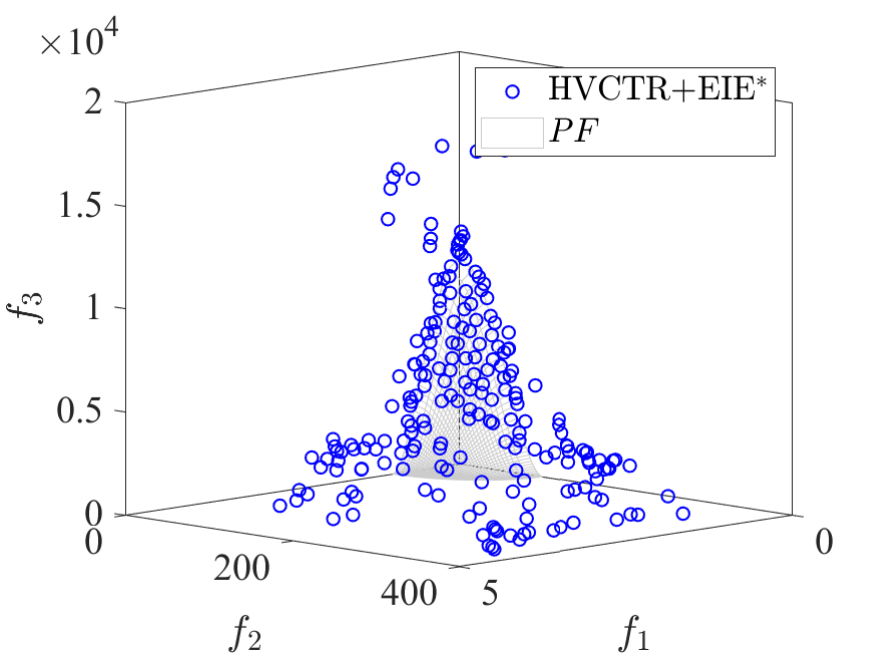}}
\subfloat[MOP16]{\includegraphics[width=0.245\linewidth]{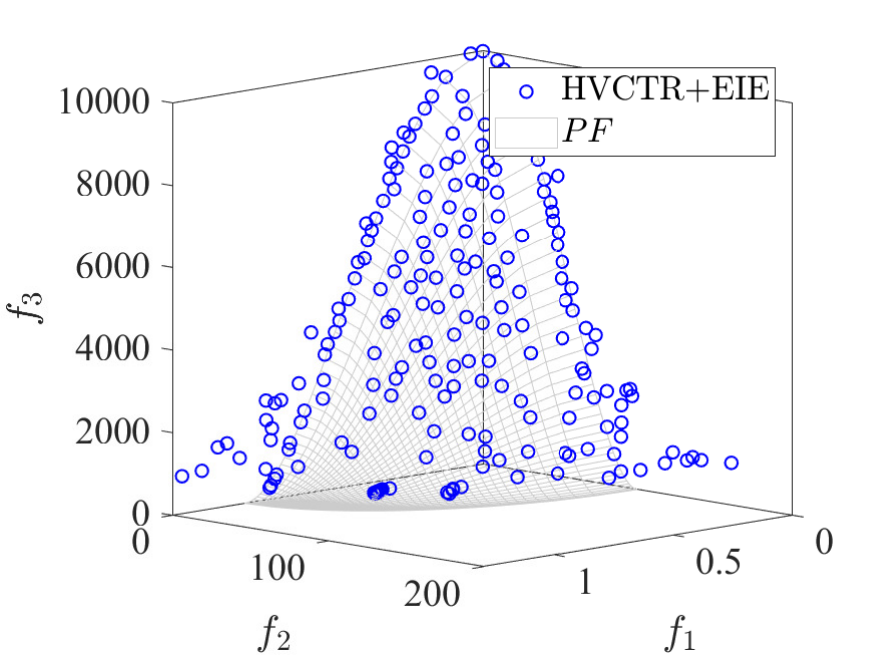}}
\caption{Plots of the non-dominated sets with the median $\operatorname{HV}$ metric values found by MOEA with EIE$^*$ and MOEA with EIE (first group).}
\label{fig:ndobjs_ews}
\end{figure*}

\begin{figure*}[ht]
\centering
\subfloat[MOP9]{\includegraphics[width=0.245\linewidth]{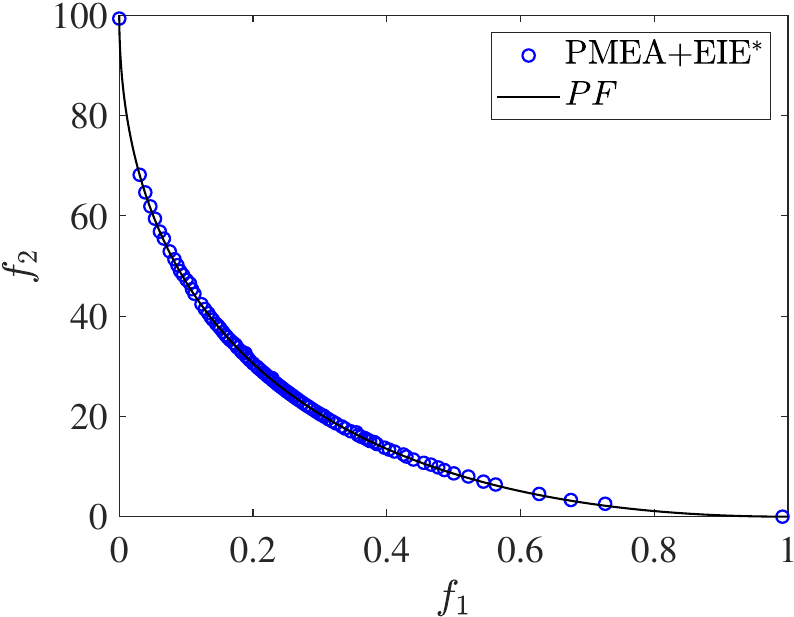}}
\subfloat[MOP9]{\includegraphics[width=0.245\linewidth]{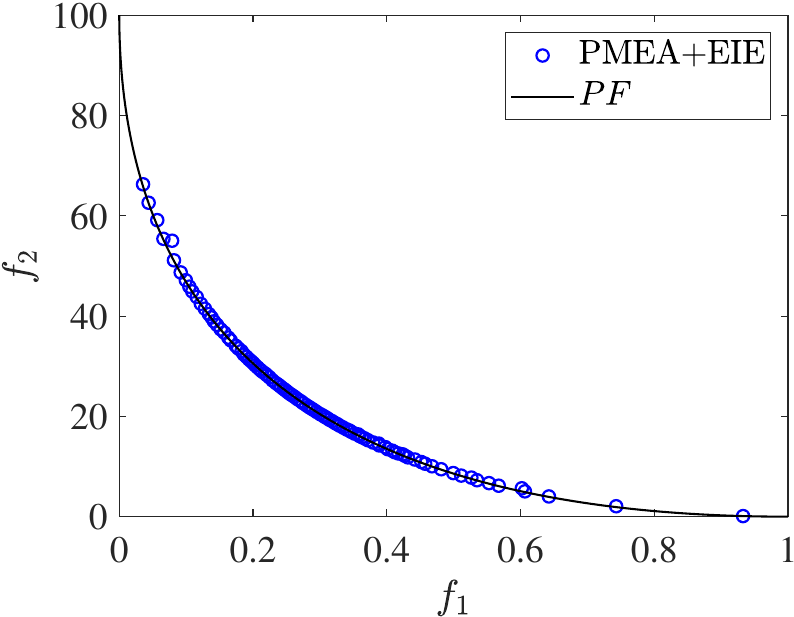}}
\subfloat[MOP11]{\includegraphics[width=0.245\linewidth]{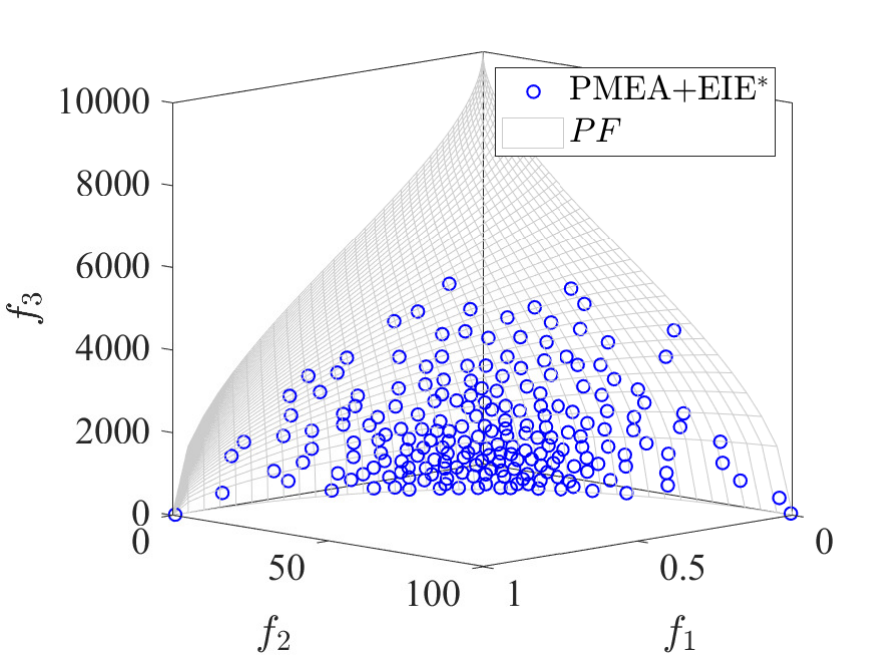}}
\subfloat[MOP11]{\includegraphics[width=0.245\linewidth]{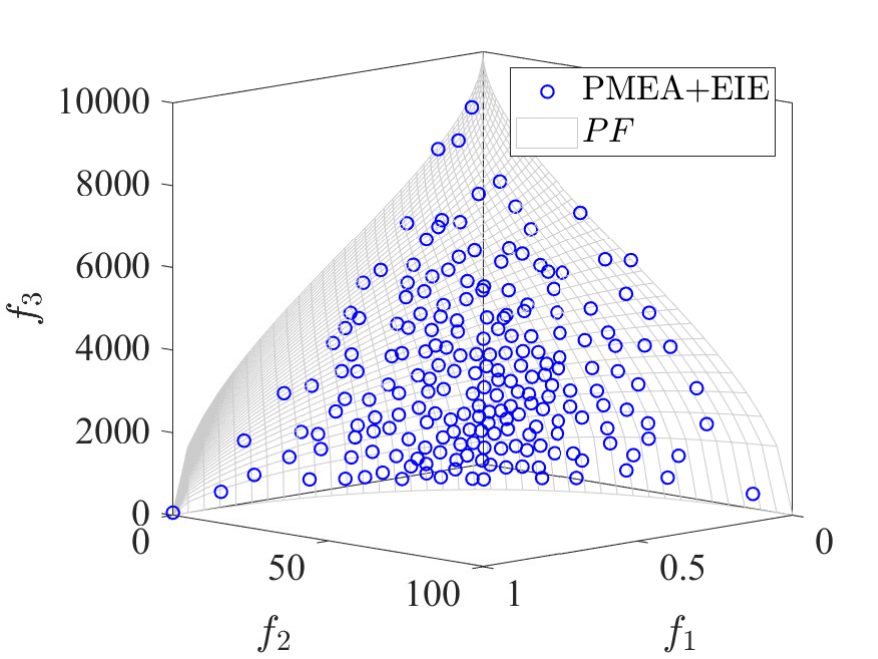}}  
\hfil
\subfloat[MOP13]{\includegraphics[width=0.245\linewidth]{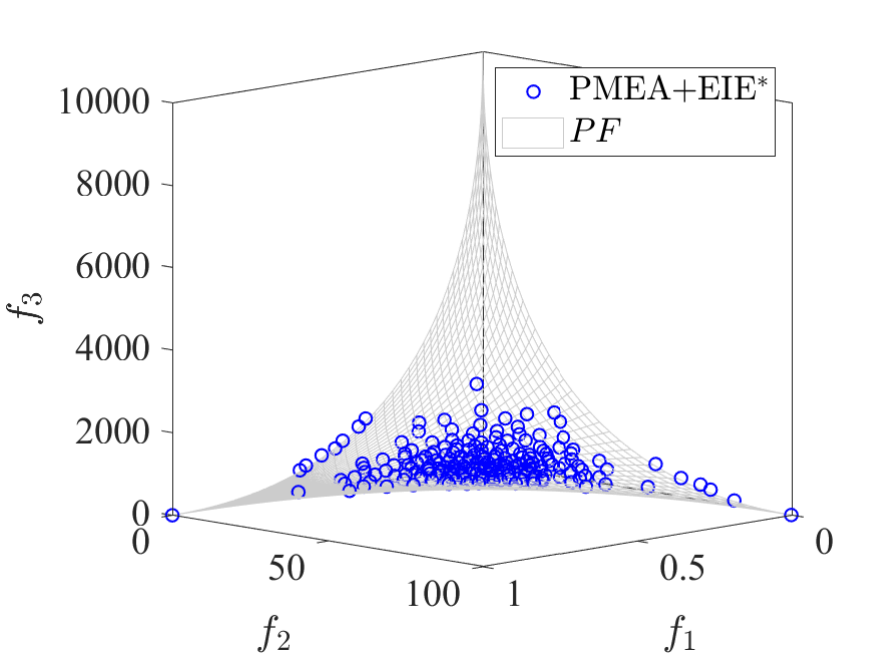}}
\subfloat[MOP13]{\includegraphics[width=0.245\linewidth]{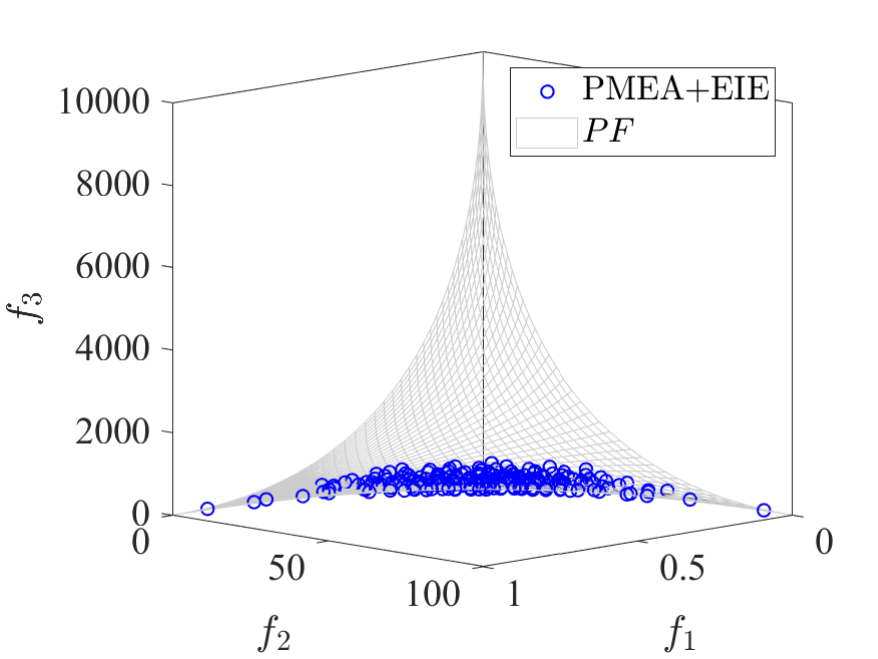}}
\subfloat[MOP14]{\includegraphics[width=0.245\linewidth]{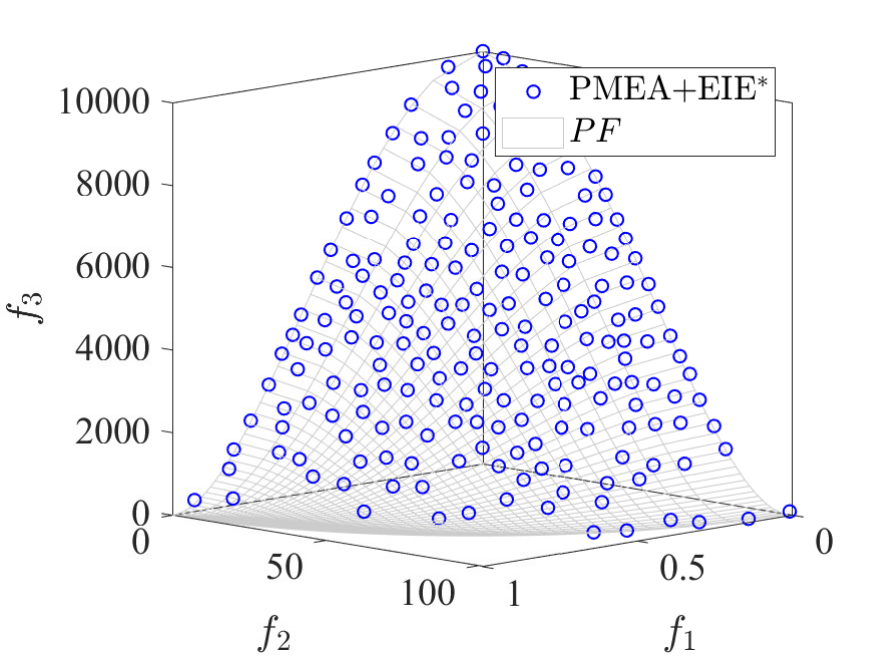}}
\subfloat[MOP14]{\includegraphics[width=0.245\linewidth]{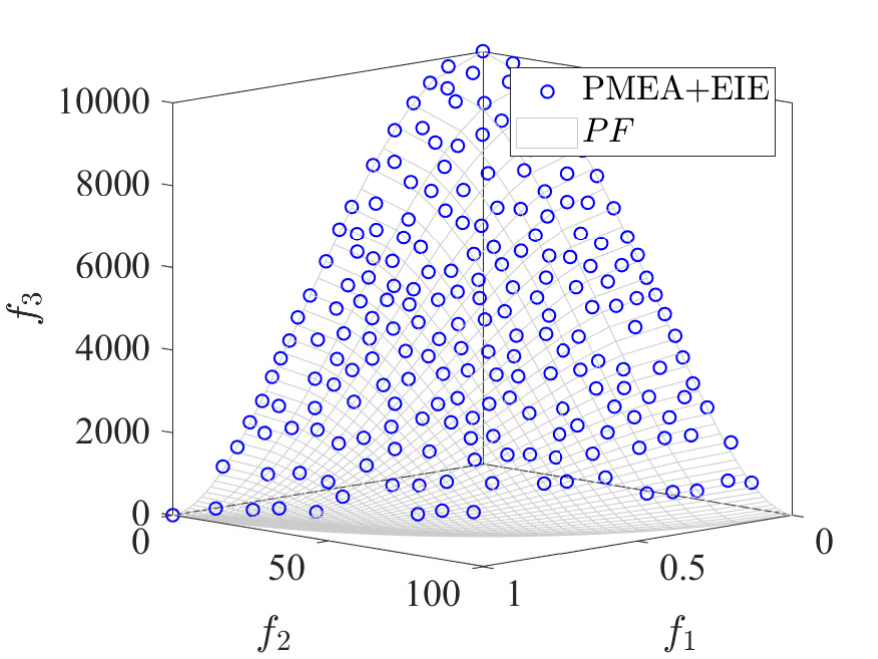}}
\hfil
\subfloat[MOP15]{\includegraphics[width=0.245\linewidth]{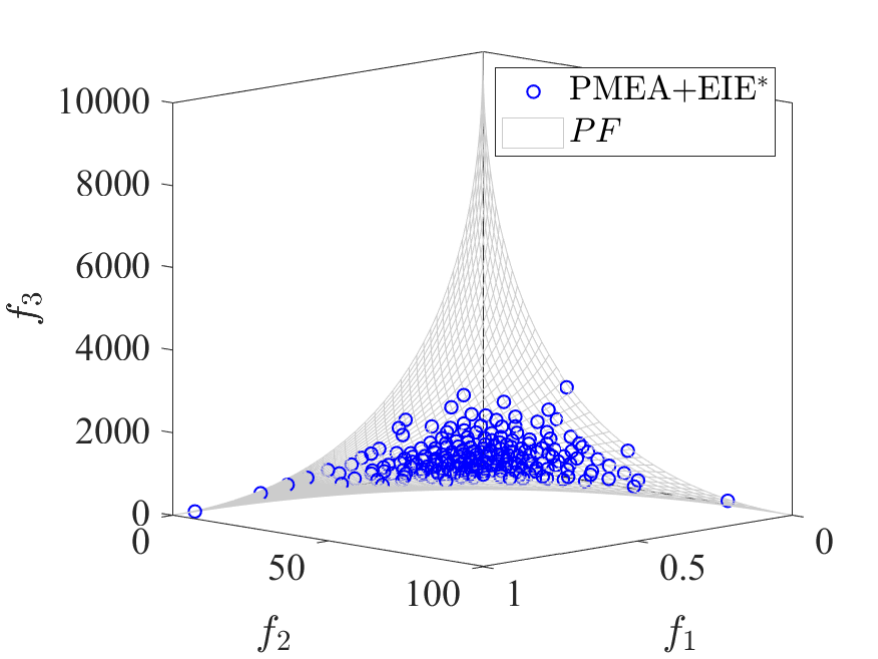}}
\subfloat[MOP15]{\includegraphics[width=0.245\linewidth]{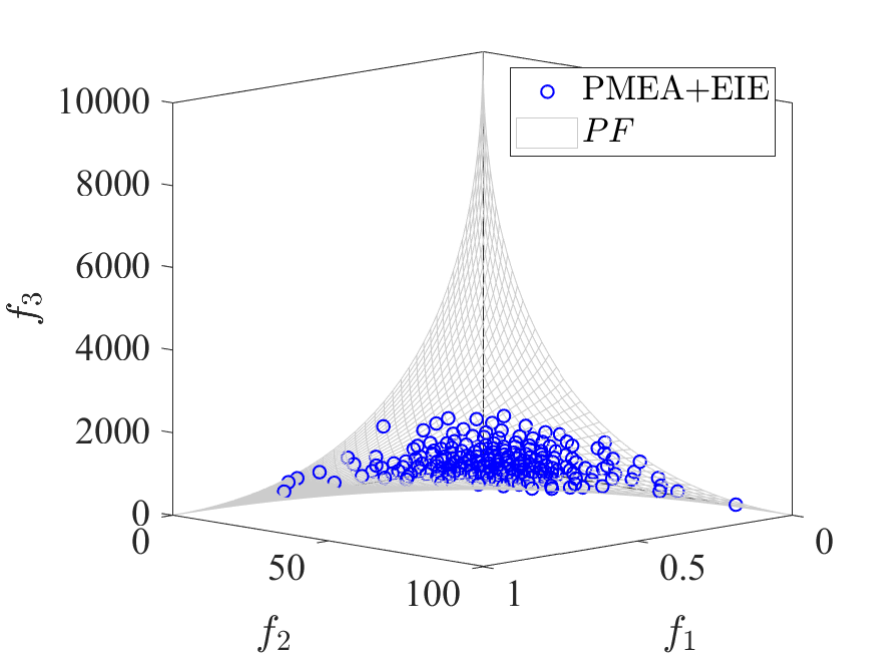}}
\subfloat[MOP11]{\includegraphics[width=0.245\linewidth]{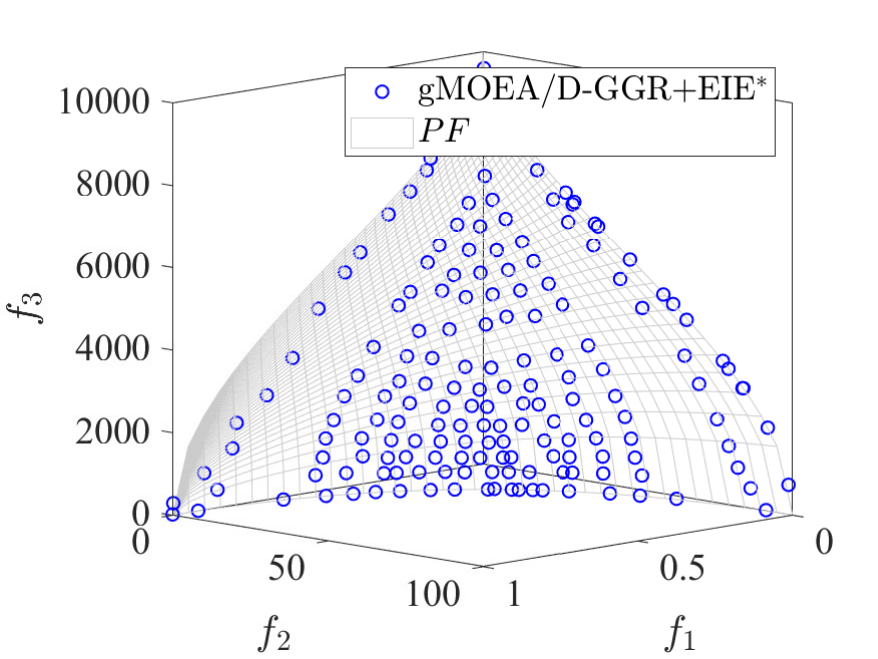}}
\subfloat[MOP11]{\includegraphics[width=0.245\linewidth]{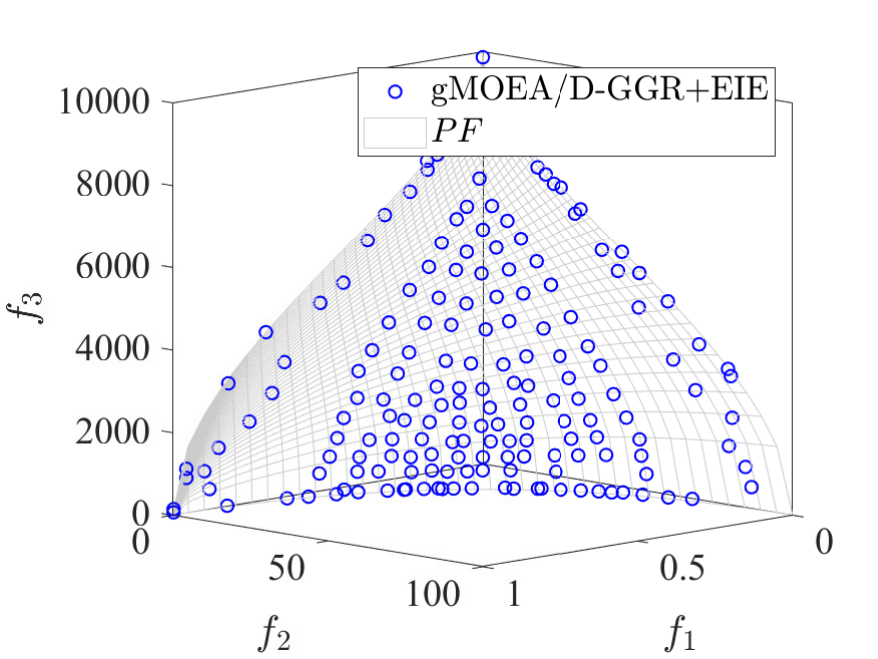}}
\hfil
\subfloat[MOP7]{\includegraphics[width=0.245\linewidth]{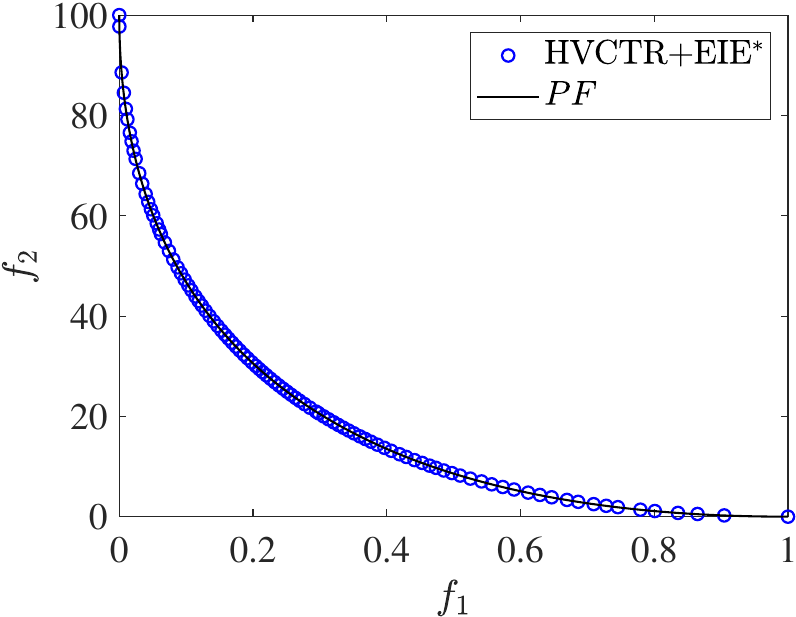}}
\subfloat[MOP7]{\includegraphics[width=0.245\linewidth]{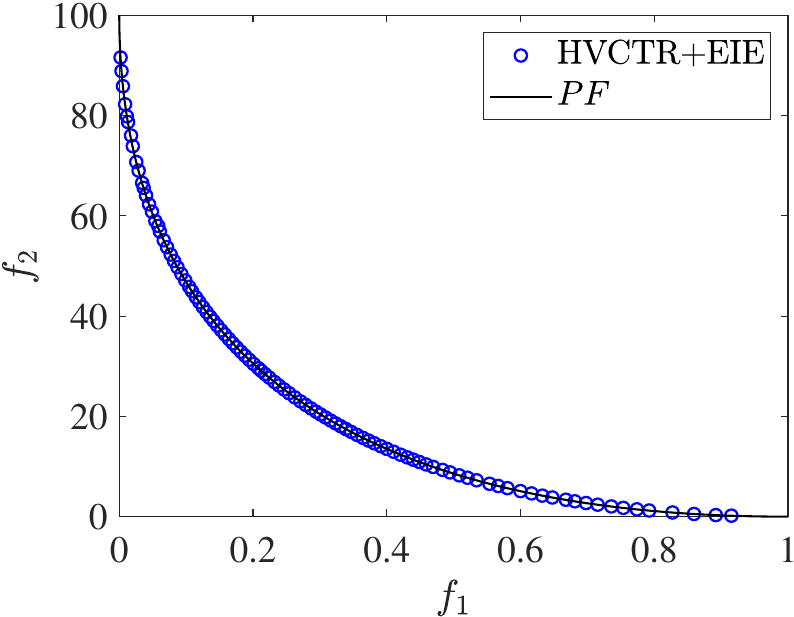}}
\subfloat[MOP9]{\includegraphics[width=0.245\linewidth]{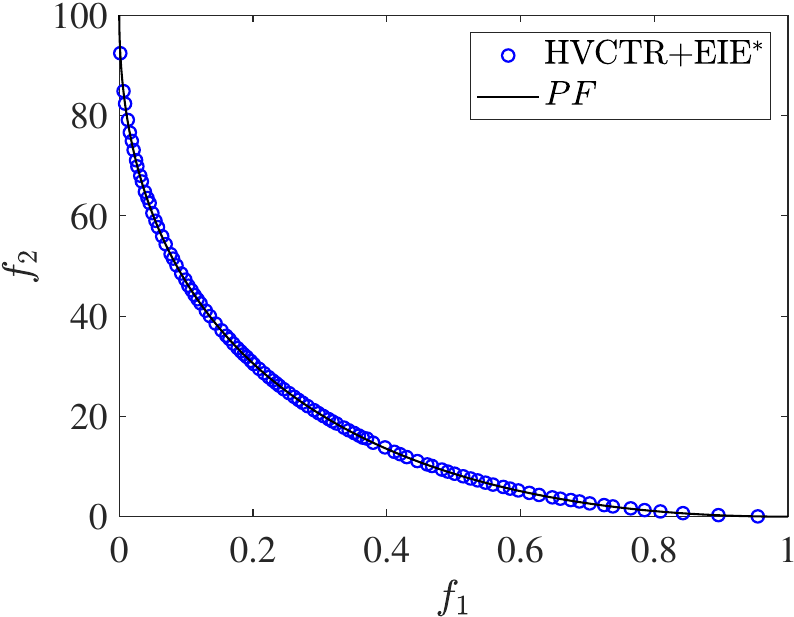}}
\subfloat[MOP9]{\includegraphics[width=0.245\linewidth]{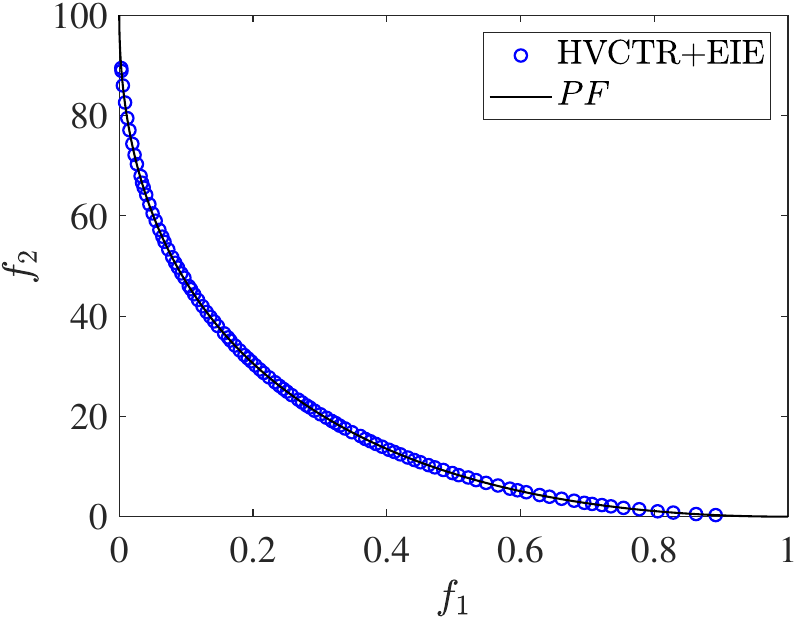}}
\hfil
\subfloat[MOP13]{\includegraphics[width=0.245\linewidth]{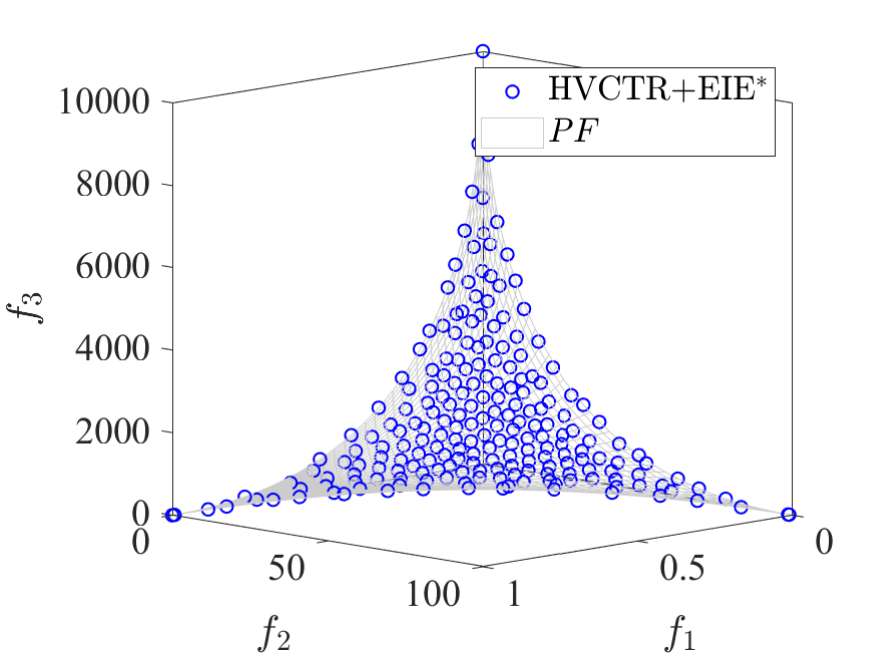}}
\subfloat[MOP13]{\includegraphics[width=0.245\linewidth]{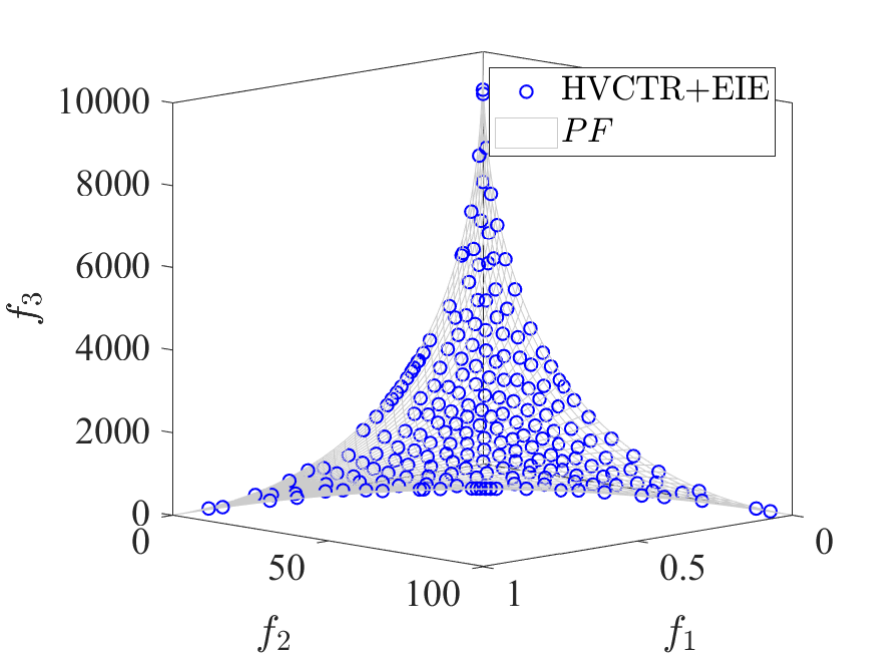}}
\subfloat[MOP15]{\includegraphics[width=0.245\linewidth]{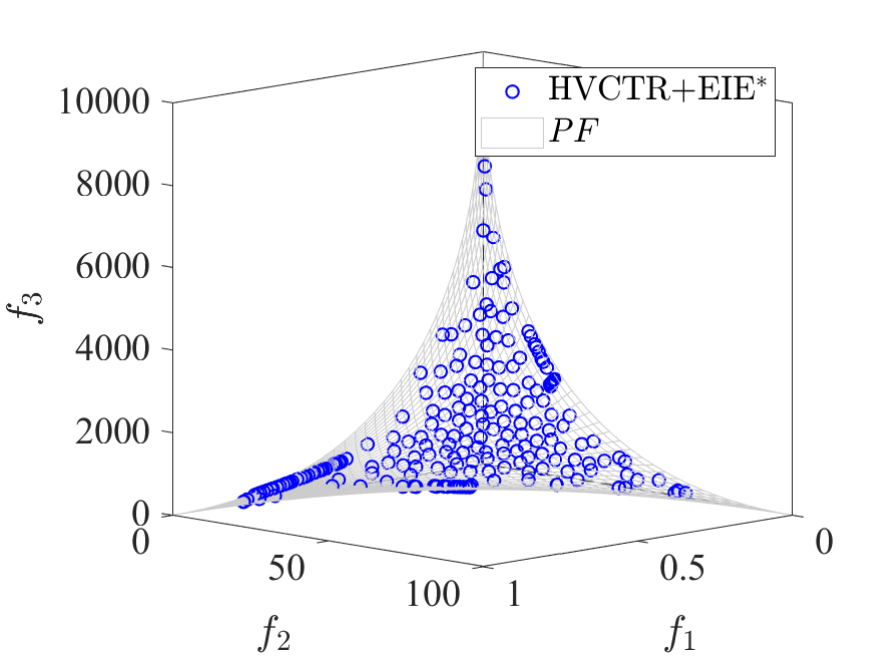}}
\subfloat[MOP15]{\includegraphics[width=0.245\linewidth]{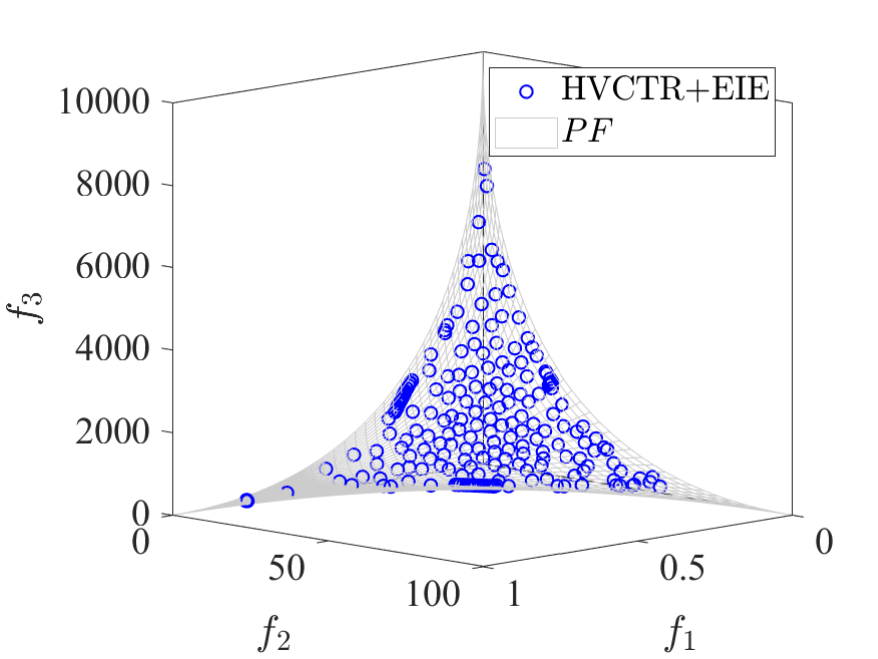}}
\caption{Plots of the non-dominated sets with the median $\operatorname{HV}$ metric values found by MOEA with EIE$^*$ and MOEA with EIE (all instances in the second group where the performance of MOEA with EIE degrades).}
\label{fig:ndobjs_ews_second}
\end{figure*}

\begin{figure*}[ht]
\centering
\subfloat[MOP11$^{-1}$]{\includegraphics[width=0.245\linewidth]{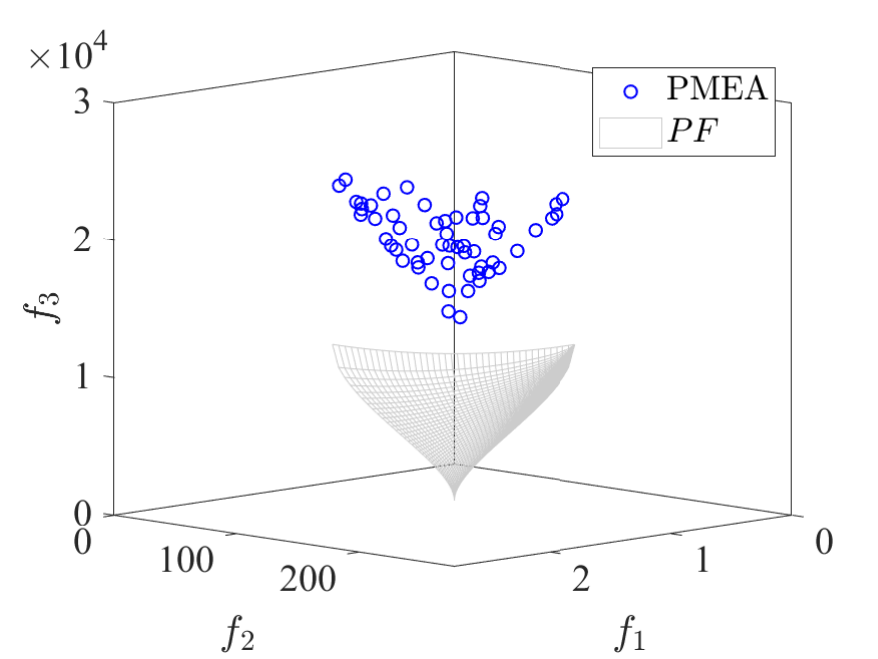}}
\subfloat[MOP11$^{-1}$]{\includegraphics[width=0.245\linewidth]{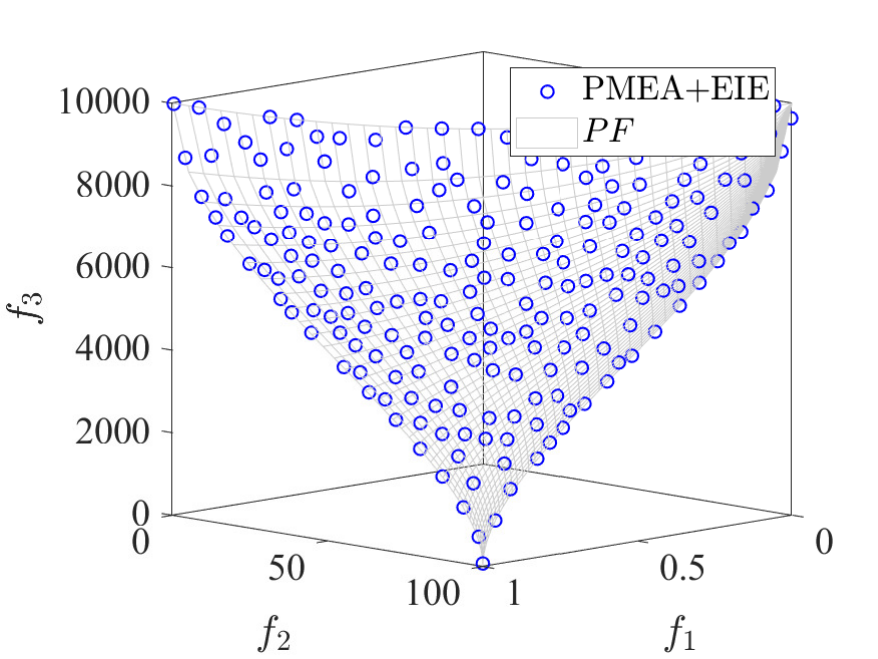}}
\subfloat[MOP15$^{-1}$]{\includegraphics[width=0.245\linewidth]{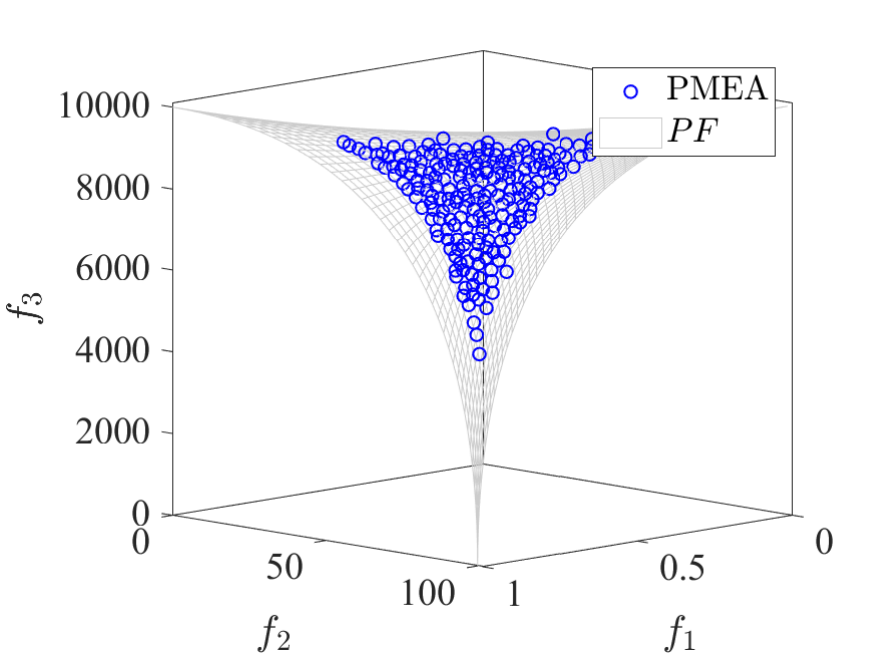}}
\subfloat[MOP15$^{-1}$]{\includegraphics[width=0.245\linewidth]{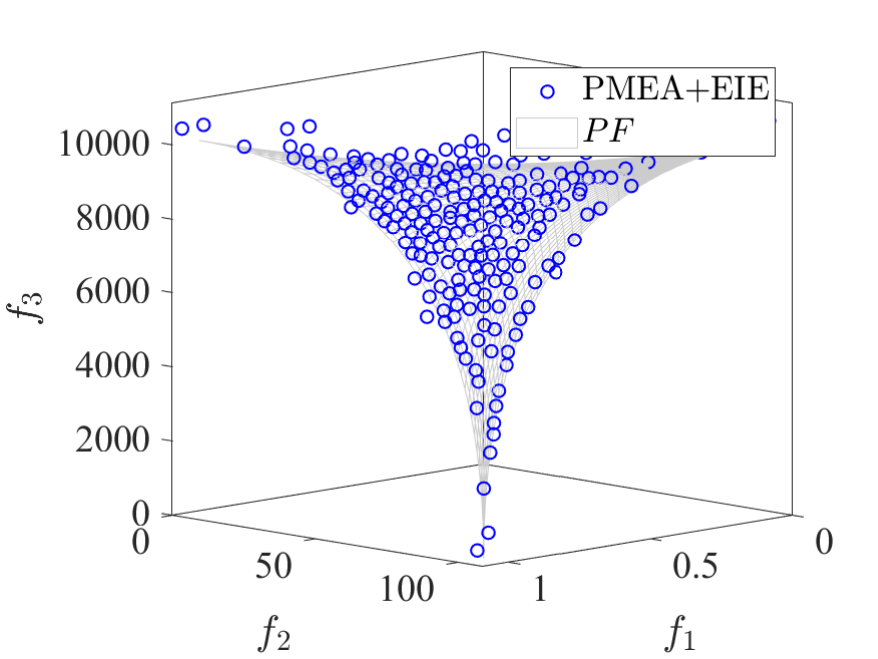}}
\hfil
\subfloat[MOP11$^{-1}$]{\includegraphics[width=0.245\linewidth]{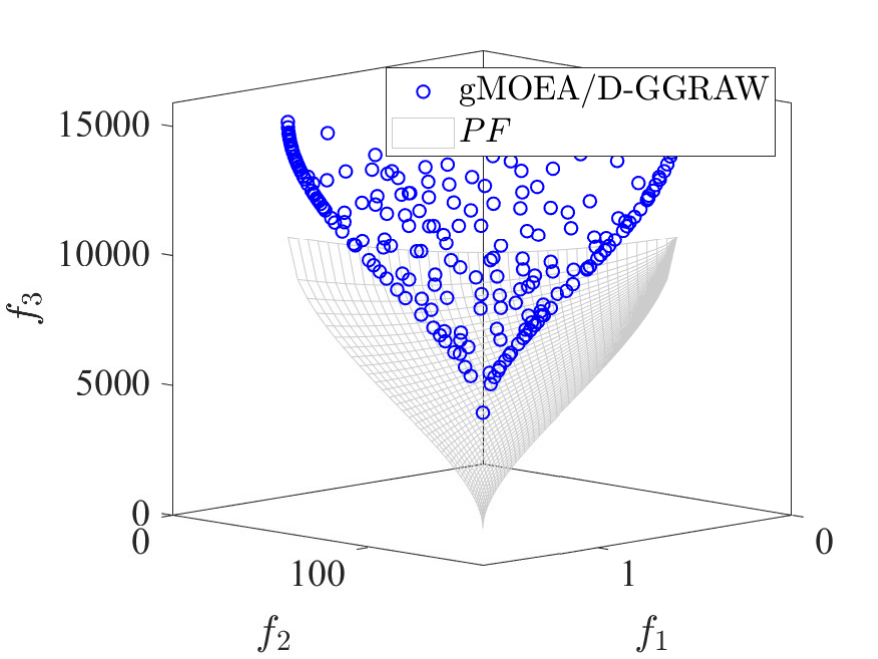}}
\subfloat[MOP11$^{-1}$]{\includegraphics[width=0.245\linewidth]{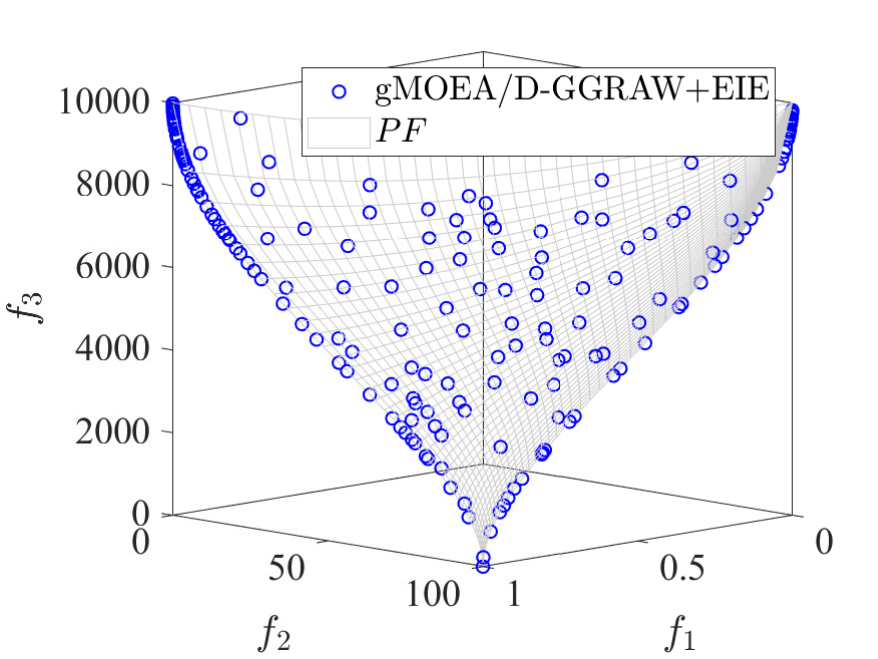}}
\subfloat[MOP15$^{-1}$]{\includegraphics[width=0.245\linewidth]{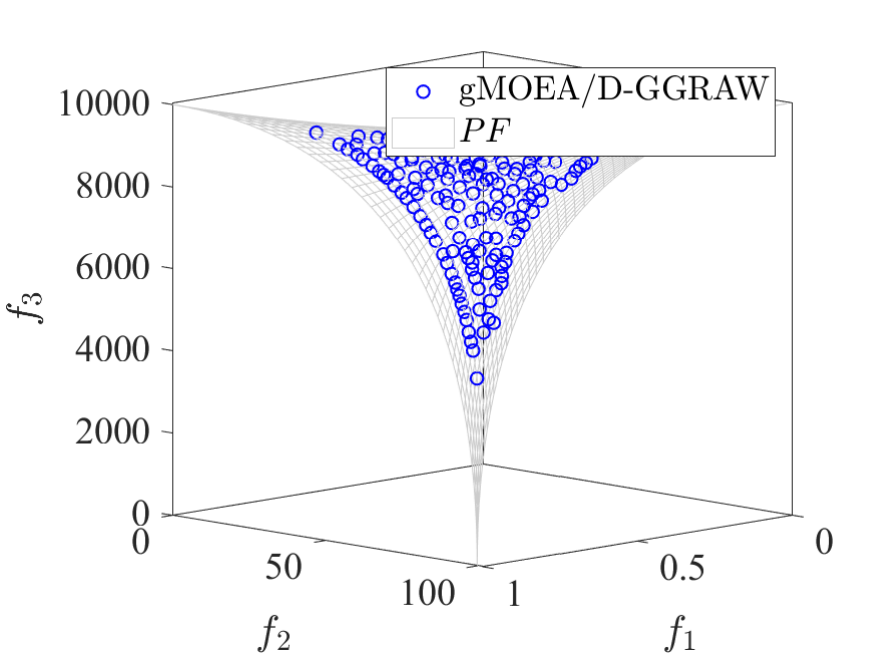}}
\subfloat[MOP15$^{-1}$]{\includegraphics[width=0.245\linewidth]{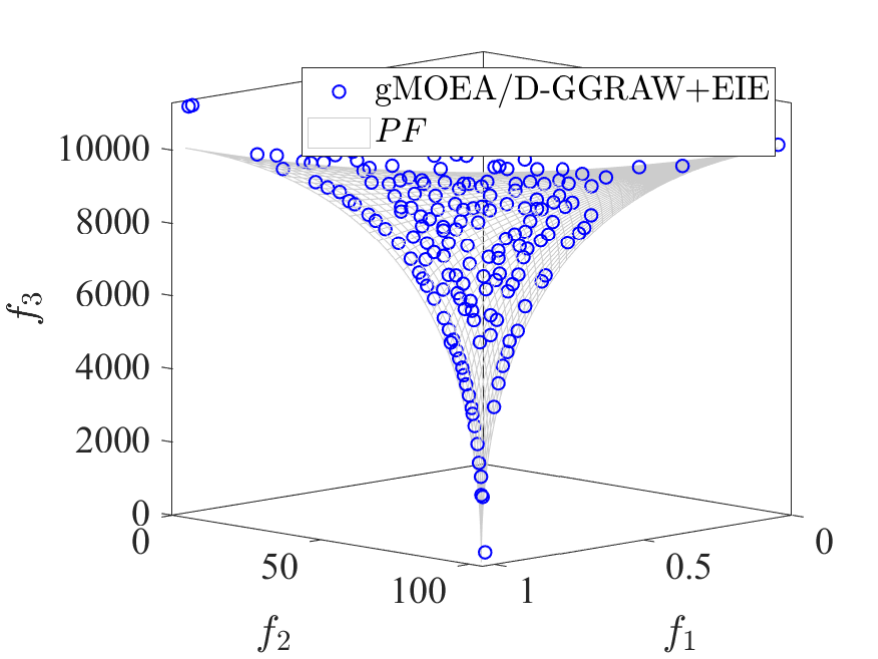}}
\hfil
\subfloat[MOP11$^{-1}$]{\includegraphics[width=0.245\linewidth]{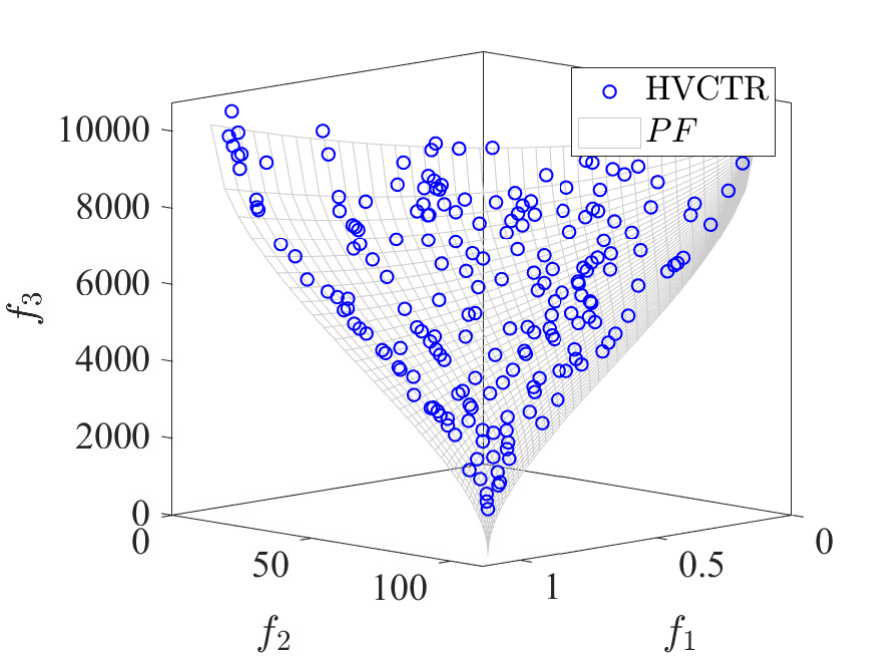}}
\subfloat[MOP11$^{-1}$]{\includegraphics[width=0.245\linewidth]{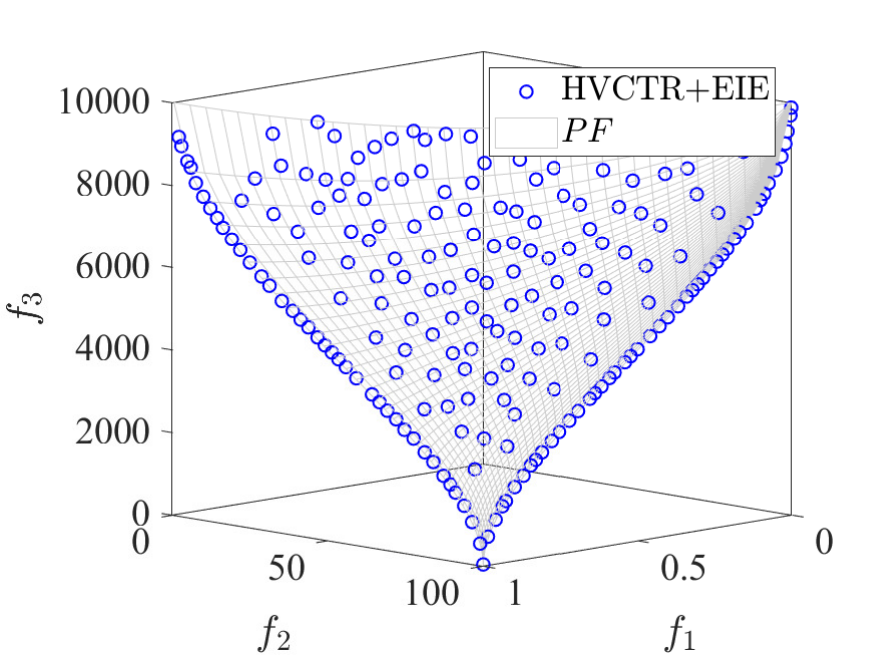}}
\subfloat[MOP15$^{-1}$]{\includegraphics[width=0.245\linewidth]{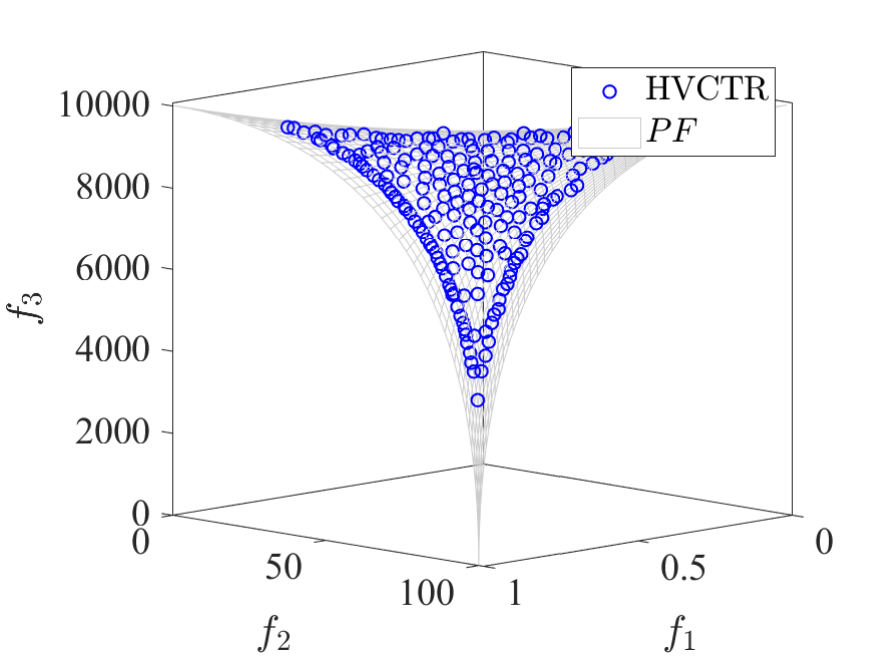}}
\subfloat[MOP15$^{-1}$]{\includegraphics[width=0.245\linewidth]{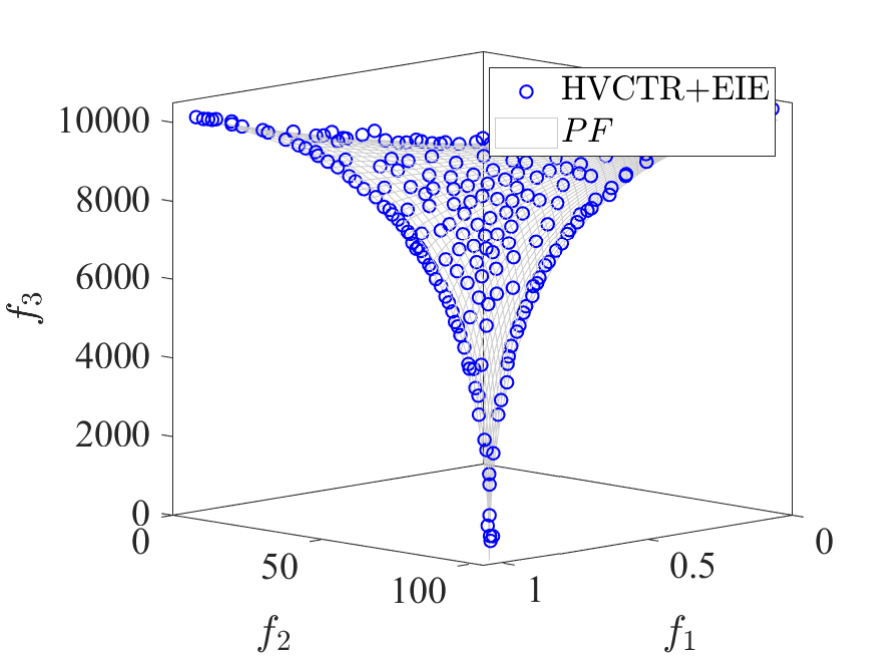}}
\caption{Plots of the non-dominated sets with the median $\operatorname{HV}$ metric values found by the MOEA and the MOEA with EIE (MOP11$^{-1}$-MOP16$^{-1}$).}
\label{fig:ndobjs_baseline_inv}
\end{figure*}
}

\vfill

\end{document}